\documentclass{article}

\usepackage{arxiv}

\usepackage[utf8]{inputenc} 
\usepackage[T1]{fontenc}    
\usepackage{hyperref}       
\usepackage{url}            
\usepackage{booktabs}       
\usepackage{amsfonts}       
\usepackage{nicefrac}       
\usepackage{microtype}      
\usepackage{lipsum}
\usepackage{graphicx}
\graphicspath{ {./images/} }

\usepackage{graphicx}
\usepackage[utf8]{inputenc}
\usepackage{caption}
\usepackage{amssymb}
\usepackage{amsmath}
\usepackage{algorithm}
\usepackage[noend]{algorithmic}
\usepackage{cite}
\usepackage{float}
\usepackage{caption,subcaption}

\usepackage{siunitx}
\usepackage{booktabs}

\title{Data embedding and prediction by sparse tropical matrix factorization}

\author{
Amra Omanović \\
  Faculty of Computer and Information Science\\
  University of Ljubljana\\
  Večna pot 113, 1000 Ljubljana, Slovenia\\
  \texttt{amra.omanovic@fri.uni-lj.si} \\
   \And
Hilal Kazan \\
  Department of Computer Engineering \\
  Antalya Bilim University \\
  Çıplaklı, Akdeniz Blv. No:290/A, 07190 Antalya, Turkey \\
  \texttt{hilal.kazan0@antalya.edu.tr} \\
  \And
Polona Oblak \\
  Faculty of Computer and Information Science\\
  University of Ljubljana\\
  Večna pot 113, 1000 Ljubljana, Slovenia\\
  \texttt{polona.oblak@fri.uni-lj.si} \\
\And
Tomaž Curk \\
Faculty of Computer and Information Science\\
  University of Ljubljana\\
  Večna pot 113, 1000 Ljubljana, Slovenia\\
  \texttt{tomaz.curk@fri.uni-lj.si} \\
}

\date{}

\begin{document}
\maketitle

\begin{abstract}
Matrix factorization methods are linear models, with limited capability to model complex relations.
In our work, we use tropical semiring to introduce non-linearity into matrix factorization models.
We propose a method called \textit{Sparse Tropical Matrix  Factorization} (\texttt{STMF}) for the estimation of missing (unknown) values.

We evaluate the efficiency of the \texttt{STMF} method on both synthetic data and biological data in the form of gene expression measurements downloaded from The Cancer Genome Atlas (TCGA) database. 
Tests on unique synthetic data showed that \texttt{STMF} approximation achieves a higher correlation than non-negative matrix factorization (\texttt{NMF}), which is unable to recover patterns effectively.
On real data, \texttt{STMF} outperforms \texttt{NMF} on six out of nine gene expression datasets. While \texttt{NMF} assumes normal distribution and tends toward the mean value, \texttt{STMF} can better fit to extreme values and distributions.

\texttt{STMF} is the first work that uses tropical semiring on sparse data.
We show that in certain cases semirings are useful because they consider the structure, which is different and simpler to understand than it is with standard linear algebra.
\end{abstract}


\section{Introduction}
Matrix factorization methods are getting increasingly popular in many research areas~\cite{koren2009matrix, xu2003document, brunet2004metagenes}. These methods generate linear models, which cannot model complex relationships.
Our work focuses on incorporating non-linearity into matrix factorization models by using tropical semiring. 

The motivation for using tropical matrix factorization can be seen in the classic example of movie rating data, where a users-by-movies matrix contains the rating users assigned to movies. In standard matrix factorization methods, it is assumed that a user's final rating is a linear combination of some factors (a person likes some movie because of the director, the genre, the lead actor, etc.). But it is also possible that some factor is so dominant that all others are irrelevant. An example given for the \texttt{Latitude} algorithm~\cite{latitude}, a person likes all Star Wars movies irrespective of actors or directors, shows that using the $\max$ operator instead of the sum might produce a better model.

We develop a method for the prediction of missing (unknown) values, called \textit{Sparse Tropical Matrix Factorization} (\texttt{STMF}). We evaluate its performance on the prediction of gene expression measurements from The Cancer Genome Atlas Research Network (TCGA) database. 
We show that the newly defined operations can discover patterns, which cannot be found with standard linear algebra.

\section{Related work}
\label{sec:related work}
Matrix factorization is a data embedding model which gives us a more compact representation of the data and simultaneously finds a latent structure.
The most popular example is the non-negative matrix factorization (\texttt{NMF})~\cite{nmf}, where the factorization is restricted to the matrices with non-negative entries. This non-negativity in resulting factor matrices makes the results easier to interpret. One of the applications of matrix factorization methods is for recommender systems, where users and items are represented in a lower-dimensional latent space~\cite{mfrecommender}. Binary matrix factorization (\texttt{BMF})\cite{bmf, bmf_1} is a variant rooted from \texttt{NMF} where factor matrices are binary, while probabilistic non-negative matrix factorization (\texttt{PMF})\cite{pmf, plsa} models the data as a multinomial distribution. Integrative approaches, which use standard linear algebra to simultaneously factorize multiple data sources and improve predictive accuracy, are reviewed in~\cite{survey}. Multi-omic and multi-view clustering methods like \texttt{MultiNMF}~\cite{multi_nmf}, \texttt{Joint NMF}~\cite{joint_nmf}, \texttt{PVC}~\cite{pvc}, \texttt{DFMF}~\cite{datafusion} and \texttt{iONMF}~\cite{orthogonal} can be used for data fusion of multiple data sources.

Lately, subtropical semiring $(\max, \cdot)$ gained interest in the field of machine learning, since it can discover interesting patterns~\cite{algorithms, capricorn}. By taking the logarithm of the subtropical semiring, we obtain the tropical one\cite{cancer}. Although these two semirings are isomorphic, the factorization in tropical semiring works differently than the factorization in subtropical semiring.
The \texttt{Cancer} algorithm~\cite{cancer} works with continuous data, performing subtropical matrix factorization (\texttt{SMF}) on the input matrix. Two main components of the algorithm are: iteratively updating the rank-1 factors one-by-one and approximate the max-times reconstruction error with a low-degree polynomial. \texttt{Latitude} algorithm~\cite{latitude} combines \texttt{NMF} and \texttt{SMF}, where factors are interpreted as \texttt{NMF} features, \texttt{SMF} features or as mixtures of both. This approach gives good results in cases where the underlying data generation process is a mixture of the two processes. In~\cite{nonliner_recomm} authors used subtropical semiring as part of a recommender system. We can consider their method to be a particular kind of neural network. Le Van et al.~\cite{semiring} presented a single generic framework that is based on the concept of semiring matrix factorization. They applied the framework on two tasks: sparse rank matrix factorization and rank matrix tiling. 

De Schutter \& De Moor~\cite{tropical} presented a heuristic algorithm to compute factorization of a matrix in the tropical semiring, which we denote as \textit{Tropical Matrix Factorization} (\texttt{TMF}). They use it to determine the minimal system order of a discrete event system (\texttt{DES}). In the last decades, there has been an increase of interest in this research area, and \texttt{DES} is modeled as a max-plus-linear (\texttt{MPL}) system~\cite{baccelli1992synchronization, heidergott2014max}. In contrast to \texttt{TMF} where approximation error is reduced gradually, convergence is not guaranteed in the \texttt{Cancer} algorithm. Both \texttt{Cancer} and \texttt{TMF} return factors that encode the most dominant components in the data. However, by their construction, they cannot be used for prediction tasks in different problem domains, such as predicting gene expression. In contrast with the \texttt{NMF} method and its variants, which require non-negative data, \texttt{TMF} can work with negative values. 

Hook~\cite{linear_regression} reviewed algorithms and applications of linear regression over the max-plus semiring, while Gärtner and Jaggi~\cite{tropical_svm} constructed a tropical analogue of support vector machines (\texttt{SVM}), which can be used to classify data into more than just two classes compared to the classical \texttt{SVM}.
Zhang et al.~\cite{tropical_geometry} in their work establish a connection between neural networks and tropical geometry. They showed that linear regions of feedforward neural networks with rectified linear unit activation correspond to vertices of polytopes associated with tropical rational functions. Therefore, to understand specific neural networks, we need to understand relevant tropical geometry. 
Since one goal in biology is not just to model the data, but also to understand the underlying mechanisms, the matrix factorization methods can give us a more straightforward interpretation than neural networks.

In our work, we answer the question stated in \texttt{Cancer}: can tropical factorization be used, in addition to data analysis, also in other data mining and machine learning tasks, e.g. matrix completion? We propose a method \texttt{STMF}, which is based on \texttt{TMF}, and it can simultaneously predict missing values, i.e. perform matrix completion.
In Table~\ref{table_diff} we compare the most relevant methods for our work. 
To the best of our knowledge \texttt{STMF} is the only method which performs prediction tasks in tropical semiring.

\begin{center}
    \setlength{\tabcolsep}{0.5em} 
    {\renewcommand{\arraystretch}{1.2}
    \scalebox{0.9}{
    \begin{tabular}{c|c|c|c|c}
    \hline
    & arithmetic & data sources  & prediction tasks & convergence \\ \hline
    \texttt{NMF}~\cite{nmf}, \texttt{BMF}~\cite{bmf}, \texttt{PMF}~\cite{pmf} & standard & single & yes & yes  \\
    \texttt{DFMF}~\cite{datafusion}, \texttt{iONMF}~\cite{orthogonal} & standard & multiple & yes & yes  \\
    \texttt{Latitude}~\cite{latitude} & standard \& subtropical & single & no & no \\
    \texttt{Cancer}~\cite{cancer} & subtropical & single & no & no \\
    \texttt{TMF}~\cite{tropical} & tropical & single & no & yes \\
    \texttt{STMF} & tropical & single & yes & yes \\ \hline
    \end{tabular}}}
\captionof{table}{A comparison between different matrix factorization methods.}
\label{table_diff}
\end{center}

\section{Methods}
\subsection{Tropical semiring and factorization}
Now, we give some formal definitions regarding the tropical semiring.
The \emph{$(\max,+)$ semiring} or \emph{tropical semiring} $\mathbb{R}_{\max}$, is the set $\mathbb{R} \cup \{-\infty\}$, equipped with $\max$ as addition ($\oplus$), and $+$ as multiplication ($\otimes$). 
For example, $2\oplus3=3$ and $1\otimes1=2$.
On the other hand, in the \emph{subtropical semiring} or \emph{$(\max,\times)$ semiring}, defined on the same set $\mathbb{R} \cup \{-\infty\}$, addition ($\max$) is defined as in the tropical semiring, but the multiplication is the standard multiplication ($\times$). Throughout the paper, symbols $+$ and $-$ refer to standard operations of addition and subtraction.
Tropical semiring can be used for optimal control\cite{optimal_control}, asymptotics\cite{asymptotics}, discrete event systems\cite{minmax} or solving a decision problem \cite{finite_automata}. Another example is the well-known game Tetris, which can be linearized using the $(\max,+)$ semiring\cite{gaubert}. 

Let $\mathbb{R}_{\max}^{m \times n}$ define the set of all $m \times n$ matrices over tropical semiring. For $A\in \mathbb{R}_{\max}^{m \times n}$ we denote by $A_{ij}$ the entry in the $i$-th row and the $j$-th column of matrix $A$. We denote the  \emph{sum of matrices} $A, B \in \mathbb{R}_{\max}^{m \times n}$ as  $A\oplus B \in \mathbb{R}_{\max}^{m \times n}$ and define its entries as
\begin{equation*}
    (A \oplus B)_{ij} = A_{ij} \oplus B_{ij} = \max\{A_{ij}, B_{ij}\},
\end{equation*}
$i=1,\ldots,m$, $j=1,\ldots,n$. 
The \emph{product of} matrices $A \in \mathbb{R}_{\max}^{m \times p}$, $B \in \mathbb{R}_{\max}^{p \times n}$ is denoted by $A\otimes B \in \mathbb{R}_{\max}^{m \times n}$ and its entries are defined as 
\begin{equation*}
    (A \otimes B)_{ij} = \bigoplus\limits_{k=1}^{p} A_{ik} \otimes B_{kj} = \max\limits_{1\leq k \leq p}\{A_{ik} + B_{kj}\},
\end{equation*}
$i=1,\ldots,m$, $j=1,\ldots,n$.

\emph{Matrix factorization over a tropical semiring} is a decomposition of a form
$R= U \otimes V$, where
$R \in \mathbb{R}_{\max}^{m \times n}$, $U \in \mathbb{R}_{\max}^{m \times r}$,  $V \in \mathbb{R}_{\max}^{r \times n}$ and $r \in \mathbb{N}_0$. 
Since for small values of $r$ such decomposition may not exist, we state tropical matrix factorization problem as:
given a matrix $R$ and factorization rank $r$, find matrices $U$ and $V$ such that 
\begin{equation}
    \label{eq:factorization}
    R\cong U \otimes V.
\end{equation}

To implement a tropical matrix factorization algorithm, we need to know how to solve tropical linear systems. Methods for solving linear systems over tropical semiring differ substantially from methods that use standard linear algebra~\cite{gaubert}. 

We define the \emph{ordering} in tropical semiring as $z \preceq w$ if and only if $z \oplus w=w$ for $z,w \in \mathbb{R}_{\max}$, and it induces the ordering on vectors and matrices over tropical semiring entry-wise.
For $A\in \mathbb{R}_{\max}^{m \times n}$ and  $c=[c_k] \in \mathbb{R}_{\max}^{m}$ the  system of linear inequalities $A\otimes x \preceq c$ always has solutions 
and we call the solutions of $A \otimes x \preceq c$ the \textit{subsolutions} of the linear system  $A\otimes x=c$. The greatest subsolution $x=[x_1\, x_2\, \ldots x_n]^T$ of  $Ax =c$ can be computed by
\begin{equation} 
x_i =  \min_{1 \leq j \leq m} (c_j - A_{ji})
\label{greatest_subsolution}
\end{equation}
for $i=1,2,\ldots,n$. We will use \eqref{greatest_subsolution} in a column-wise form to solve the matrix equations.

\texttt{TMF} starts with an initial guess for the matrix $U$ in \eqref{eq:factorization}, denoted by $U_0$ and then computes $V$ as the greatest subsolution of $U_0 \otimes X = R$. Then authors use the iterative procedure by selecting and adapting an entry of $U$ or $V$ and recomputing it as the greatest subsolution of $Y \otimes V = R$ and $U \otimes X = R$, respectively. The \emph{$b$-norm} of matrix $W$, defined as $||W||_b=\sum_{i,j}\vert W_{ij} \vert$ is used as the objective function to get a good approximation of the input data.

\subsection{Our contribution}
In our work, we implement and modify \texttt{TMF} so that it can be applied in data mining tasks. We propose a sparse version of \texttt{TMF}, which can work with missing values. 
 
In \textit{Sparse Tropical Matrix Factorization} (\texttt{STMF}), which is available on \url{https://github.com/Ejmric/STMF}, we update the factor matrices $U$ and $V$ based on the  selected given entry of the input data matrix $R$ to predict the missing values in $R$.
In Algorithm~\ref{stmf}, we present the pseudocode of \texttt{STMF} in which for each given entry $(i,j)$ of $R$ we first update $U$ and $V$ based on the element from the $i$\textsuperscript{th} row of the left factor $U$ (\texttt{ULF}, see Algorithm~\ref{ULF}). If the update of the factors does not improve the approximation of $R$, then we update $U$ and $V$ based on the element from the $j$\textsuperscript{th} column of the right factor $V$
(\texttt{URF}, see Algorithm~\ref{URF}). 

\begin{algorithm}[H]
\caption{Sparse tropical matrix factorization (\texttt{STMF})}
\begin{algorithmic}
\REQUIRE data matrix $R$ $\in \mathbb{R}^{m \times n}$, factorization rank $r$ 
\ENSURE factorization $U$ $\in \mathbb{R}^{m \times r}$, $V$ $\in \mathbb{R}^{r \times n}$
\STATE $perm \gets$ order columns in $R$ by the minimum value in the increasing order
\STATE $R \gets R[:, perm]$
\STATE initialize $U$ 
        and compute $V$ 
\WHILE {not converged}
\STATE \textbf{for} each \textit{given entry} $(i,j)$ of $R$
\begin{ALC@g}
\STATE $(U', V', decreased) \gets \text{ULF}(R, i, j, r, U, V)$
\STATE \textbf{if} $decreased$ is True 
\begin{ALC@g}
   \STATE \textbf{break}
\end{ALC@g}
\STATE $(U', V', decreased) \gets \text{URF}(R, i, j, r, U, V)$\\
\STATE \textbf{if} $decreased$ is True
\begin{ALC@g}
   \STATE \textbf{break}
\end{ALC@g}
\end{ALC@g}
\STATE \textbf{if} $\left\lVert R - U' \otimes V' \right\rVert_b$ decreases
\begin{ALC@g}
\STATE $(U, V) \gets (U', V')$
\end{ALC@g}
\STATE \textbf{else exit} (no solution found)
\ENDWHILE
\RETURN $U, V[:, perm^{-1}]$
\end{algorithmic}
\label{stmf}
\end{algorithm}

Algorithms \texttt{ULF} and \texttt{URF} differ from the corresponding \texttt{TMF}'s versions in the way  they solve linear systems. Since some of the entries of matrix $A$ are not given, we define $(\min, +)$ matrix multiplication $\otimes^{*}$ as
\begin{equation*}
    (A \otimes^{*} B)_{ij} = \min\limits_{A_{ik}, B_{kj} \text{are given}}\{A_{ik} + B_{kj}\}
\end{equation*}
for matrices $A \in \mathbb{R}_{\max}^{m \times p}$ and $B \in \mathbb{R}_{\max}^{p \times n}$, $i=1,\ldots,m$, $j=1,\ldots,n$.
Newly-defined operator $\otimes^*$ can be seen as a generalization of Equation~\eqref{greatest_subsolution}, and it is used for solving linear systems by skipping unknown values.
We assume that at least one element in each row/column is known.

\begin{algorithm}[H]
\caption{Update $U, V$ based on the element from $i$\textsuperscript{th} row of left factor $U$ (\texttt{ULF})}
\begin{algorithmic}
\REQUIRE data matrix $R$  $\in \mathbb{R}^{m \times n}$, position  $(i,j)$, factorization rank $r$, left factor matrix $U$  $\in \mathbb{R}^{m \times r}$, right factor matrix $V$ $\in \mathbb{R}^{r \times n}$
\ENSURE $U', V', decreased$ \\
$decreased = $ False
\FOR{$k\gets 1, r$}
   \STATE $U' \leftarrow U$
   \STATE $U'_{ik} \leftarrow R_{ij} - V_{kj}$
   \STATE $V' \leftarrow (-U')^{T} \otimes^{*}  R$
   \STATE $U' \leftarrow R \otimes^{*} (-V')^{T}$
   \STATE \textbf{if} $\left\lVert R - U' \otimes V' \right\rVert_b$ decreases
   \begin{ALC@g}
   \STATE $decreased \leftarrow $ True
   \STATE \textbf{break}
   \end{ALC@g}
\ENDFOR
\RETURN $U', V', decreased$
\end{algorithmic}
\label{ULF}
\end{algorithm}

\begin{algorithm}[H]
\caption{Update $U, V$ based on the element from $j$\textsuperscript{th} column of right factor $V$ (\texttt{URF})}
\begin{algorithmic}
\REQUIRE data matrix $R$  $\in \mathbb{R}^{m \times n}$, position $(i,j)$, factorization rank $r$, left factor matrix $U$ $\in \mathbb{R}^{m \times r}$, right factor matrix $V$ $\in \mathbb{R}^{r \times n}$
\ENSURE $U', V', decreased$ \\
$decreased = $ False
\FOR{$k\gets 1, r$}
   \STATE $V' \leftarrow V$
   \STATE $V'_{kj} \leftarrow R_{ij} - U_{ik}$
   \STATE $U' \leftarrow R \otimes^{*} (-V')^{T}$
   \STATE $V' \leftarrow (-U')^{T} \otimes^{*} R$
   \STATE \textbf{if} $\left\lVert R - U' \otimes V' \right\rVert_b$ decreases
   \begin{ALC@g}
   \STATE $decreased \leftarrow $ True
   \STATE \textbf{break}
   \end{ALC@g}
\ENDFOR
\RETURN $U', V', decreased$
\end{algorithmic}
\label{URF}
\end{algorithm}

Among the different matrix initialization strategies, we obtained the best performance with Random Acol strategy~\cite{randomacol, datafusion}. Random Acol computes each column of the initialized matrix $U$ as an element-wise average of a random subset of columns of the data matrix $R$. It is a widely used method for initializations in matrix factorization methods since it gives better insight into the original data matrix than simple random initialization.

In contrast to \texttt{Cancer}, where convergence is not guaranteed, the update rules of \texttt{STMF}, similar to \texttt{TMF}, gradually reduce the approximation error. This is ensured by the fact that factor matrices $U$ and $V$ are only updated in the case when   $\left\lVert R - U \otimes V \right\rVert_b$ monotonously decreases. 

\subsection{Distance correlation}
It is well known that Pearson and Spearman correlation coefficients can misrepresent non-linear relationships~\cite{corrs_comparison}.
Since in real data, we often deal with non-linearity, our choice is to use so-called \textit{distance correlation}. Distance correlation~\cite{dcor} is a straightforward measure of association that uses the distances between observations as part of its calculation. It is a better alternative for detecting a wide range of relationships between variables.

Let $X$ and $Y$ be the matrices each with $n$ rows and $A$ and $B$ their matrices of Euclidean distances with the row/column means subtracted, and grand mean added. After matrix centering the \emph{distance covariance} $V_{xy}$ is defined as 
\begin{equation*}
V_{XY}^2 = \frac{1}{n^2}\sum_{i,j=1}^{n} A_{ij} B_{ij},
\end{equation*}{}
and distance correlation ${\rm dcor}$ as
\begin{equation*}
    {\rm dcor}(X,Y) = \sqrt{\frac{V_{XY}^2}{V_X V_Y}},
\end{equation*}
where $V_X$ and $V_Y$ represent distance variances of matrices $X$ and $Y$.
Distance correlation is $0$ only if the two corresponding variables are independent.

Distance correlation cannot be used to compare specific rows between $X$ and $Y$, because it requires the entire matrix to be centered first. In such cases we use Euclidean norm between rows of centered original and rows of centered approximated data.

\subsection{Synthetic data}
We create two types of synthetic datasets of rank 3: one smaller of size $200 \times 100$ and five larger of size $500 \times 300$. We use the $(\max, +)$ multiplication of two random non-negative matrices sampled from a uniform distribution over $[0, 1)$ to generate each synthetic dataset.

\subsection{Real data}
We download the preprocessed TCGA data~\cite{survey} for nine cancer types, where for each cancer type three types of omic data are present: gene expression, methylation and miRNA data. We transpose the data sources, so that in each data source, the rows represent patients and columns represent features. 
The first step of data preprocessing is to take the subset of patients for which we have all three data sources. In our experiments we use only gene expression data.
After filtering the patients, we substitute each gene expression value $x$ in the original data with the $\log_2(x+1)$.  With log-transformation, we make the gene expression data conform more closely to the normal distribution, and by adding one, we reduce the bias of zeros. We also perform polo clustering, which is an optimal linear leaf ordering~\cite{polo}, to re-order rows and columns on the preprocessed data matrix. Polo clustering results in a more interpretative visualization of factor matrices.

Next, we use feature agglomeration to merge similar genes by performing clustering~\cite{murtagh2014ward}. We use Ward linkage and split genes into 100 clusters (see Supplementary Figure S~\ref{datasets_agglom}), the center of each cluster representing a \emph{meta-gene}. With this approach, we minimize the influence of non-informative, low variance genes on distance calculations and reduce the computational requirements. 

\begin{table}[h!]
\centering
\caption{Size of gene expression data in the form of $patients \times meta\text{-}genes$ for eight cancer subtypes, and for the subset of \texttt{PAM50} genes in \texttt{BIC}.}
      \begin{tabular}{cc}
        \hline
           \texttt{cancer subtype} & \texttt{size} \\ \hline
        Acute Myeloid Leukemia (\texttt{AML}) & $171 \times 100$ \\
        Colon Adenocarcinoma (\texttt{COLON}) & $221 \times 100$ \\
        Glioblastoma Multiforme (\texttt{GBM}) & $274 \times 100$ \\
        Liver Hepatocellular Carcinoma (\texttt{LIHC}) & $410 \times 100$ \\
        Lung Squamous Cell Carcinoma (\texttt{LUSC}) & $344 \times 100$ \\
        Ovarian serous cystadenocarcinoma (\texttt{OV}) & $291 \times 100$ \\
        Skim Cutaneous Melanoma (\texttt{SKCM}) & $450 \times 100$ \\
        Sarcoma (\texttt{SARC}) & $261 \times 100$ \\
        Breast Invasive Carcinoma (\texttt{BIC}) & $541 \times 50$ \\
        \hline 
        \label{datasets}
      \end{tabular}
\end{table}

For Breast Invasive Carcinoma (\texttt{BIC}), we do not perform feature agglomeration since a list of 50 genes, called \texttt{PAM50}~\cite{parker2009supervised}, classify breast cancers into one of five subtypes: LumA, LumB, Basal, Her2, and Normal~\cite{cerami, gao}, resulting in our \texttt{BIC} data matrix of size $541 \times 50$. These five subtypes differ significantly in a couple of genes in our data, which leads to the value close to zero for silhouette score~\cite{rousseeuw1987silhouettes} (see Supplementary Figure S~\ref{bic_silh_plot}). The sizes of the final nine datasets are listed in Table~\ref{datasets}.

\subsection{Performance evaluation}
Experiments were performed for varying values of the factorization rank. 
The smaller synthetic dataset experiments were run 10 times, with 500 iterations each, and on larger synthetic datasets, experiments were run 50 times, with 500 iterations each.
Experiments for real data were run five times, with 500 iterations.
For both datasets, we mask $20 \%$ of data as missing, which we then use as a test set to evaluate the two tested methods. 
The remaining $80 \%$ represent the training set.
We choose a rank based on the approximation error on training data, which represents a fair/optimal choice for both methods, \texttt{STMF} and \texttt{NMF} so that we can compare them, knowing both of them to have the same number of parameters. 

We compute the distance correlation and Euclidean norm between the original and approximated data matrix to evaluate the predictive performance. 

\section{Results}
First, we use synthetic data to show the correctness of the \texttt{STMF} algorithm. We use the smaller dataset to show that \texttt{STMF} can discover the tropical structure. The larger datasets are needed to show how the order of rows and columns affects the result. We then apply it to real data to compare the performance and interpretability of models obtained with \texttt{STMF} and \texttt{NMF}. 

\subsection*{Synthetic data}
\begin{figure}[h!]
        \centering
        \includegraphics[scale=0.5]{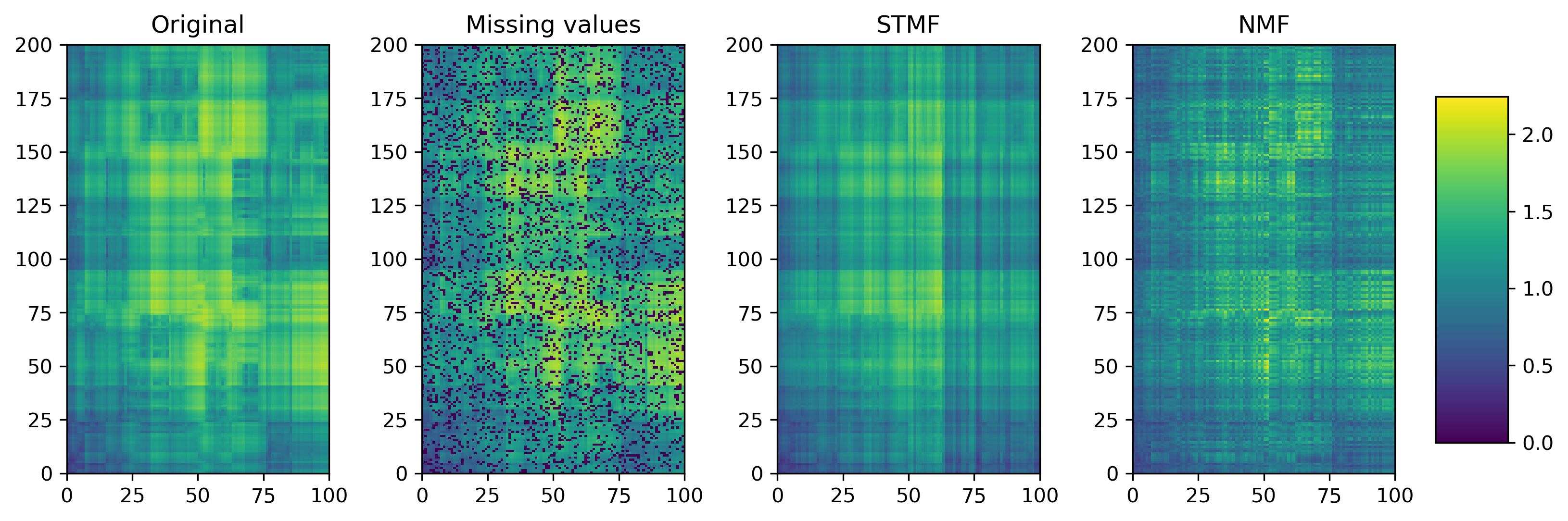}
        \caption{A comparison between \texttt{STMF}'s and \texttt{NMF}'s predictions of best rank 4 approximations on $200 \times 100$  synthetic $(\max,+)$ matrix with $20\%$ missing values.}
        \label{synth}
\end{figure}

\begin{figure}[h!]
\centering
\begin{subfigure}{\textwidth}
\centering
\includegraphics[width=0.8\textwidth]{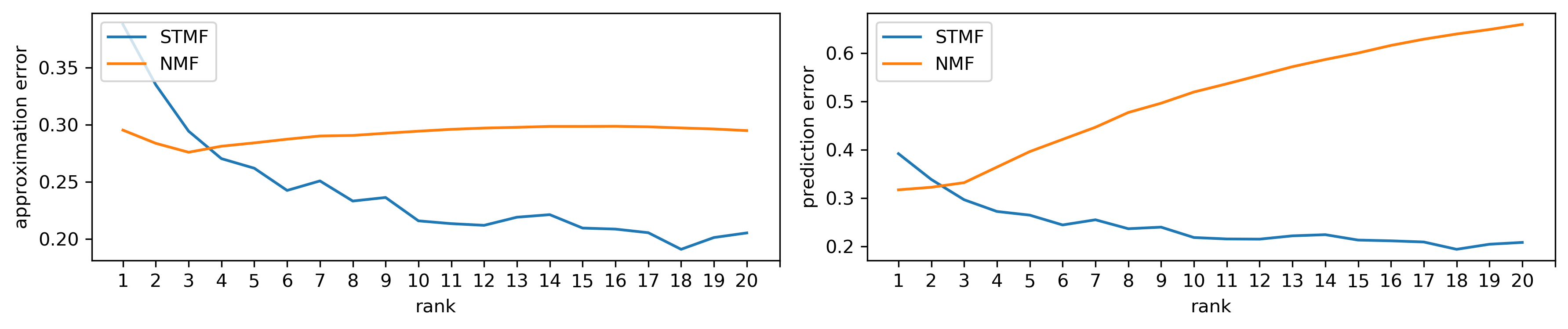}
\caption{Median RMSE for approximation (training) and prediction (test) data.}
\end{subfigure}

\begin{subfigure}{\textwidth}
\centering
\includegraphics[width=0.4\textwidth]{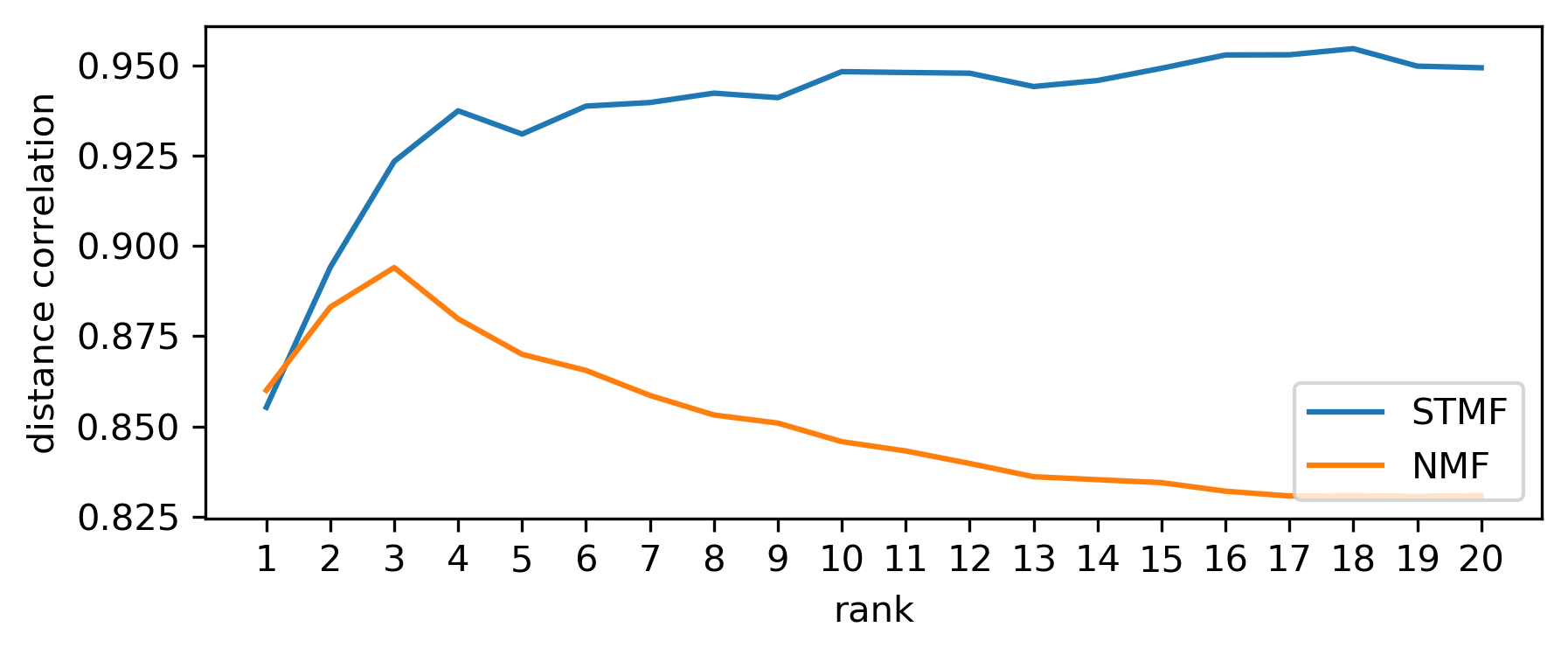}
\caption{Median distance correlation.}
\end{subfigure}

\caption{Comparison of \texttt{STMF} (blue) and \texttt{NMF}  (orange) on synthetic $(\max,+)$ matrix of size $200 \times 100$ and rank 3.}
\label{error_and_corr}
\end{figure}

The objective of synthetic experiments is to show that \texttt{STMF} can identify the $(\max,+)$ structure when it exists. 
Even on a relatively small $200 \times 100$ matrix results show that \texttt{NMF} cannot successfully recover extreme values compared to \texttt{STMF}, see Figure~\ref{synth}. As the results show \texttt{STMF} achieves a smaller prediction root-mean-square-error (RMSE) and higher distance correlation (Figure~\ref{error_and_corr}).

Experiments on synthetic data show that changing the execution order of \texttt{URF} and \texttt{ULF} in the computation of \texttt{STMF} does not affect the result of the algorithm. 

\begin{figure}[h!]
        \centering
        \includegraphics[scale=0.4]{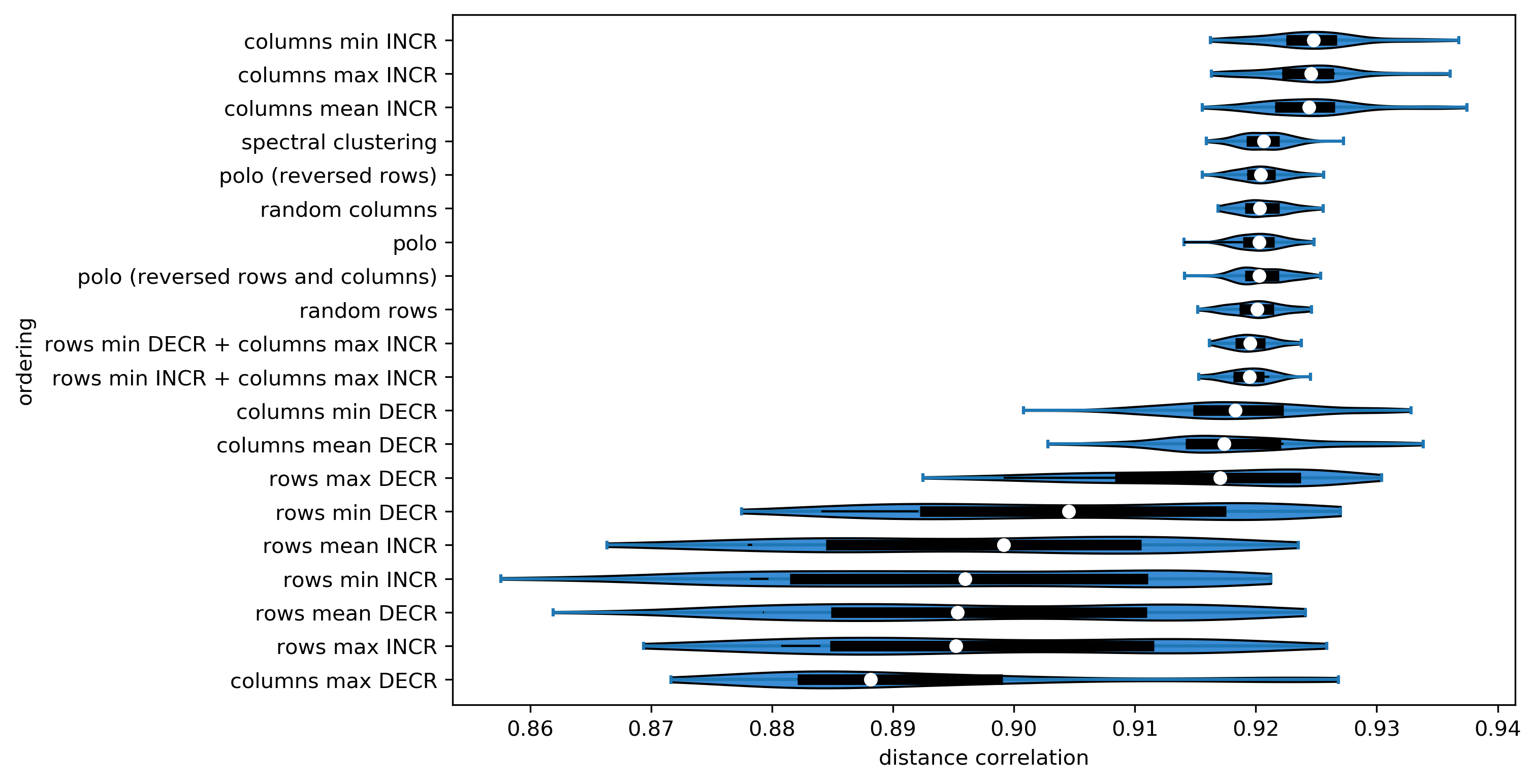}
        \caption{Effect of ordering strategy on achieved distance correlation by \texttt{STMF}, on $500 \times 300$  synthetic  $(\max, +)$ matrix. Top three performing strategies order columns by increasing values.}
        \label{violin_plots}
\end{figure}

The result of \texttt{STMF} depends on the order of matrix entries. We perform different types of permutation techniques to order columns and rows on five large synthetic datasets (see Supplementary Figure S~\ref{synth_permutation}). 
Top three strategies are to sort columns by increasing values of their minimum, maximum, and mean value (Figure~\ref{violin_plots}). Moreover, in four out of five datasets, the best results were obtained by  ordering columns in increasing order by their minimum value (see Supplementary Figure S~\ref{synth_violin_plots}). This strategy represents the first step of \texttt{STMF} method (Algorithm~\ref{stmf}).

\subsection{Real data}
\begin{figure}[h!]
        \centering
        \includegraphics[scale=0.5]{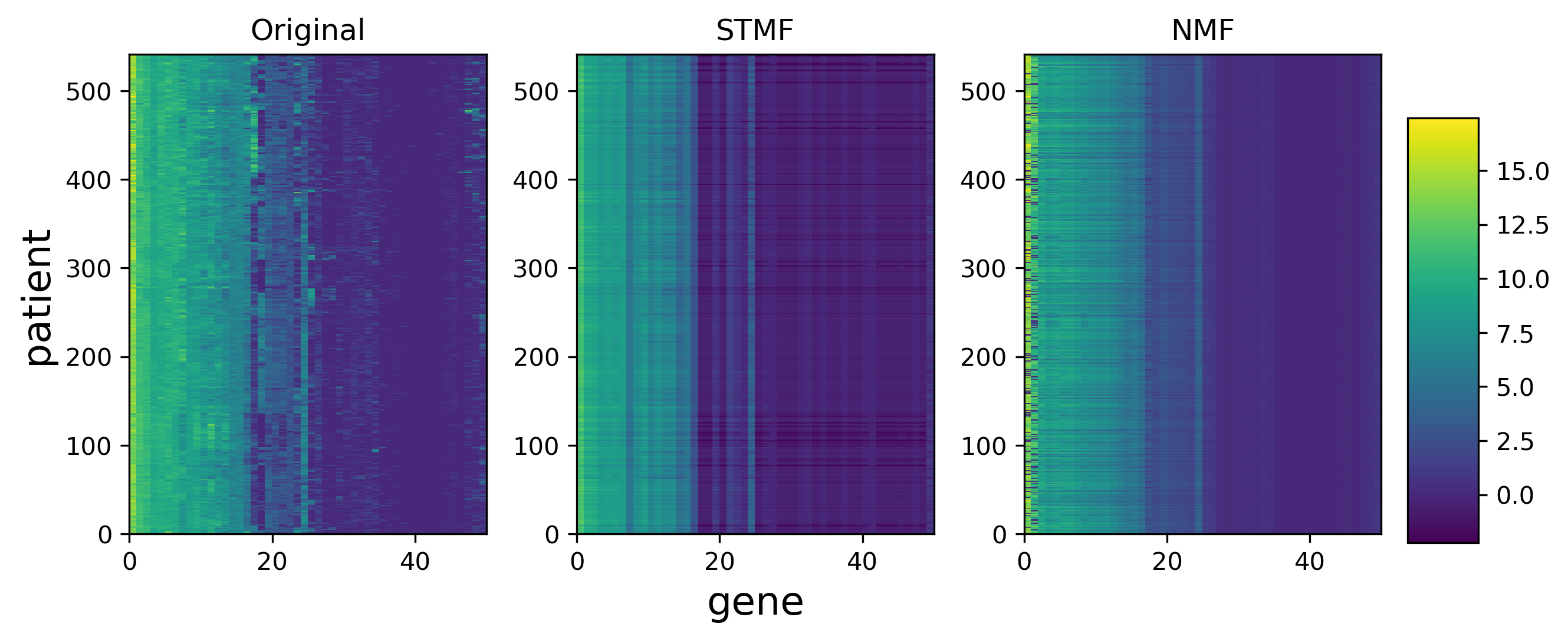}
        \caption{Best rank 3 approximation matrices $R_{\texttt{STMF}}$ and $R_{\texttt{NMF}}$ from \texttt{STMF} and \texttt{NMF} on the prediction  of the gene expression signal on Breast Invasive Carcinoma (\texttt{BIC}) tumor.}
        \label{bic}
\end{figure}

Figure~\ref{bic} shows the results on \texttt{BIC} matrix, with \texttt{PAM50} genes and $541$ patients. Our findings confirm that \texttt{STMF} expresses some extreme values. We see that \texttt{STMF} successfully recovers large values, while \texttt{NMF} has the largest error where gene expression values are high.
Note that \texttt{NMF} tends towards the mean value. Half of the original data is close to zero (plotted in dark blue), which is a reason that \texttt{NMF} cannot successfully predict high (yellow) values.
For all other datasets approximation matrices are available in Supplementary, Section~\ref{supp_section_two}.

\begin{figure}[htb] 
    \centering
    \begin{subfigure}{\textwidth}
    \centering
    \includegraphics[width=0.8\textwidth]{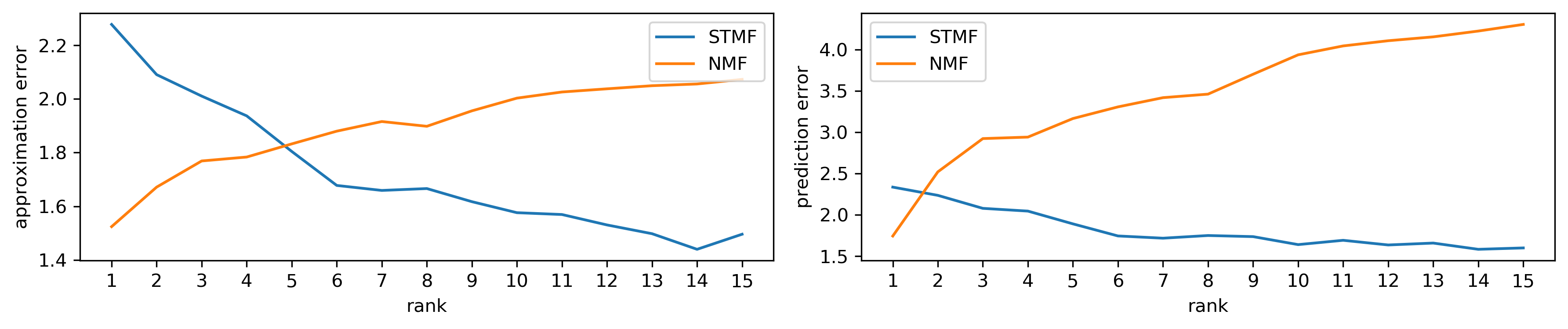}
    \caption{Median RMSE for approximation (training) data and prediction (test) data.}
    \label{metrics_rmse}
\end{subfigure}\hfill%

\begin{subfigure}{\textwidth}
    \centering
    \includegraphics[width=0.8\textwidth]{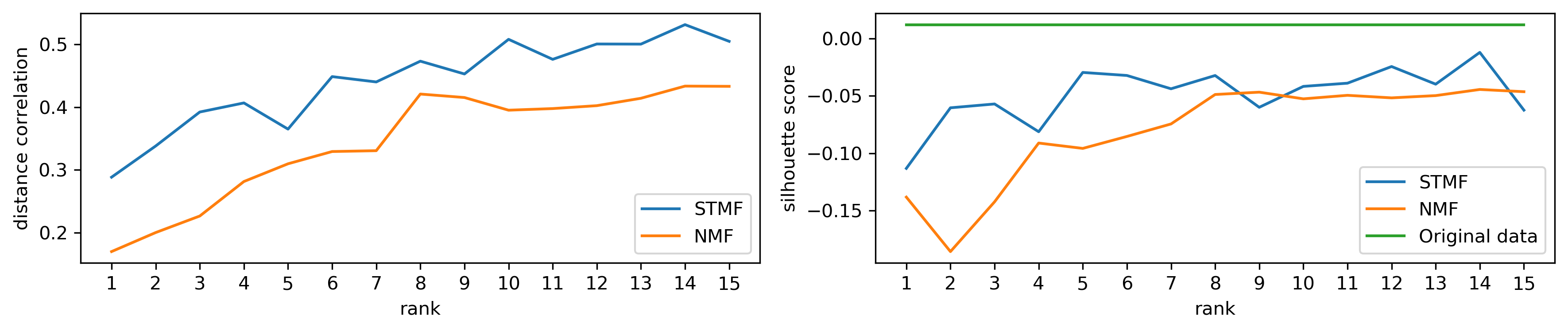}
    \caption{Median distance correlation and median silhouette score.}
    \label{metrics_corr_silh}
\end{subfigure}

    \caption{Comparison of performance  of  \texttt{STMF} (blue) and \texttt{NMF} (orange) on \texttt{BIC} matrix.}
    \label{metrics}
\end{figure}

In Figure~\ref{metrics_rmse} we see that \texttt{NMF} has smaller approximation error than \texttt{STMF}, but larger prediction error. So, \texttt{NMF} better approximates/fits the data, but \texttt{STMF} is not prone to overfitting, since its prediction error is smaller.
On the other hand, in Figure~\ref{metrics_corr_silh}, \texttt{STMF} has better distance correlation and silhouette score values. This means that \texttt{STMF} can find clusters of patients with the same subtype better than \texttt{NMF}, which tends to describe every patient to be similar to the average one.
For all other datasets similar graphs are available in Supplementary, Section~\ref{supp_section_two}.

\begin{figure}[h!] 
    \centering
    \begin{subfigure}{.5\textwidth}
    \includegraphics[width=\textwidth]{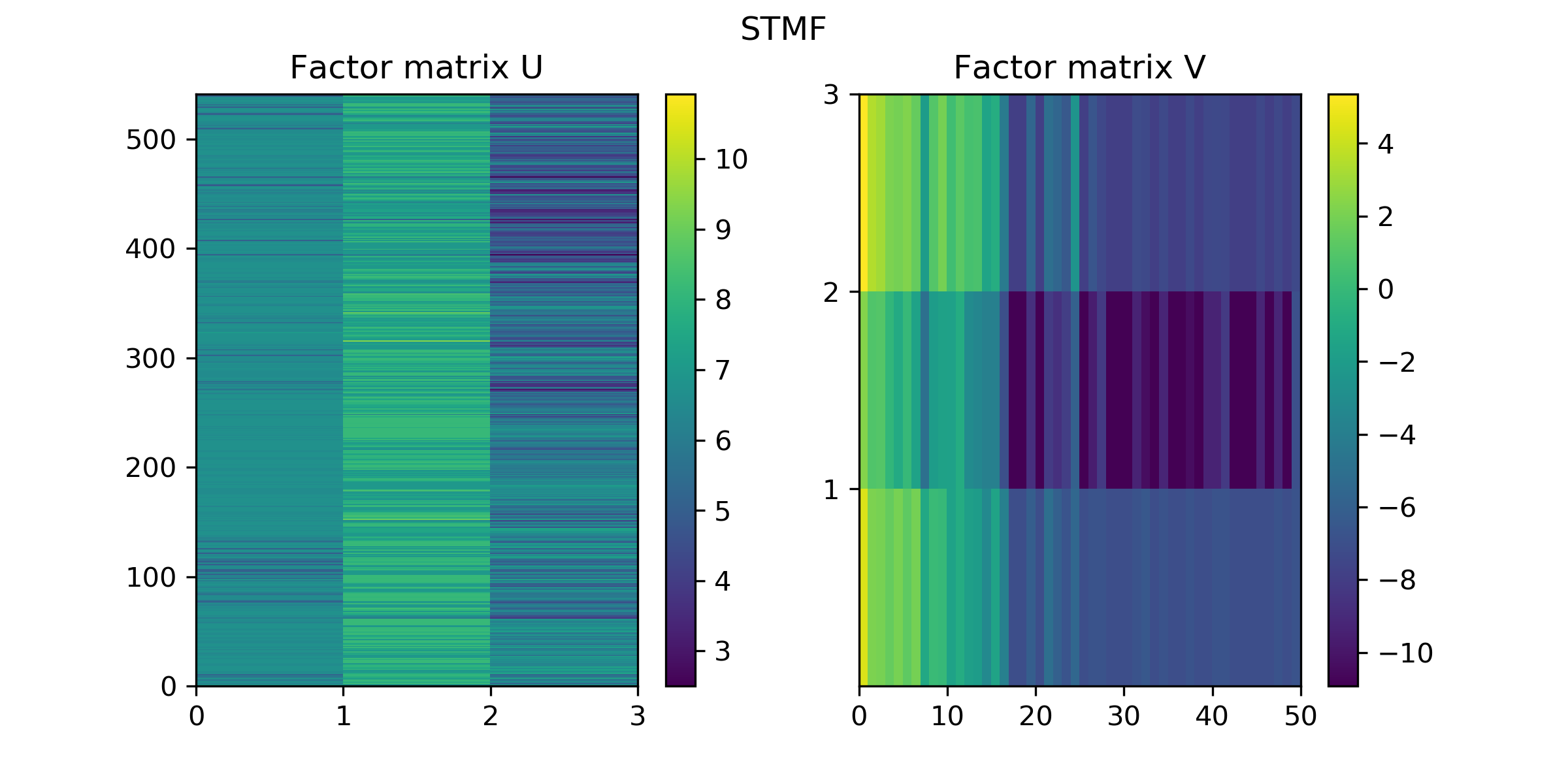}
    \caption{Factor matrices $U_{\texttt{STMF}}, V_{\texttt{STMF}}$ from \texttt{STMF}.}
    \label{u_v_stmf}
\end{subfigure}\hfill%
\begin{subfigure}{.5\textwidth}
    \includegraphics[width=\textwidth]{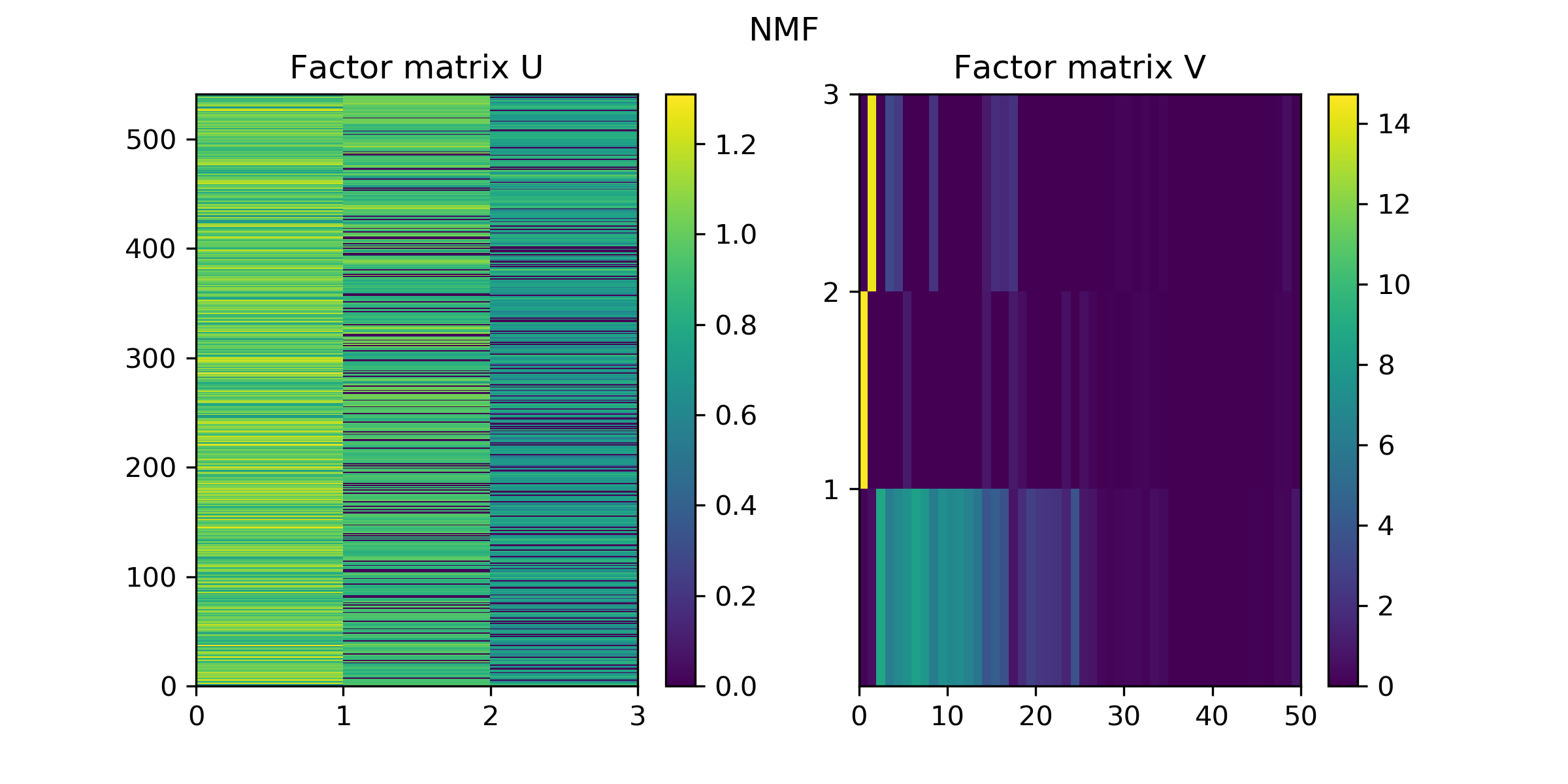}
    \caption{Factor matrices $U_{\texttt{NMF}}, V_{\texttt{NMF}}$ from \texttt{NMF}.}
    \label{u_v_nmf}
\end{subfigure}
    \caption{Factor matrices $U_{\texttt{STMF}}, V_{\texttt{STMF}}$ and  $U_{\texttt{NMF}}, V_{\texttt{NMF}}$ obtained by \texttt{STMF} and \texttt{NMF} algorithms,
    respectively.}
    \label{factor_matrices}
\end{figure}

We factorize \texttt{BIC} matrix, illustrated in Figure~\ref{bic}. In Figure~\ref{factor_matrices} we show \texttt{STMF}'s factor matrices denoted by $U_{\texttt{STMF}}, V_{\texttt{STMF}}$, and \texttt{NMF}'s factor matrices denoted by $U_{\texttt{NMF}}, V_{\texttt{NMF}}$. We see that these factor matrices are substantially different.
Basis factor $V_{\texttt{STMF}}$ (first and third row) is visually the most similar to the original matrix than any other factor alone. Factor $V_{\texttt{STMF}}$ detects low and high values of gene expression, while factor $V_{\texttt{NMF}}$ detects high values in the first two columns (second and third row, respectively) and low values in remaining columns (first row). Coefficient factors $U_{\texttt{STMF}}$ and $U_{\texttt{NMF}}$ contribute to a good approximation of the original matrix.
For all other datasets factor matrices are available in Supplementary, Section~\ref{supp_section_two}.

\begin{figure}[h!] 
    \centering
    \begin{subfigure}{.48\textwidth}
    \centering
    \includegraphics[width=\textwidth]{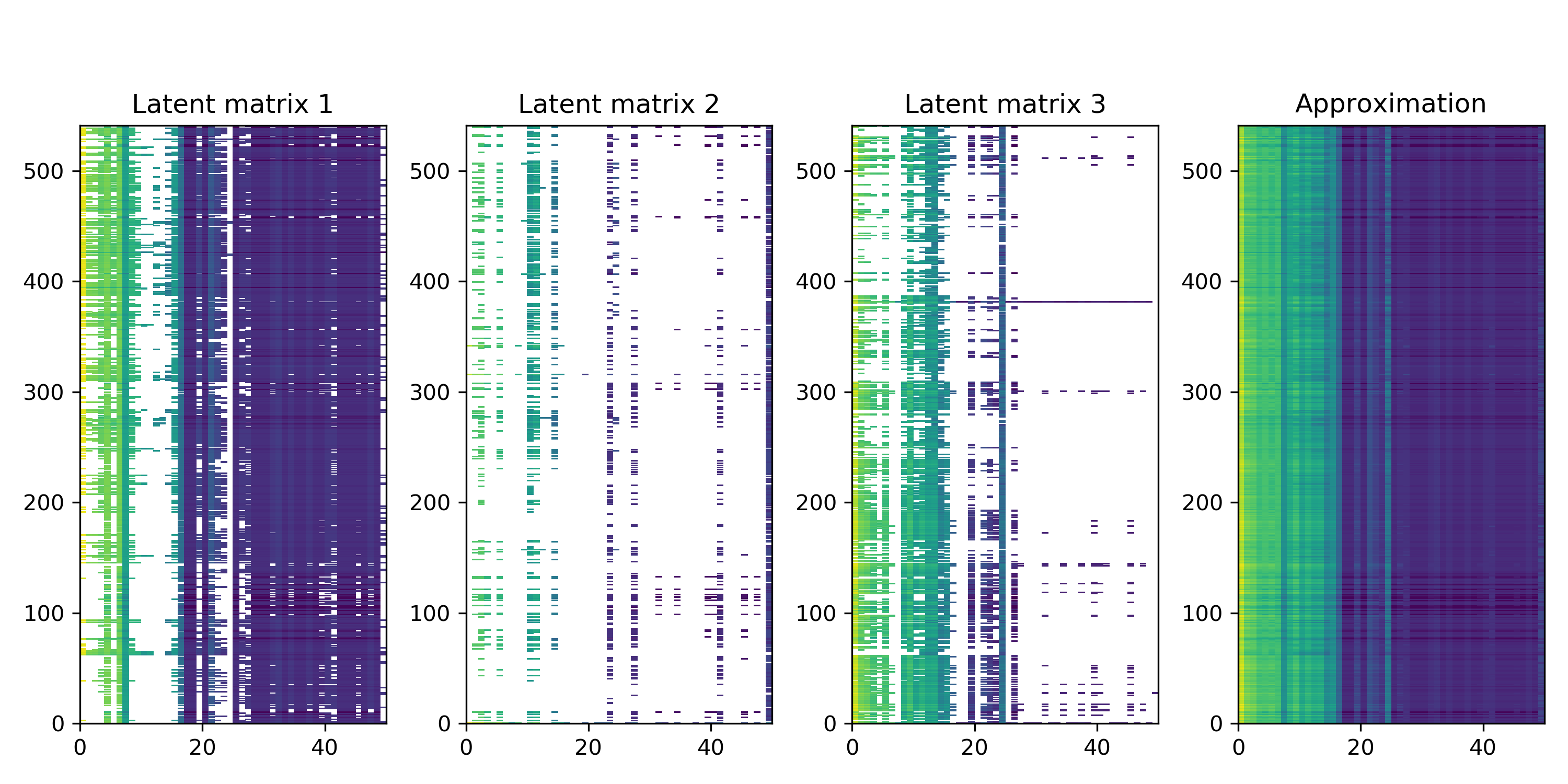}
    \caption{Latent matrices $R_{\texttt{STMF}}^{(i)}$, $i \in \{1,2, 3\}$, where white color represents the elements which do not contribute to the approximation $R_{\texttt{STMF}}$.}
    \label{latent_stmf}
\end{subfigure}\hfill%
\vfill%
\begin{subfigure}{.48\textwidth}
\centering
    \includegraphics[width=\textwidth]{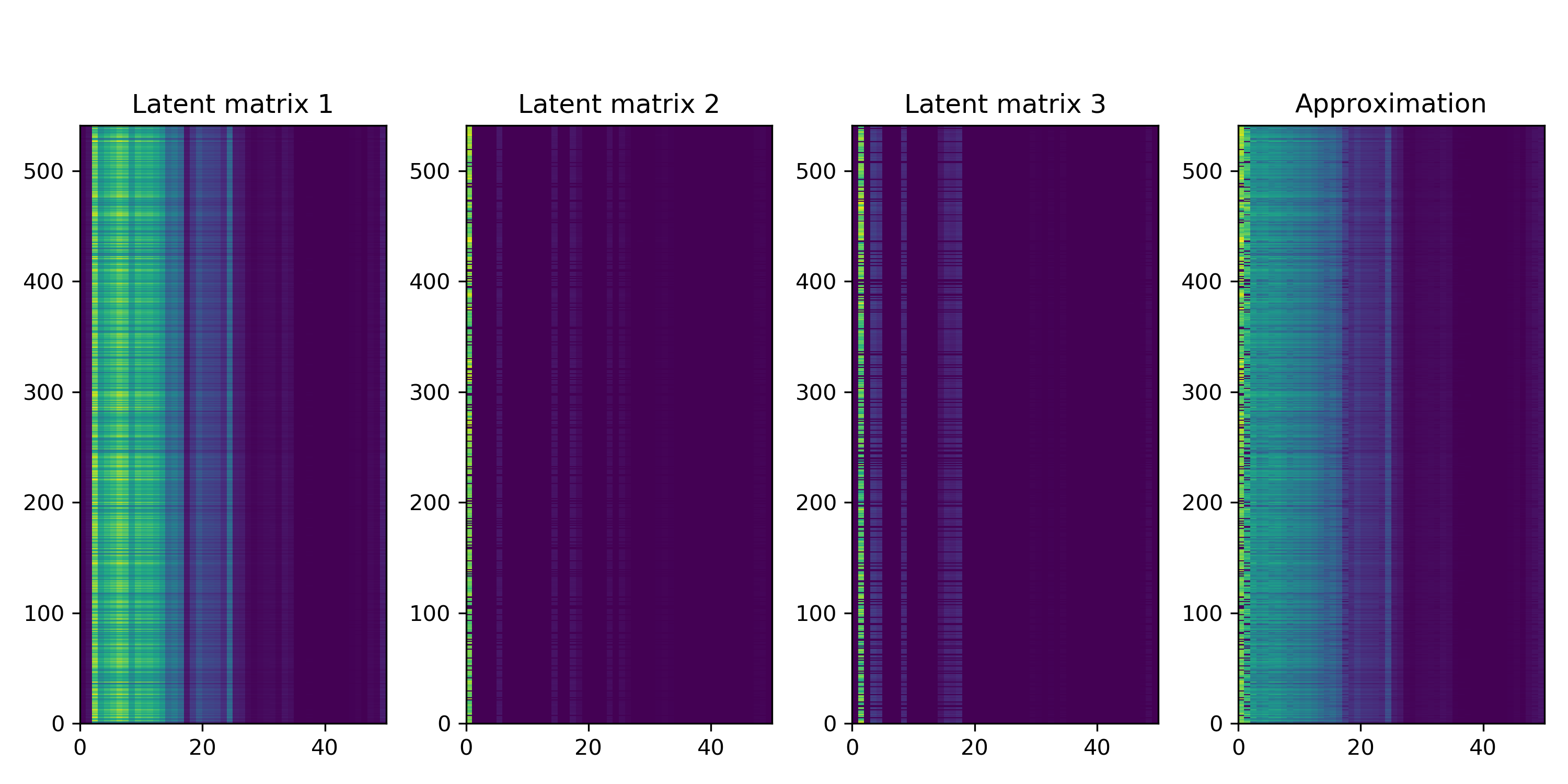}
    \caption{Latent matrices $R_{\texttt{NMF}}^{(i)}$, $i \in \{1,2, 3\}$.}
    \label{latent_nmf}
\end{subfigure}
    \caption{\texttt{STMF}'s and \texttt{NMF}'s latent matrices.}
    \label{latent_matrices}
\end{figure}

To see which part of data is explained by which factorization rank, we define a latent matrix $R^{(i)}$ as a reconstruction using only one latent component from the approximation matrix, where $i \in \{1,\ldots, r\}$, and $r$ is the factorization rank. $R^{(i)}$ can be seen as a projection on the direction of the $i$-th factor. For example, $R_{\texttt{STMF}}^{(1)}$ matrix in Figure~\ref{latent_stmf} is a result of the (max, +) product, which represent sums of each pair of elements, of the first column of $U_{\texttt{STMF}}$ and the first row of $V_{\texttt{STMF}}$ (Figure~\ref{factor_matrices}). In the case of \texttt{NMF}, instead of sum, there is multiplication (see Figure~\ref{latent_nmf}). If we compute an element-wise maximum of all $R_{\texttt{STMF}}^{(i)}$ we get the $R_{\texttt{STMF}}$, while element-wise sum of all $R_{\texttt{NMF}}^{(i)}$ results in $R_{\texttt{NMF}}$. In this way, we see which latent matrix $R^{(i)}$ explains which part of the data. On the \texttt{BIC} matrix, we see that both methods, \texttt{STMF}, and \texttt{NMF}, describe most of the data with the first latent matrix (Figure~\ref{latent_matrices}).
For all other datasets latent matrices are available in Supplementary, Section~\ref{supp_section_two}.

In Table~\ref{real_table} we present the results of experiments on nine datasets presented in Table~\ref{datasets}. We see that \texttt{STMF} outperforms \texttt{NMF} on six datasets, while \texttt{NMF} achieves better results on the remaining three datasets, \texttt{LUSC}, \texttt{SKCM} and \texttt{SARC}.

Solving linear systems using $\otimes^{*}$ emphasizes the low (blue) and high (yellow) gene expression values of patients in Figure~\ref{bic}. In this way, \texttt{STMF} can, in some cases, recover better the original data, while \texttt{NMF}'s results are diluted. However, a limitation of  \texttt{STMF} compared to  \texttt{NMF} is in its computational efficiency (Table \ref{table_runtime}).

\begin{table}[h!]
\centering
\caption{The comparison of the average running time in seconds with the best choice of rank $r$ for different matrix factorization methods on nine datasets.}
      \begin{tabular}{cccccc}
        \hline
           dataset & rank $r$ & \texttt{STMF} & \texttt{NMF} \\ \hline
        \texttt{AML} & $3$ & 117.953 & 0.336 \\
        \texttt{COLON} & $3$ & 153.398 & 0.312 \\
        \texttt{GBM} & $3$ & 191.204 & 0.353 \\
        \texttt{LIHC} & $2$ & 236.655 & 0.467 \\
        \texttt{LUSC} & $3$ & 239.329 & 0.456 \\
        \texttt{OV} & $4$ & 251.328 & 0.336 \\
        \texttt{SKCM} & $3$ & 310.401 & 0.395 \\
        \texttt{SARC} & $3$ & 186.475 & 0.398 \\
        \texttt{BIC} & $3$ & 221.669 & 0.526 \\
        \hline 
        \label{table_runtime}
      \end{tabular}
\end{table}

\begin{table}[h!]
 \centering
\caption{The comparison of the distance correlation with the best choice of rank $r$ for different matrix factorization methods on nine datasets.}
    \begin{tabular}{cccccc}
    \hline
    & & \multicolumn{3}{c}{\texttt{STMF}} & \multicolumn{1}{c}{\texttt{NMF}} \\
    \cmidrule(r){3-5}
    dataset & rank $r$ & {Min.} & {Median} & {Max.} &  \\ \hline
        \texttt{AML} & $3$ & 0.650 & \textbf{0.831} &  0.845 & 0.636 \\
        \texttt{COLON} & $3$ & 0.585 & \textbf{0.647} & 0.688 & 0.586 \\
        \texttt{GBM} & $3$ & 0.684 & \textbf{0.702} & 0.762 & 0.325 \\
        \texttt{LIHC} & $2$ & 0.493 & \textbf{0.515} & 0.588 & 0.311 \\
        \texttt{LUSC} & $3$ & 0.498 & 0.562 & 0.731  & \textbf{0.697} \\
        \texttt{OV} & $4$ & 0.420 & \textbf{0.569} & 0.601 & 0.347 \\
        \texttt{SKCM} & $3$ & 0.480 & 0.521 & 0.605 &  \textbf{0.633} \\
        \texttt{SARC} & $3$ & 0.493 & 0.584 & 0.610 & \textbf{0.649} \\
        \texttt{BIC} & $3$ & 0.350 & \textbf{0.392} & 0.531 & 0.227 \\
        \hline 
        \label{real_table}
    \end{tabular}
\end{table}

\begin{figure}[h!]
        \centering
        \includegraphics[scale=0.3]{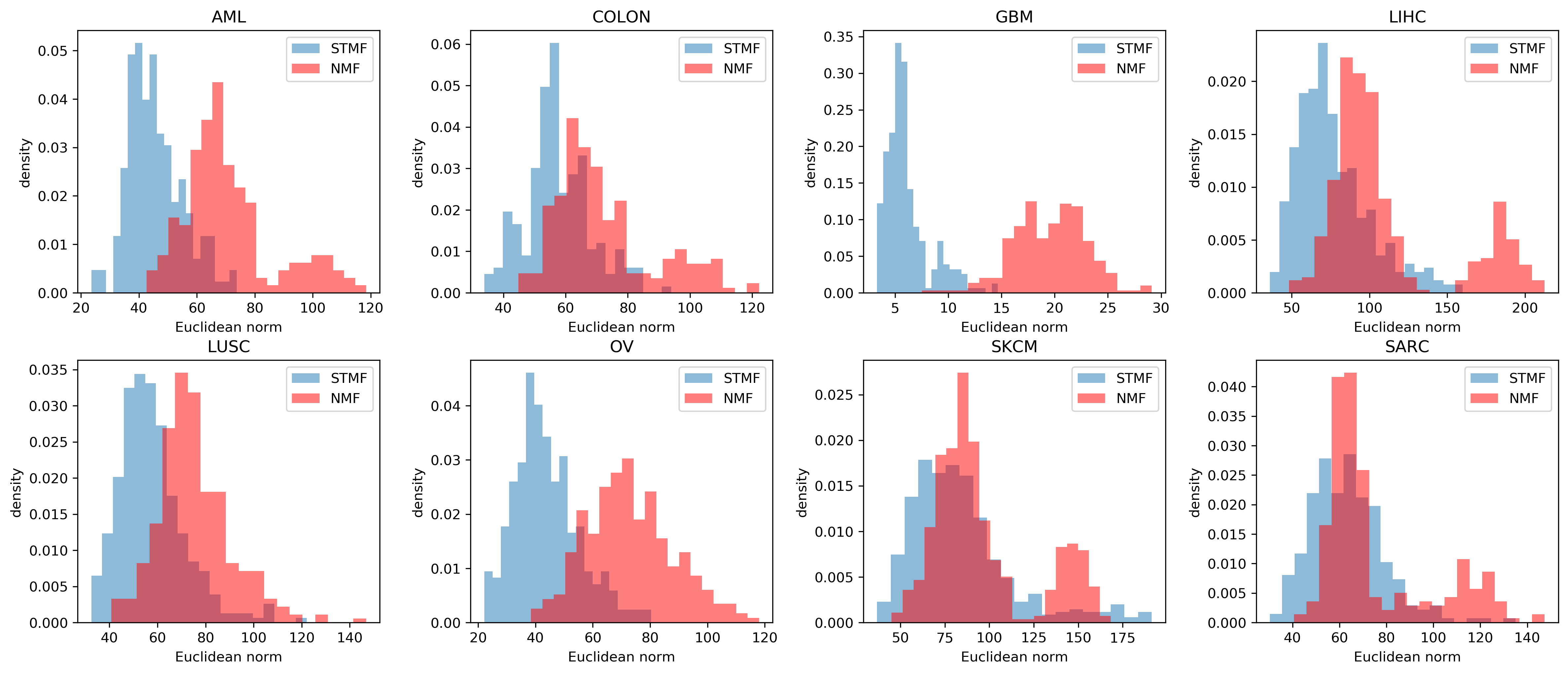}
        \caption{Euclidean norm of difference between centered original data and centered approximations of rank $r$ (chosen in Table \ref{real_table}) for different  datasets.}
        \label{datasets_dist}
\end{figure}

In Figure~\ref{datasets_dist} we plot the distribution of Euclidean norm of difference between centered original data and centered approximations of rank $r$ (chosen in Table \ref{real_table}) for different datasets. We see that even if we use another metric like Euclidean norm, computed for each row (patient) separately, results still show that \texttt{STMF} outperforms \texttt{NMF}, as it is shown in Table \ref{real_table} using distance correlation.

\begin{figure}[h!]
        \centering
        \includegraphics[scale=0.4]{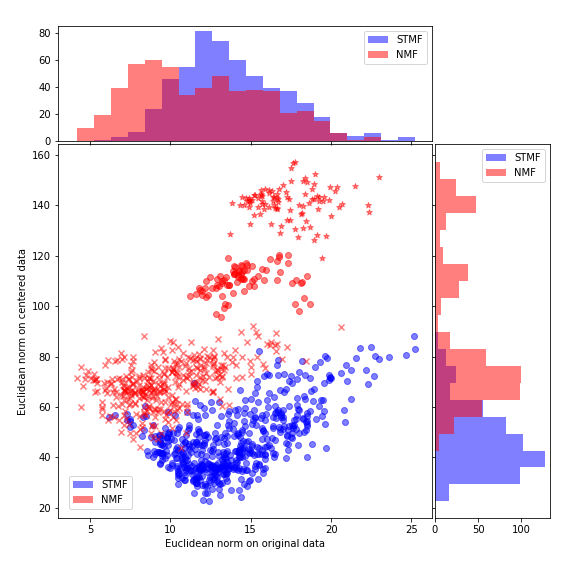}
        \caption{Euclidean norm of difference between original \texttt{BIC} data and approximations ($x$-axis) and Euclidean norm of difference between centered original data and centered approximations ($y$-axis).}
        \label{experiment_euclidean_norm}
\end{figure}

In Figure~\ref{experiment_euclidean_norm} we explore the difference between the original, approximated and centered \texttt{BIC} dataset. For every row (patient) we present the Euclidean norm  of the difference between the rows in the original and the approximated matrix on $x$-axis, which can be interpreted as the accuracy of the approximated values.
In contrast, on $y$-axis we present the Euclidean norm of the difference between the corresponding rows in the centered original and centered approximated matrix, which can be interpreted as the average error of the reconstruction of the original pattern.
We see that for each row (patient) the \texttt{STMF}'s value on $y$-axis is smaller than the \texttt{NMF}'s value, indicating that \texttt{STMF} better approximates the original patterns. The rows in the \texttt{STMF}'s approximation in Figure~\ref{bic} with predominantly low values have large approximation errors ($y$-axis) in Figure~\ref{experiment_euclidean_norm} while having a comparable approximation of the original pattern as \texttt{NMF}'s approximation of the original pattern.

\begin{figure}[h!] 
    \centering
    \begin{subfigure}{.5\textwidth}
    \includegraphics[width=\textwidth]{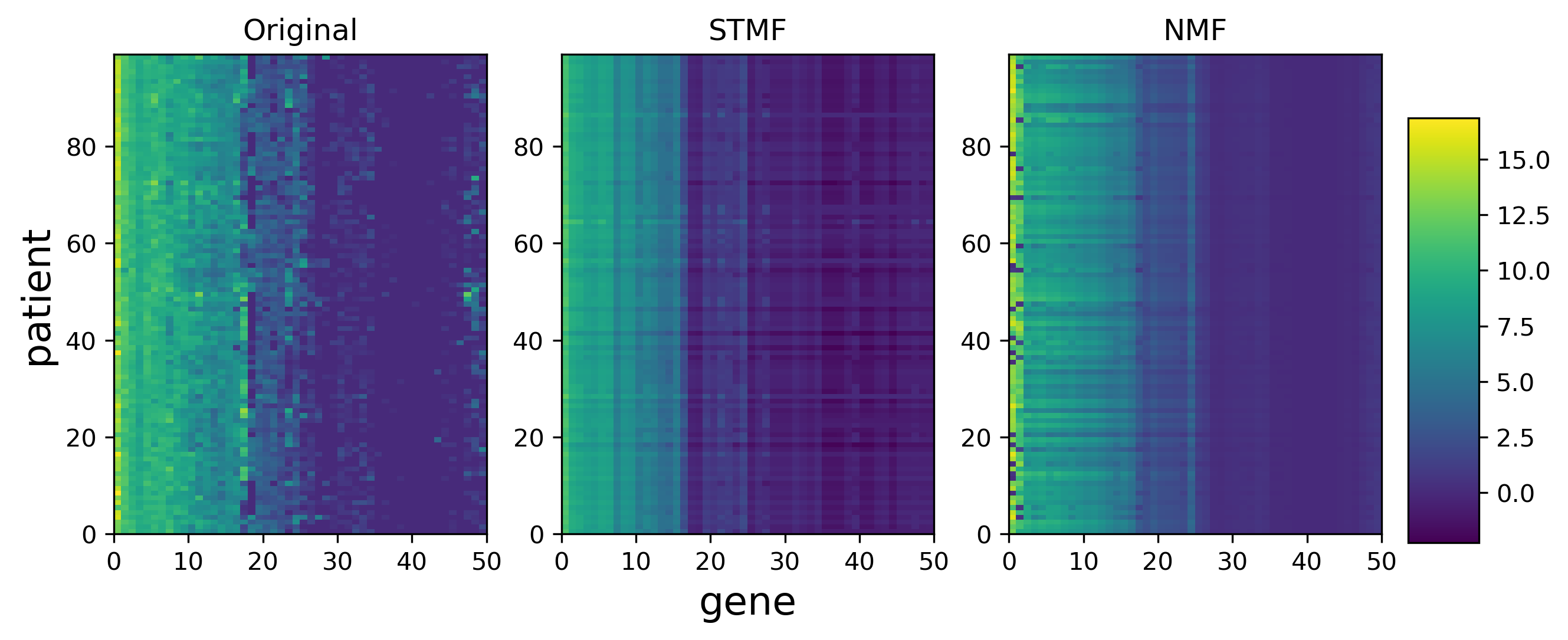}
    \caption{Matrix approximations of the first cluster.}
    \label{first_cluster}
\end{subfigure}\hfill%
\begin{subfigure}{.5\textwidth}
    \includegraphics[width=\textwidth]{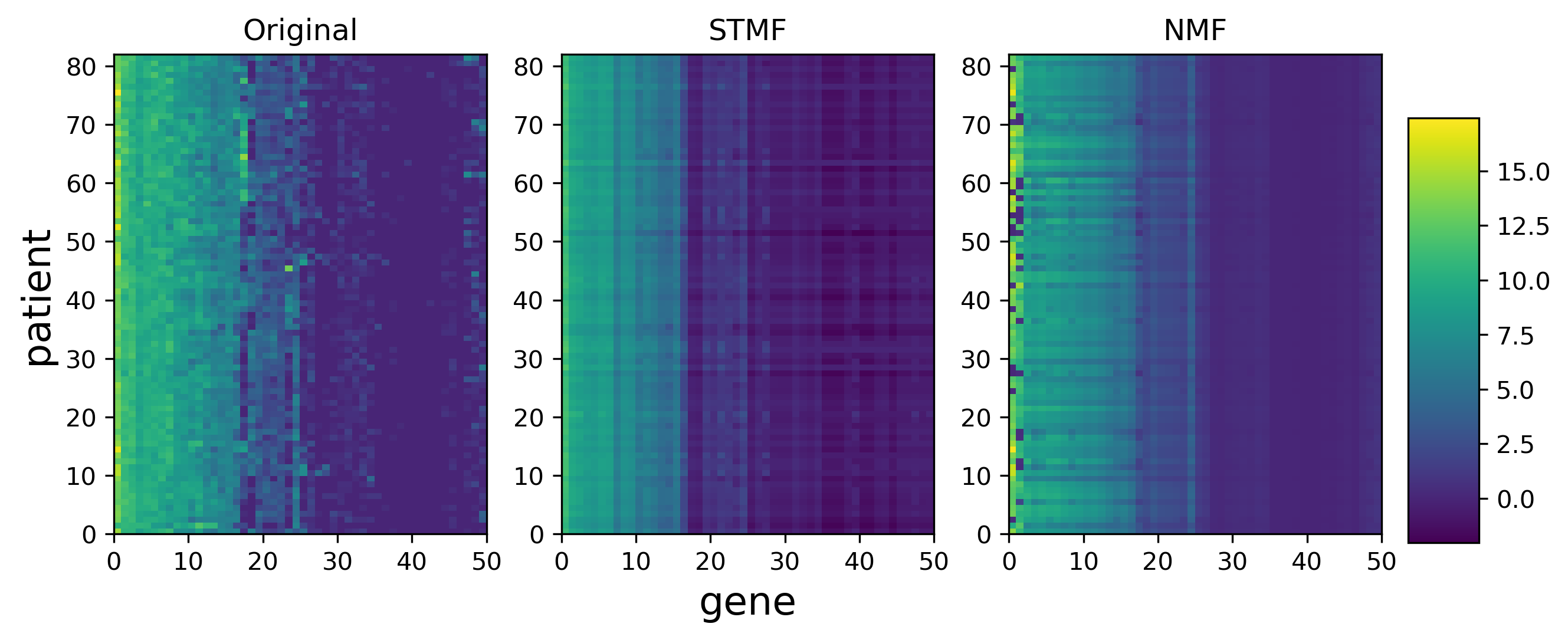}
    \caption{Matrix approximations of the second cluster.}
    \label{second_cluster}
\end{subfigure}
    \caption{Comparison of \texttt{STMF}'s and \texttt{NMF}'s  approximations of specific patients from two  clusters generated by \texttt{NMF}, shown in Figure~\ref{experiment_euclidean_norm}.  First cluster has a center positioned around  $(17, 141)$, while second cluster is positioned around $(15, 109)$.}
    \label{clusters}
\end{figure}

We see that \texttt{NMF} has two clusters of patients with large values on $y$-axis, denoted by red stars and red circles. These are the rows (patients) where the \texttt{NMF}'s predicted pattern differs significantly from the original pattern, more than the \texttt{STMF}'s predictions, but at the same time \texttt{NMF} is achieving smaller approximation error than \texttt{STMF}.
In Figure~\ref{clusters} we plot the patients corresponding to these two clusters and compare approximations with original data. 
It can be seen that \texttt{NMF} cannot model high (yellow) values in a couple of first columns, while low (blue) values are larger (light blue) compared to the original matrix, which has around half of the data plotted with dark blue.
Comparison with Pearson and Spearman correlation is shown in Supplementary Figure S~\ref{bic_pear_spear}, where \texttt{STMF} achieves higher Pearson correlation, but lower Spearman correlation. Clusters of patients are also visible in both figures using these two correlations confirming results in Figure~\ref{experiment_euclidean_norm}. 
For all other datasets plots are available in Supplementary, Section~\ref{supp_section_two}.

\section*{Conclusion}
Standard linear algebra is used in the majority of data mining and machine learning tasks. Utilizing different types of semirings has the potential to reveal previously undiscovered patterns.
The motivation for using tropical semiring in matrix factorization methods is that resulting factors should give us the most dominant features that are specific and essential for each factor. In that way, factors are likely easier to interpret. 

We propose a method called \texttt{STMF}, which can work with missing values. We implement \texttt{STMF} by extending \texttt{TMF} algorithm to be able to handle unknown values.
Results show that \texttt{NMF} could not successfully recover the patterns on specific synthetic data, while the approximation with \texttt{STMF} achieves a higher correlation value. Results on TCGA data show that \texttt{STMF} outperforms \texttt{NMF} in the prediction task. Also, the results obtained by \texttt{NMF} tend toward the mean value, while the approximations obtained by \texttt{STMF} better express extreme values. Our proposed approach identifies strong patterns that aid the visual interpretation of results. In this way, we can discover sharp, high-variance components in the data. To the best of our knowledge, \texttt{STMF} is the first work using tropical semiring in sparse (biomedical) data.

A limitation of our \texttt{STMF} method is the fact that can be used only on single data source. Integrative data fusion methods are based on co-factorization of multiple data matrices. Using standard linear algebra, \texttt{DFMF} is a variant of penalized matrix tri-fa\-cto\-ri\-za\-tion, which simultaneously factorizes data matrices to reveal hidden associations. It can model multiple relations between multiple object types, while relations between some object types can be completely missing. 
In our future work, we will investigate ways to modify the \texttt{STMF} method for data fusion of multiple data sources focusing on the fusion of methylation, miRNA, and gene expression data.

We believe that future research will show that semirings are useful in many scenarios and that they find the structure that is different and easier to interpret than with standard linear algebra.

\section*{Author's contributions}
AO, TC and PO designed the study. HK guided the selection and the processing of the data. AO wrote the software application and performed experiments. AO and TC analyzed and interpreted the results on real data. AO wrote the initial draft of the paper and all authors edited and approved the final manuscript.

\section*{Acknowledgements}
We would like to express very great appreciation to Rafsan Ahmed from Antalya Bilim University (Department of Computer Engineering, Machine Learning Laboratory, now at Lund University), who helped us acquire and preprocess the real data.

The results published here are in whole or part based upon data generated by the TCGA Research Network: \url{https://www.cancer.gov/tcga}.

\section*{Funding}
This work is supported by the Slovene Research Agency, Young Researcher Grant (52096) awarded to AO, and research core funding (no~P1-0222 to PO and no~P2-0209 to TC).

\section*{Availability of data and materials}
This paper uses the real TCGA data available on \url{http://acgt.cs.tau.ac.il/multi_omic_benchmark/download.html}. \texttt{PAM50} data can be found on the \url{https://github.com/CSB-IG/pa3bc/tree/master/bioclassifier_R/}. \texttt{BIC} subtypes are collected from \url{https://www.cbioportal.org/}. \texttt{STMF} code, \texttt{PAM50} data and \texttt{BIC} subtypes are available on \url{https://github.com/Ejmric/STMF}.

\section*{Ethics approval and consent to participate}
Not applicable.

\section*{Consent for publication}
Not applicable.

\section*{Competing interests}
The authors declare that they have no competing interests.

\bibliographystyle{unsrt}  

\bibliography{references}

\begin{thebibliography}{10}

\bibitem{koren2009matrix}
Yehuda Koren, Robert Bell, and Chris Volinsky.
\newblock Matrix factorization techniques for recommender systems.
\newblock {\em Computer}, 42(8):30--37, 2009.

\bibitem{xu2003document}
Wei Xu, Xin Liu, and Yihong Gong.
\newblock Document clustering based on non-negative matrix factorization.
\newblock In {\em Proceedings of the 26th annual international ACM SIGIR
  conference on Research and development in informaion retrieval}, pages
  267--273, 2003.

\bibitem{brunet2004metagenes}
Jean-Philippe Brunet, Pablo Tamayo, Todd~R Golub, and Jill~P Mesirov.
\newblock Metagenes and molecular pattern discovery using matrix factorization.
\newblock {\em Proceedings of the national academy of sciences},
  101(12):4164--4169, 2004.

\bibitem{latitude}
Sanjar Karaev, James Hook, and Pauli Miettinen.
\newblock Latitude: A model for mixed linear-tropical matrix factorization.
\newblock In {\em Proceedings of the 2018 SIAM International Conference on Data
  Mining}, pages 360--368. SIAM, 2018.

\bibitem{nmf}
Daniel~D Lee and H~Sebastian Seung.
\newblock Learning the parts of objects by non-negative matrix factorization.
\newblock {\em Nature}, 401(6755):788, 1999.

\bibitem{mfrecommender}
Yehuda Koren, Robert Bell, and Chris Volinsky.
\newblock Matrix factorization techniques for recommender systems.
\newblock {\em Computer}, (8):30--37, 2009.

\bibitem{bmf}
Zhongyuan Zhang, Tao Li, Chris Ding, and Xiangsun Zhang.
\newblock Binary matrix factorization with applications.
\newblock In {\em Seventh IEEE International Conference on Data Mining (ICDM
  2007)}, pages 391--400. IEEE, 2007.

\bibitem{bmf_1}
Zhong-Yuan Zhang, Tao Li, Chris Ding, Xian-Wen Ren, and Xiang-Sun Zhang.
\newblock Binary matrix factorization for analyzing gene expression data.
\newblock {\em Data Mining and Knowledge Discovery}, 20(1):28, 2010.

\bibitem{pmf}
Hans Laurberg, Mads~Gr{\ae}sb{\o}ll Christensen, Mark~D Plumbley, Lars~Kai
  Hansen, and S{\o}ren~Holdt Jensen.
\newblock Theorems on positive data: On the uniqueness of nmf.
\newblock {\em Computational intelligence and neuroscience}, 2008, 2008.

\bibitem{plsa}
Eric Gaussier and Cyril Goutte.
\newblock Relation between plsa and nmf and implications.
\newblock In {\em Proceedings of the 28th annual international ACM SIGIR
  conference on Research and development in information retrieval}, pages
  601--602. ACM, 2005.

\bibitem{survey}
Nimrod Rappoport and Ron Shamir.
\newblock Multi-omic and multi-view clustering algorithms: review and cancer
  benchmark.
\newblock {\em Nucleic acids research}, 46(20):10546--10562, 2018.

\bibitem{multi_nmf}
Jialu Liu, Chi Wang, Jing Gao, and Jiawei Han.
\newblock Multi-view clustering via joint nonnegative matrix factorization.
\newblock In {\em Proceedings of the 2013 SIAM International Conference on Data
  Mining}, pages 252--260. SIAM, 2013.

\bibitem{joint_nmf}
Shihua Zhang, Chun-Chi Liu, Wenyuan Li, Hui Shen, Peter~W Laird, and
  Xianghong~Jasmine Zhou.
\newblock Discovery of multi-dimensional modules by integrative analysis of
  cancer genomic data.
\newblock {\em Nucleic acids research}, 40(19):9379--9391, 2012.

\bibitem{pvc}
Shao-Yuan Li, Yuan Jiang, and Zhi-Hua Zhou.
\newblock Partial multi-view clustering.
\newblock In {\em Twenty-Eighth AAAI Conference on Artificial Intelligence},
  2014.

\bibitem{datafusion}
Marinka {\v{Z}}itnik and Bla{\v{z}} Zupan.
\newblock Data fusion by matrix factorization.
\newblock {\em IEEE transactions on pattern analysis and machine intelligence},
  37(1):41--53, 2015.

\bibitem{orthogonal}
Martin Stra{\v{z}}ar, Marinka {\v{Z}}itnik, Bla{\v{z}} Zupan, Jernej Ule, and
  Toma{\v{z}} Curk.
\newblock Orthogonal matrix factorization enables integrative analysis of
  multiple rna binding proteins.
\newblock {\em Bioinformatics}, 32(10):1527--1535, 2016.

\bibitem{algorithms}
Sanjar Karaev and Pauli Miettinen.
\newblock Algorithms for approximate subtropical matrix factorization.
\newblock {\em arXiv preprint arXiv:1707.08872}, 2017.

\bibitem{capricorn}
Sanjar Karaev and Pauli Miettinen.
\newblock Capricorn: An algorithm for subtropical matrix factorization.
\newblock In {\em Proceedings of the 2016 SIAM International Conference on Data
  Mining}, pages 702--710. SIAM, 2016.

\bibitem{cancer}
Sanjar Karaev and Pauli Miettinen.
\newblock Cancer: Another algorithm for subtropical matrix factorization.
\newblock In {\em Joint European Conference on Machine Learning and Knowledge
  Discovery in Databases}, pages 576--592. Springer, 2016.

\bibitem{nonliner_recomm}
Jason Weston, Ron~J Weiss, and Hector Yee.
\newblock Nonlinear latent factorization by embedding multiple user interests.
\newblock In {\em Proceedings of the 7th ACM conference on Recommender
  systems}, pages 65--68, 2013.

\bibitem{semiring}
Thanh Le~Van, Siegfried Nijssen, Matthijs Van~Leeuwen, and Luc De~Raedt.
\newblock Semiring rank matrix factorization.
\newblock {\em IEEE Transactions on Knowledge and Data Engineering},
  29(8):1737--1750, 2017.

\bibitem{tropical}
Bart De~Schutter and Bart De~Moor.
\newblock Matrix factorization and minimal state space realization in the
  max-plus algebra.
\newblock In {\em Proceedings of the 1997 American Control Conference (Cat. No.
  97CH36041)}, volume~5, pages 3136--3140. IEEE, 1997.

\bibitem{baccelli1992synchronization}
Fran{\c{c}}ois Baccelli, Guy Cohen, Geert~Jan Olsder, and Jean-Pierre Quadrat.
\newblock Synchronization and linearity: an algebra for discrete event systems.
\newblock 1992.

\bibitem{heidergott2014max}
Bernd Heidergott, Geert~Jan Olsder, and Jacob Van Der~Woude.
\newblock {\em Max Plus at work: modeling and analysis of synchronized systems:
  a course on Max-Plus algebra and its applications}, volume~48.
\newblock Princeton University Press, Princeton and Oxford, 2014.

\bibitem{linear_regression}
James Hook.
\newblock Linear regression over the max-plus semiring: algorithms and
  applications.
\newblock {\em arXiv preprint arXiv:1712.03499}, 2017.

\bibitem{tropical_svm}
Bernd G{\"a}rtner and Martin Jaggi.
\newblock Tropical support vector machines.
\newblock Technical report, Technical Report ACS-TR-362502-01, 2008.

\bibitem{tropical_geometry}
Liwen Zhang, Gregory Naitzat, and Lek-Heng Lim.
\newblock Tropical geometry of deep neural networks.
\newblock {\em arXiv preprint arXiv:1805.07091}, 2018.

\bibitem{optimal_control}
Philippe Declerck and Mohamed Khalid~Didi Alaoui.
\newblock Optimal control synthesis of timed event graphs with interval model
  specifications.
\newblock {\em IEEE Transactions on Automatic Control}, 55(2):518--523, 2009.

\bibitem{asymptotics}
Marianne Akian, Ravindra Bapat, and St{\'e}phane Gaubert.
\newblock Asymptotics of the perron eigenvalue and eigenvector using
  max-algebra.
\newblock {\em Comptes Rendus de l'Acad{\'e}mie des Sciences-Series
  I-Mathematics}, 327(11):927--932, 1998.

\bibitem{minmax}
Jeremy Gunawardena.
\newblock Min-max functions.
\newblock {\em Discrete Event Dynamic Systems}, 4(4):377--407, 1994.

\bibitem{finite_automata}
Hing Leung.
\newblock Limitedness theorem on finite automata with distance functions: an
  algebraic proof.
\newblock {\em Theoretical Computer Science}, 81(1):137--145, 1991.

\bibitem{gaubert}
Stephane Gaubert and Max Plus.
\newblock Methods and applications of (max,+) linear algebra.
\newblock In {\em Annual symposium on theoretical aspects of computer science},
  pages 261--282. Springer, 1997.

\bibitem{randomacol}
Amy~N Langville, Carl~D Meyer, Russell Albright, James Cox, and David Duling.
\newblock Algorithms, initializations, and convergence for the nonnegative
  matrix factorization.
\newblock {\em arXiv preprint arXiv:1407.7299}, 2014.

\bibitem{corrs_comparison}
Michael Clark.
\newblock A comparison of correlation measures.
\newblock {\em Center for Social Research, University of Notre Dame}, 4, 2013.

\bibitem{dcor}
G{\'a}bor~J Sz{\'e}kely, Maria~L Rizzo, et~al.
\newblock Brownian distance covariance.
\newblock {\em The annals of applied statistics}, 3(4):1236--1265, 2009.

\bibitem{polo}
Ziv Bar-Joseph, David~K Gifford, and Tommi~S Jaakkola.
\newblock Fast optimal leaf ordering for hierarchical clustering.
\newblock {\em Bioinformatics}, 17(suppl\_1):S22--S29, 2001.

\bibitem{murtagh2014ward}
Fionn Murtagh and Pierre Legendre.
\newblock Ward’s hierarchical agglomerative clustering method: which
  algorithms implement ward’s criterion?
\newblock {\em Journal of classification}, 31(3):274--295, 2014.

\bibitem{parker2009supervised}
Joel~S Parker, Michael Mullins, Maggie~CU Cheang, Samuel Leung, David Voduc,
  Tammi Vickery, Sherri Davies, Christiane Fauron, Xiaping He, Zhiyuan Hu,
  et~al.
\newblock Supervised risk predictor of breast cancer based on intrinsic
  subtypes.
\newblock {\em Journal of clinical oncology}, 27(8):1160, 2009.

\bibitem{cerami}
Ethan Cerami, Jianjiong Gao, Ugur Dogrusoz, Benjamin~E Gross, Selcuk~Onur
  Sumer, B{\"u}lent~Arman Aksoy, Anders Jacobsen, Caitlin~J Byrne, Michael~L
  Heuer, Erik Larsson, et~al.
\newblock The cbio cancer genomics portal: an open platform for exploring
  multidimensional cancer genomics data, 2012.

\bibitem{gao}
Jianjiong Gao, B{\"u}lent~Arman Aksoy, Ugur Dogrusoz, Gideon Dresdner, Benjamin
  Gross, S~Onur Sumer, Yichao Sun, Anders Jacobsen, Rileen Sinha, Erik Larsson,
  et~al.
\newblock Integrative analysis of complex cancer genomics and clinical profiles
  using the cbioportal.
\newblock {\em Science signaling}, 6(269):pl1--pl1, 2013.

\bibitem{rousseeuw1987silhouettes}
Peter~J. Rousseeuw.
\newblock Silhouettes: A graphical aid to the interpretation and validation of
  cluster analysis.
\newblock {\em Journal of Computational and Applied Mathematics}, 20:53 -- 65,
  1987.

\bibitem{boutsidis2008svd}
Christos Boutsidis and Efstratios Gallopoulos.
\newblock Svd based initialization: A head start for nonnegative matrix
  factorization.
\newblock {\em Pattern recognition}, 41(4):1350--1362, 2008.

\end{thebibliography}

\clearpage
\appendix

\section*{Data embedding and prediction by sparse tropical matrix factorization \\ Supplementary Material}

Amra Omanović, Hilal Kazan, Polona Oblak and Tomaž Curk

\renewcommand{\figurename}{Supplementary Figure S}

\section{Synthetic data}

In Supplementary Figure S~\ref{synth_original_factors} we present original factor matrices of a smaller synthetic dataset. Approximation matrices of rank 4 are shown in Supplementary Figure S~\ref{synth_orig_miss_approx_rank_4}, while factor matrices are in Supplementary Figure S~\ref{synth_factor_matrices} and latent  matrices in Supplementary Figure S~\ref{synth_latent_matrices}.

\begin{figure}[!htb]
    \centering
    \captionsetup{justification=centering}
    \includegraphics[scale=0.23]{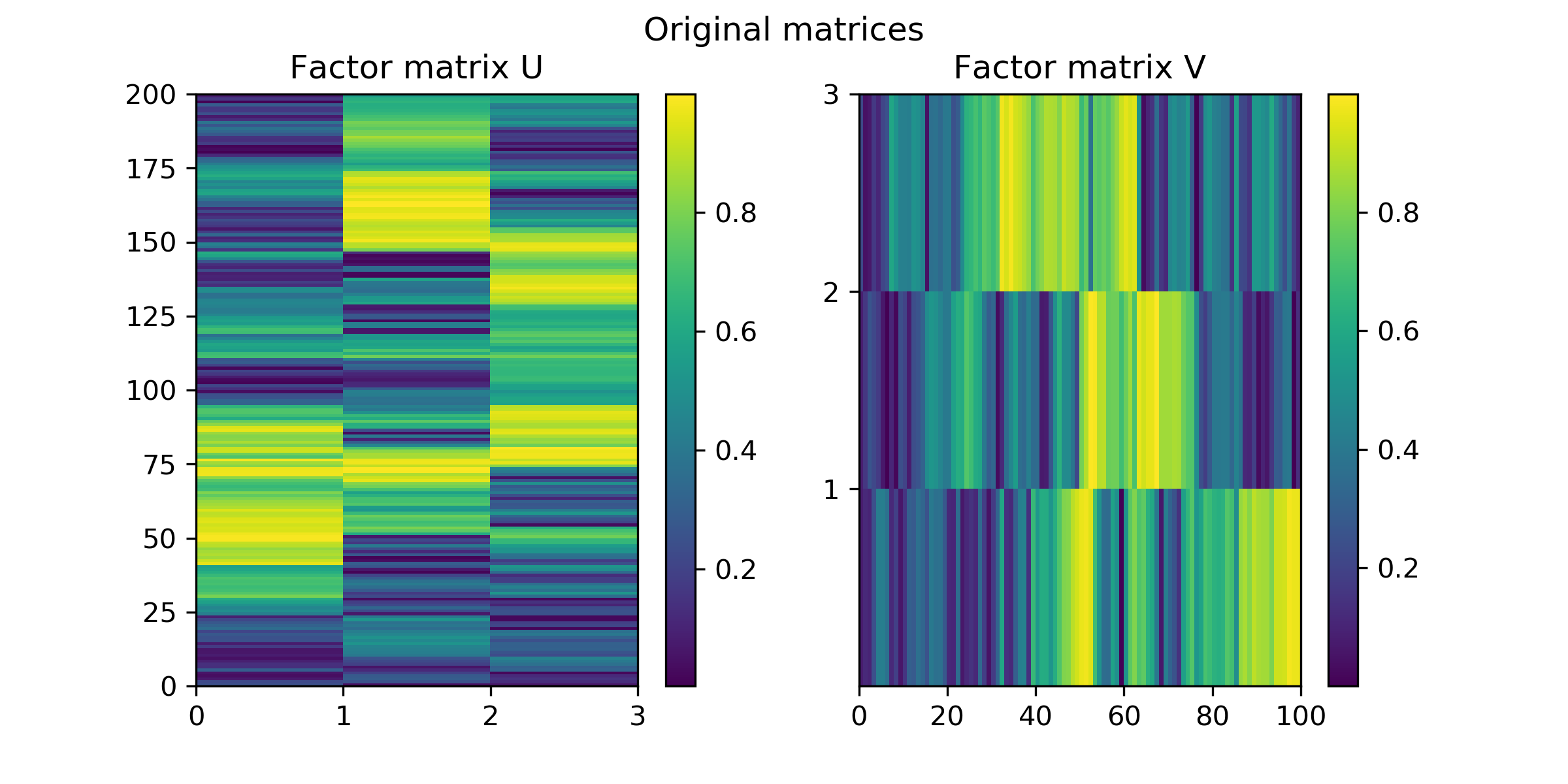}
    \caption{Original factor matrices.}
    \label{synth_original_factors}
\end{figure}

\begin{figure}[!htb]
    \centering
    \captionsetup{justification=centering}
    \includegraphics[scale=0.23]{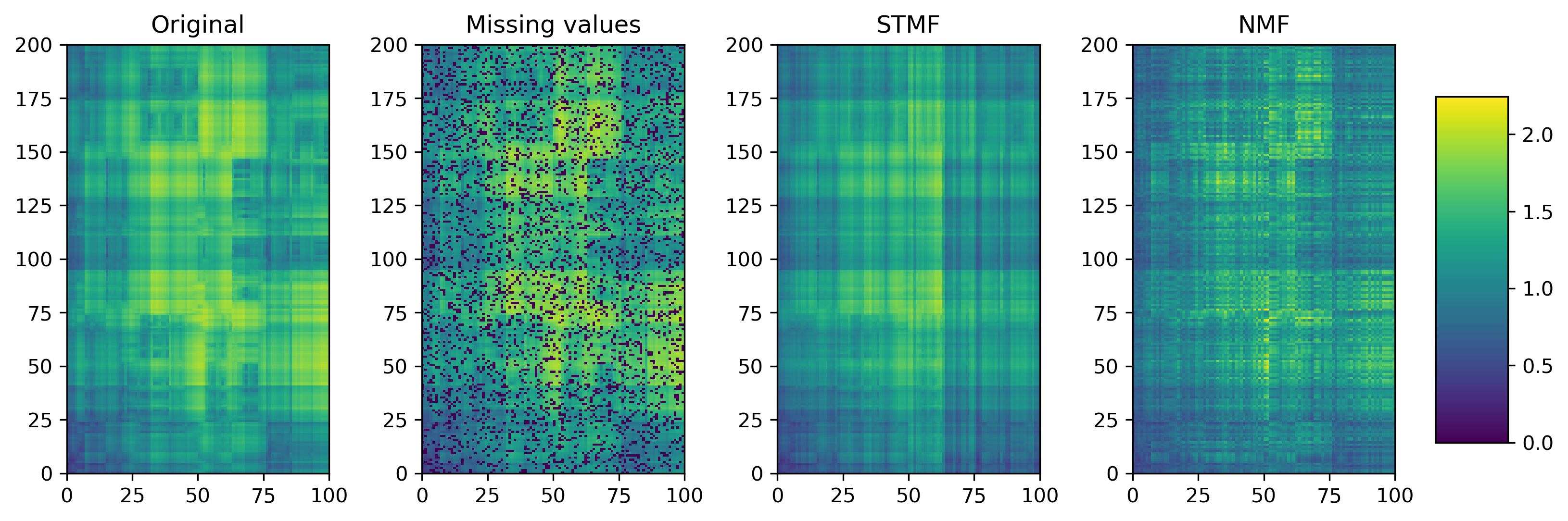}
    \caption{A comparison between \texttt{STMF}'s and \texttt{NMF}'s \\ predictions of rank 4 approximations on  $(200, 100)$ \\ synthetic $(\max,+)$ matrix with $20\%$ missing values.}
    \label{synth_orig_miss_approx_rank_4}
\end{figure}

\begin{figure}[!htb] 
    \centering
    \captionsetup{justification=centering}
    \begin{subfigure}{.5\textwidth}
    \includegraphics[height=0.42\textwidth, width=\textwidth]{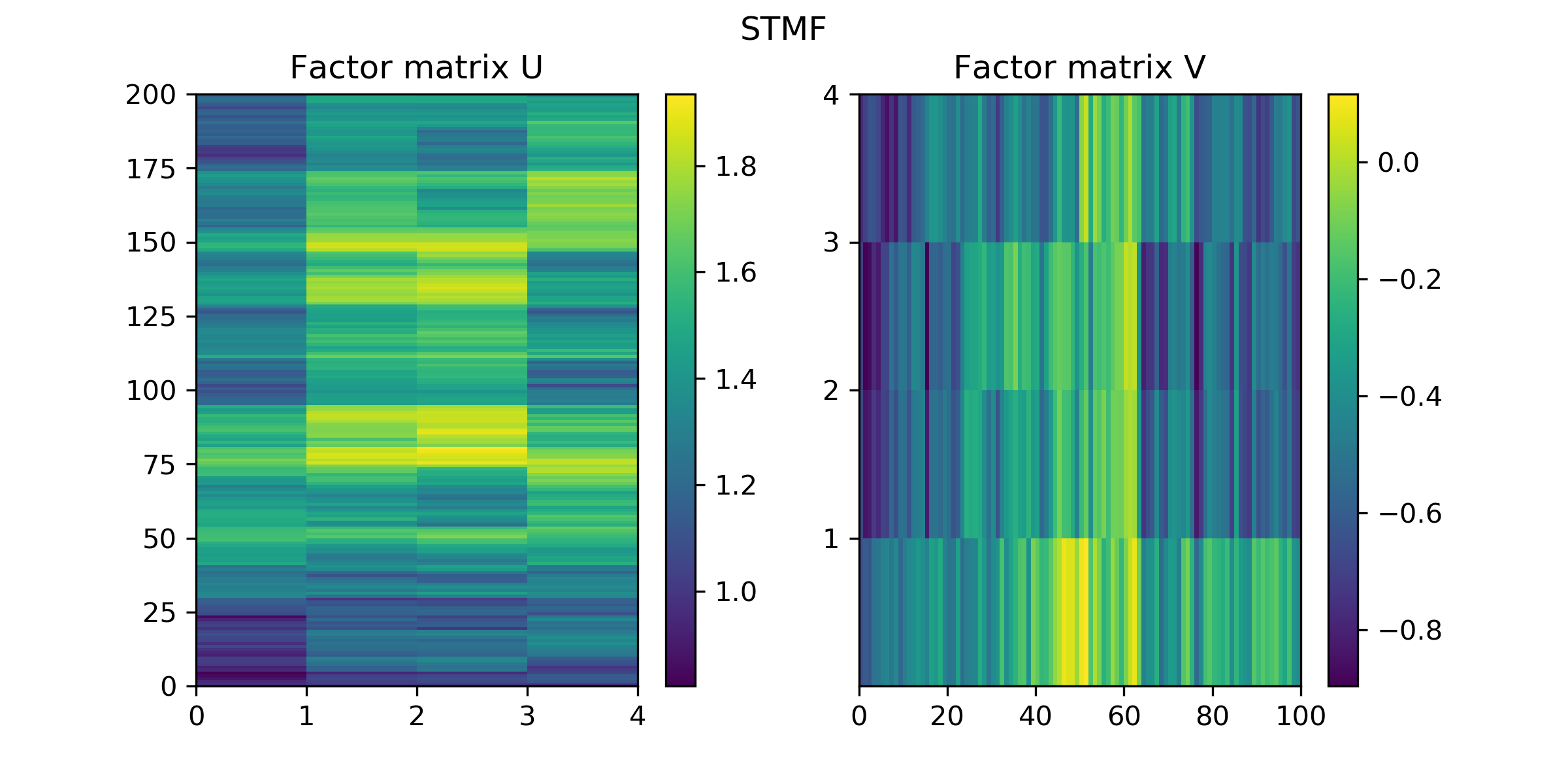}
    \caption{Factor matrices $U_{\texttt{STMF}}, V_{\texttt{STMF}}$ from \texttt{STMF}}
\end{subfigure}\hfill%
\begin{subfigure}{.5\textwidth}
    \includegraphics[height=0.42\textwidth, width=\textwidth]{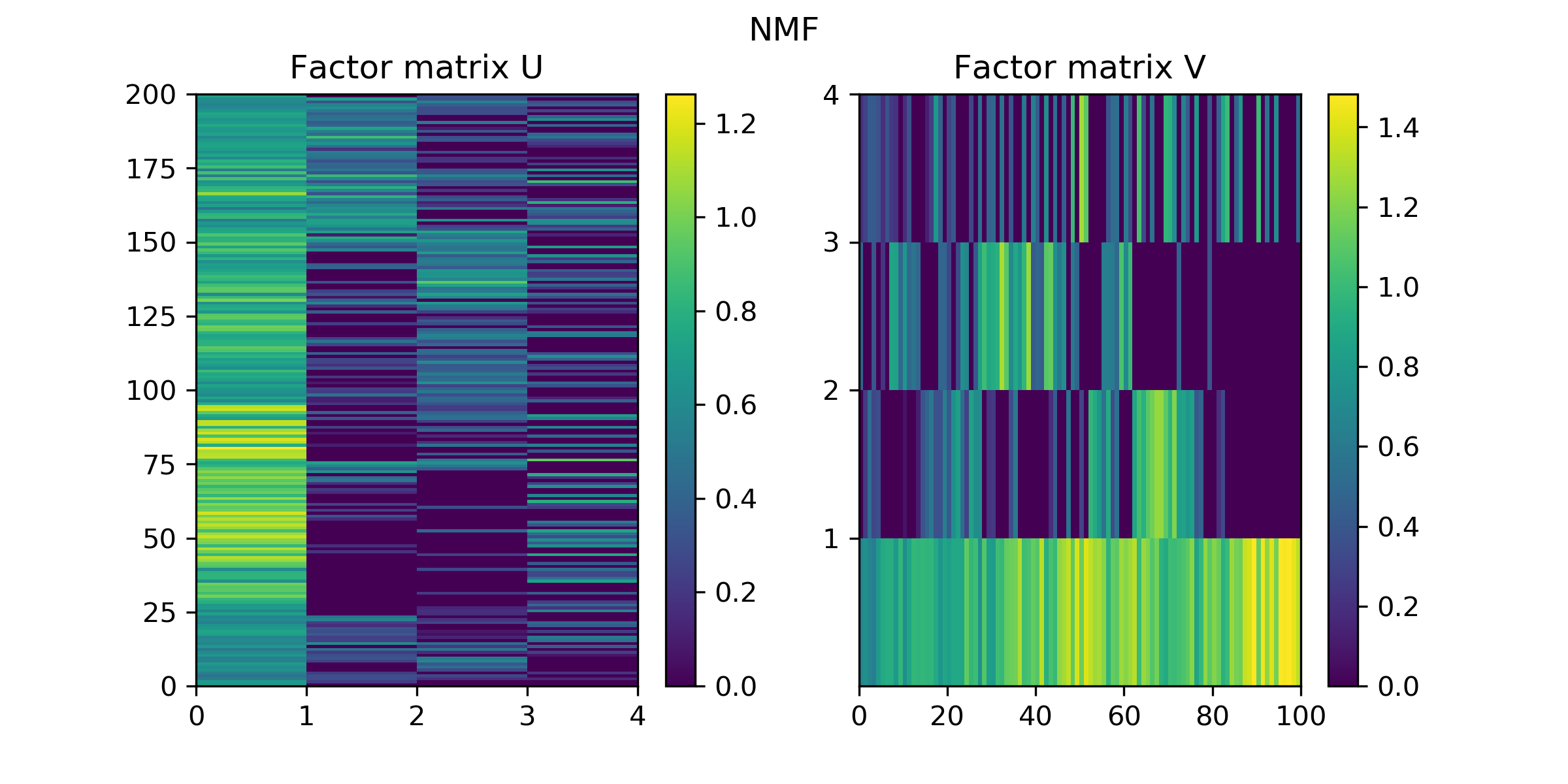}
    \caption{Factor matrices $U_{\texttt{NMF}}, V_{\texttt{NMF}}$ from \texttt{NMF}}
\end{subfigure}
    \caption{Factor matrices $U_{\texttt{STMF}}, V_{\texttt{STMF}}$ and  $U_{\texttt{NMF}}, V_{\texttt{NMF}}$ from \texttt{STMF} and \texttt{NMF},  respectively.}
    \label{synth_factor_matrices}
\end{figure}

\begin{figure}[!htb] 
    \centering
    \begin{subfigure}[t]{.5\textwidth}
        \includegraphics[height=0.42\textwidth, width=\textwidth]{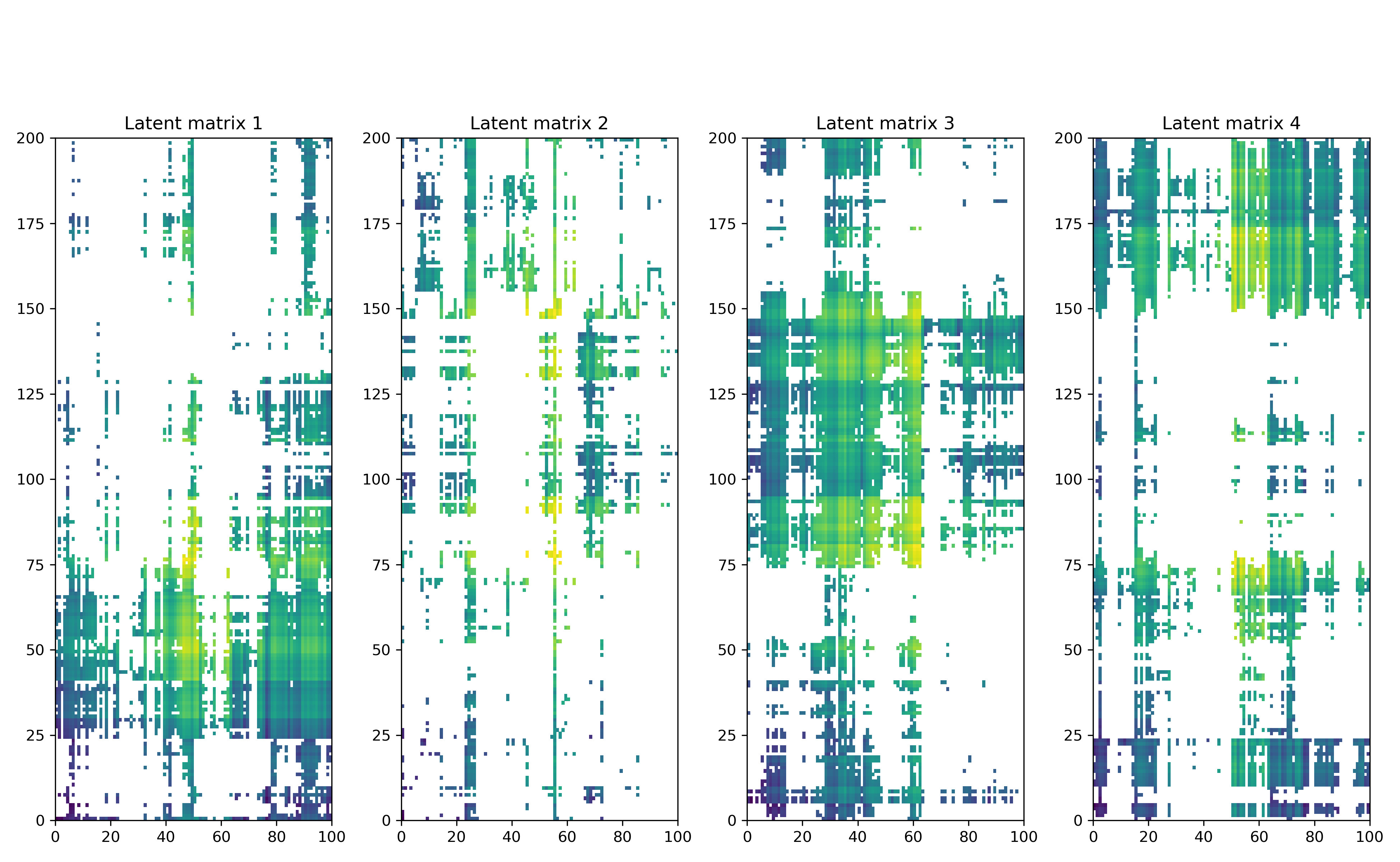}
        \caption{Latent matrices $R_{\texttt{STMF}}^{(i)}$, $i \in \{1, 4\}$, where \\ white represents the element which does not \\ contribute to the approximation $R_{\texttt{STMF}}$.}
    \end{subfigure}\hfill%
    \begin{subfigure}[t]{.5\textwidth}
        \includegraphics[height=0.42\textwidth, width=\textwidth]{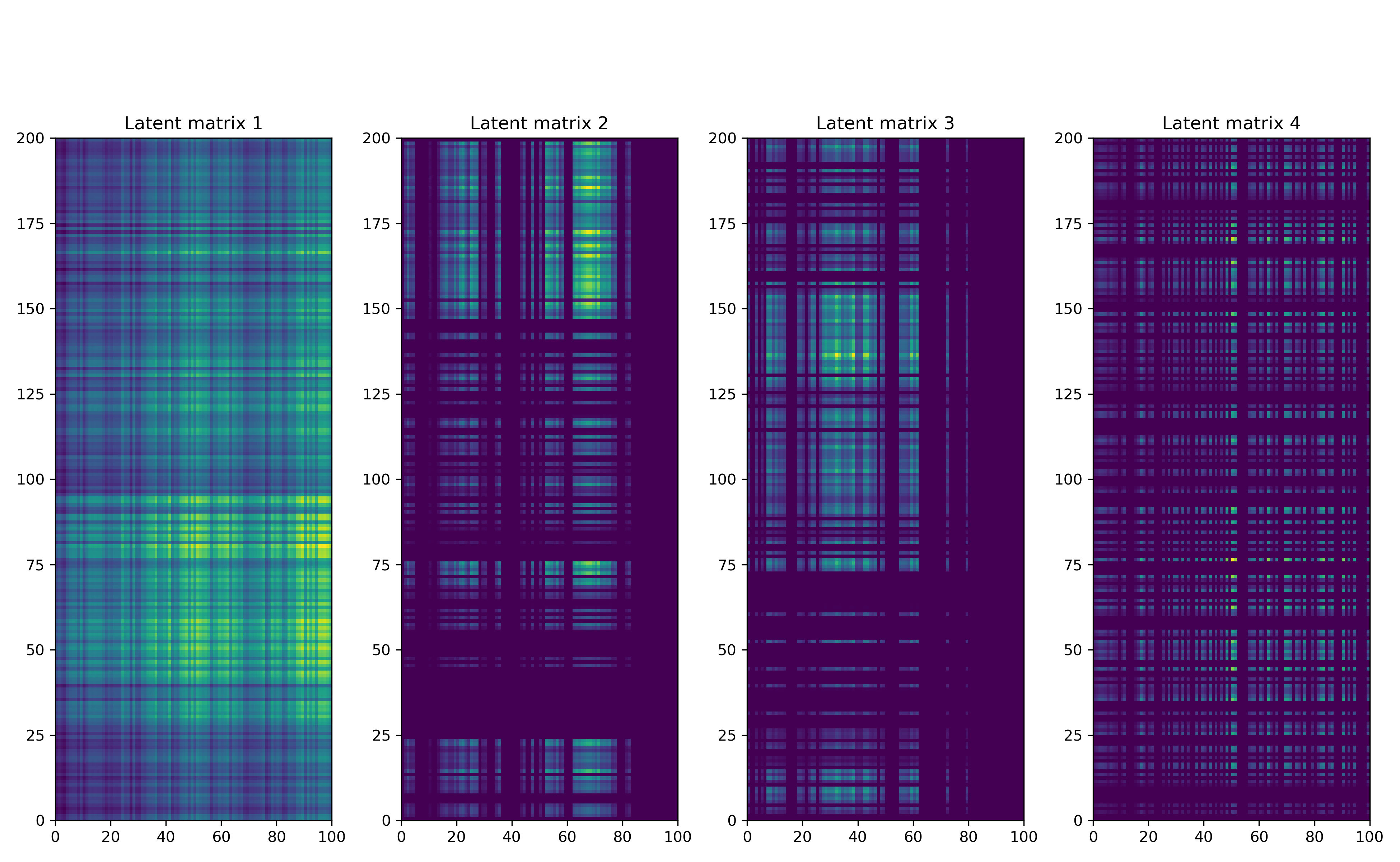}
        \caption{Latent matrices $R_{\texttt{NMF}}^{(i)}$, $i \in \{1, 4\}$.}
    \end{subfigure}
    \caption{\texttt{STMF}'s and \texttt{NMF}'s latent matrices.}
    \label{synth_latent_matrices}
\end{figure}

\subsection{Ordering techniques}
In Supplementary Figure S~\ref{synth_permutation} we present five large synthetic datasets used for the ordering techniques experiment. Effect of different ordering strategies for these five datasets is shown in Supplementary Figure S~\ref{synth_violin_plots}.
\begin{figure}[!htb]
    \centering
    \captionsetup{justification=centering}
    \includegraphics[scale=0.25]{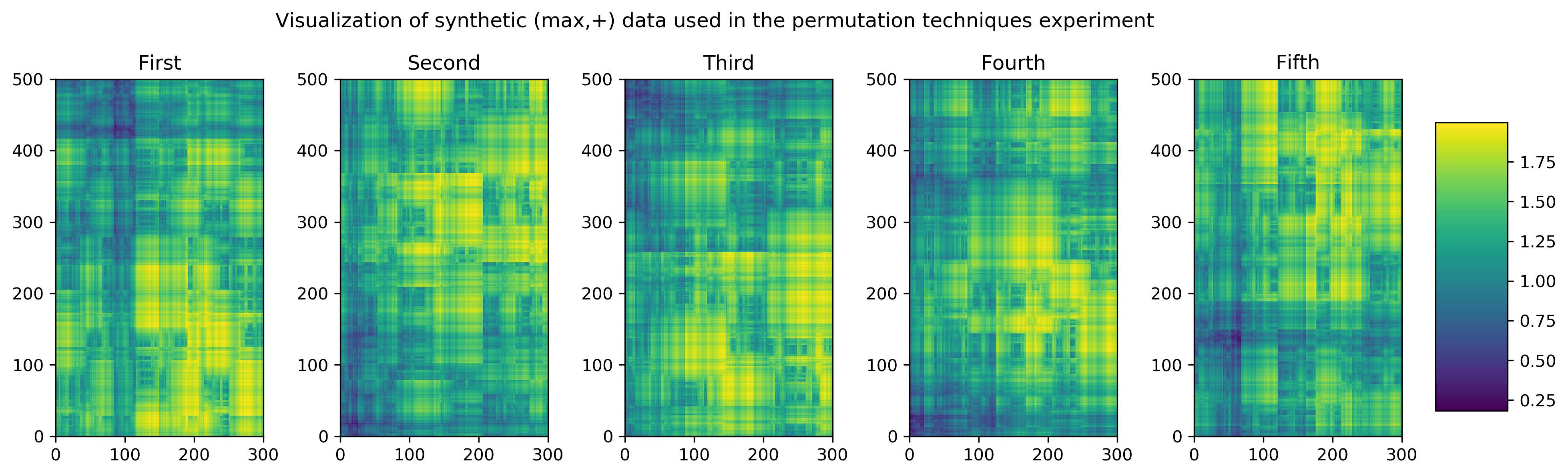}
    \caption{Synthetic data used in the ordering techniques experiment.}
    \label{synth_permutation}
\end{figure}

\begin{figure*}
        \centering
          \begin{subfigure}[b]{0.475\textwidth}
            \centering
            \includegraphics[width=\textwidth]{images/ordering/violin_plots_analysis_100.png}
        \end{subfigure}
        \vfill
        \begin{subfigure}[b]{0.475\textwidth}
            \centering
            \includegraphics[width=\textwidth]{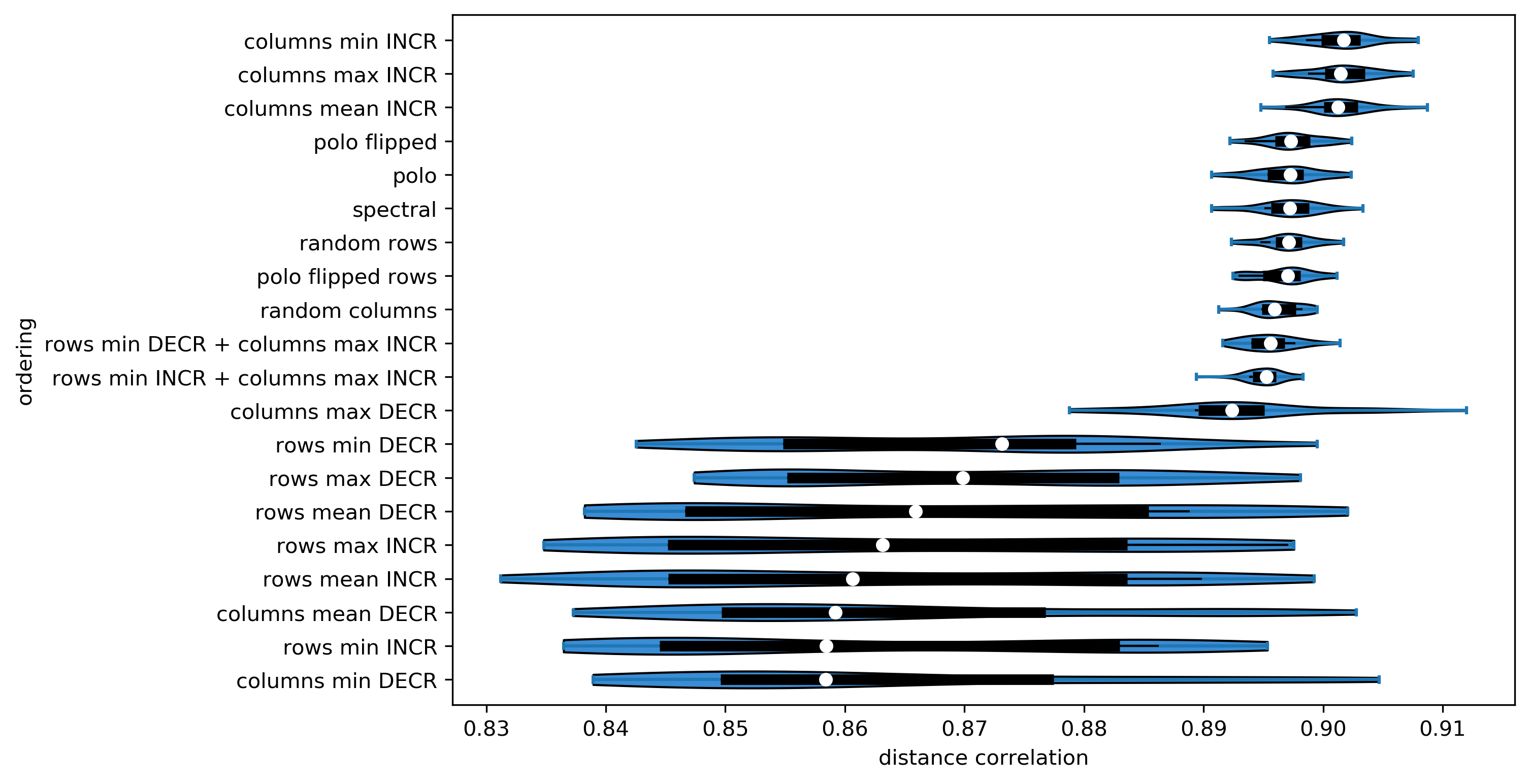}
        \end{subfigure}
        \hfill
        \begin{subfigure}[b]{0.475\textwidth}  
            \centering 
            \includegraphics[width=\textwidth]{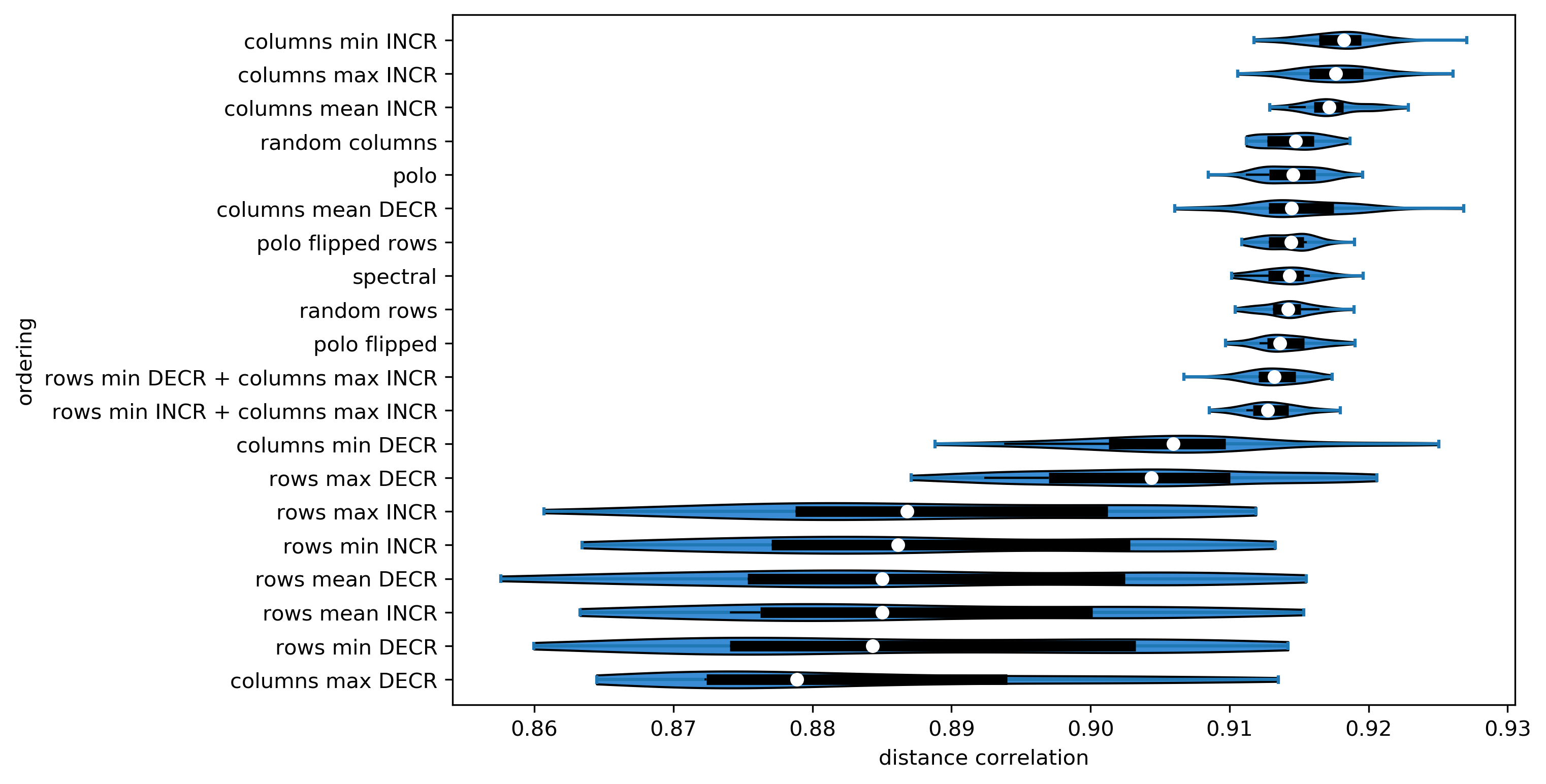}
        \end{subfigure}
        \vskip\baselineskip
        \begin{subfigure}[b]{0.475\textwidth}   
            \centering 
            \includegraphics[width=\textwidth]{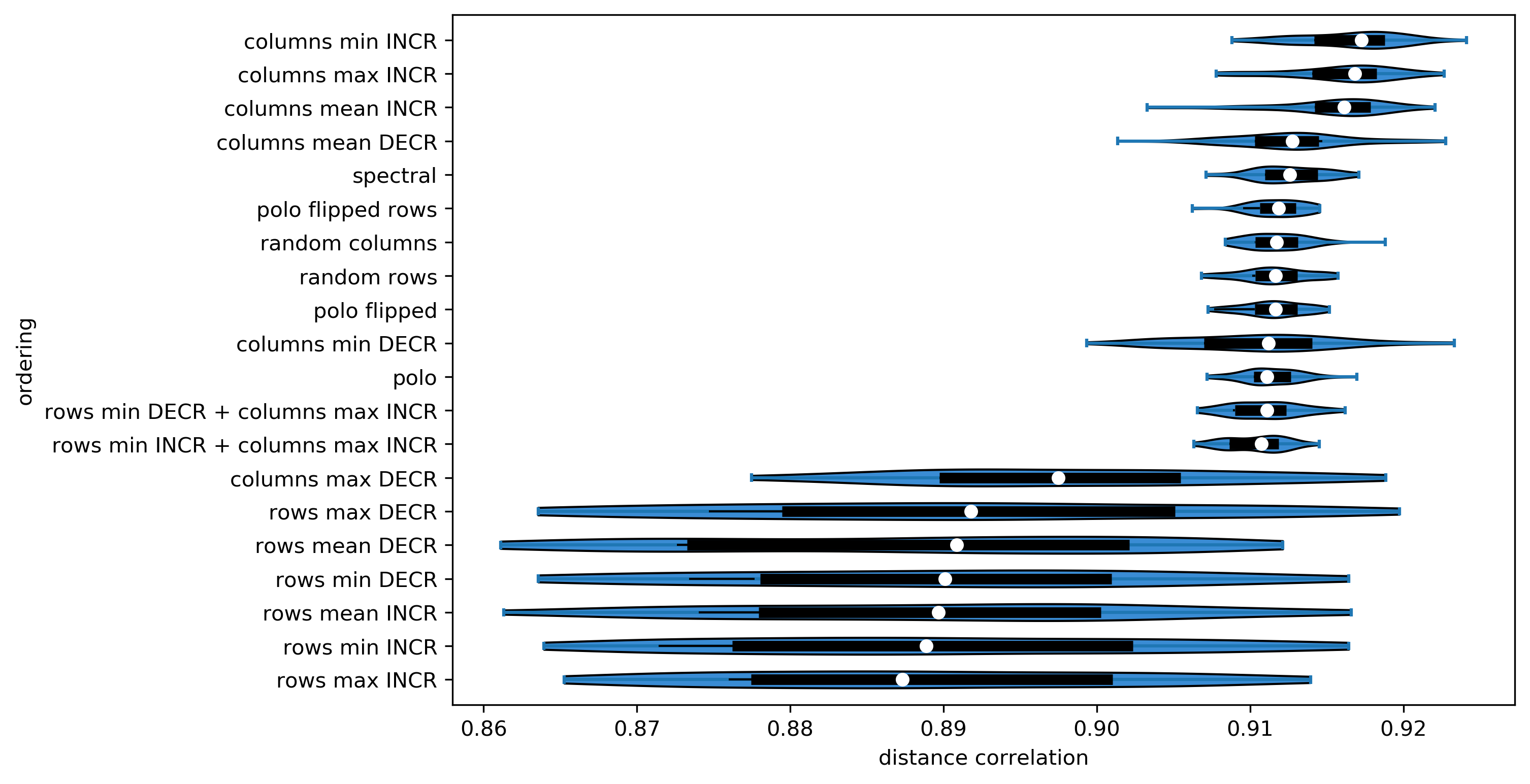}
        \end{subfigure}
        \hfill
        \begin{subfigure}[b]{0.475\textwidth}
            \centering 
            \includegraphics[width=\textwidth]{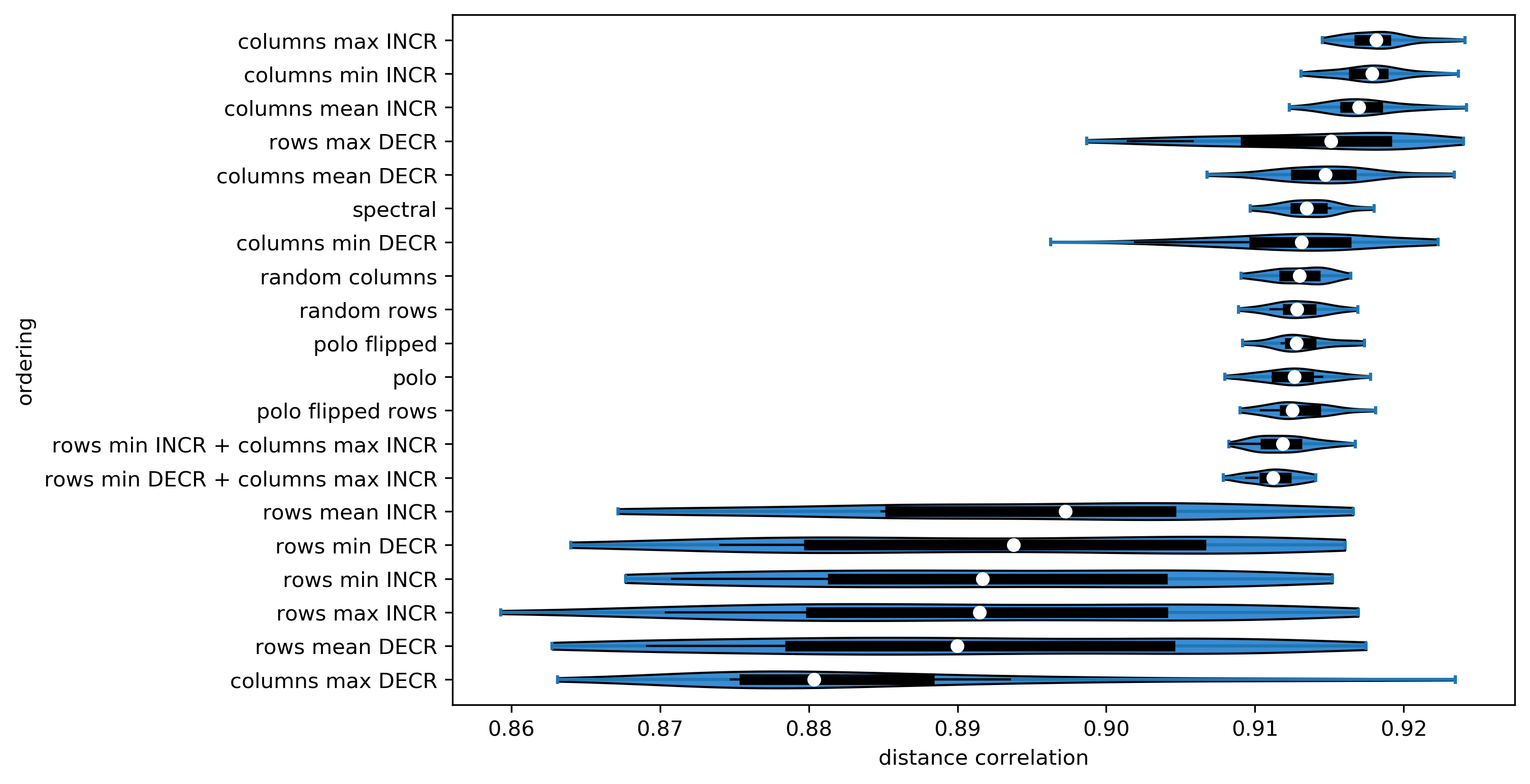}
        \end{subfigure}
        \caption[Effect of ordering strategy on achieved distance correlation by \texttt{STMF}, on five $500 \times 300$ synthetic  $(\max, +)$ matrices from Figure~\ref{synth_permutation}.]
        {\small Effect of ordering strategy on achieved distance correlation by \texttt{STMF}, on five $500 \times 300$ synthetic  $(\max, +)$ matrices from Figure~\ref{synth_permutation}.} 
        \label{synth_violin_plots}
\end{figure*}

\clearpage
\section{Real data}
In this section, we present results on real data using best approximation matrices of the corresponding rank. For the \texttt{STMF} method, we use Random Acol initialization described in the paper, and for the \texttt{NMF} method, we use a method \texttt{NNDSVD}~\cite{boutsidis2008svd} designed to enhance the initialization stage of \texttt{NMF}.
\label{supp_section_two}

\subsection{\texttt{BIC}}

In Supplementary Figure S~\ref{bic_silh_plot} we present the silhouette plot of \texttt{BIC} data which contains five clusters. Values can range from -1 to 1, where a value of 0 indicates that the sample (patient) is on or very close to the decision boundary between two neighboring clusters.
\begin{figure}[!htb]
    \centering
    \captionsetup{justification=centering}
    \includegraphics[scale=0.3]{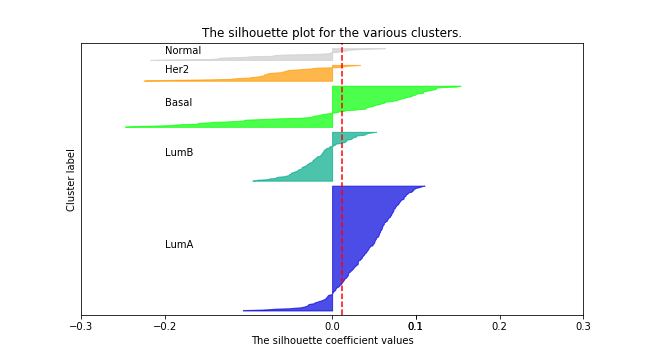}
    \caption{The silhouette plot of \texttt{BIC}  data.}
    \label{bic_silh_plot}
\end{figure}

In Supplementary Figure S~\ref{datasets_agglom}, we plot distributions of original data and feature agglomeration data for all eight datasets.
\begin{figure}[!htb]
    \centering
    \captionsetup{justification=centering}
    \includegraphics[scale=0.25]{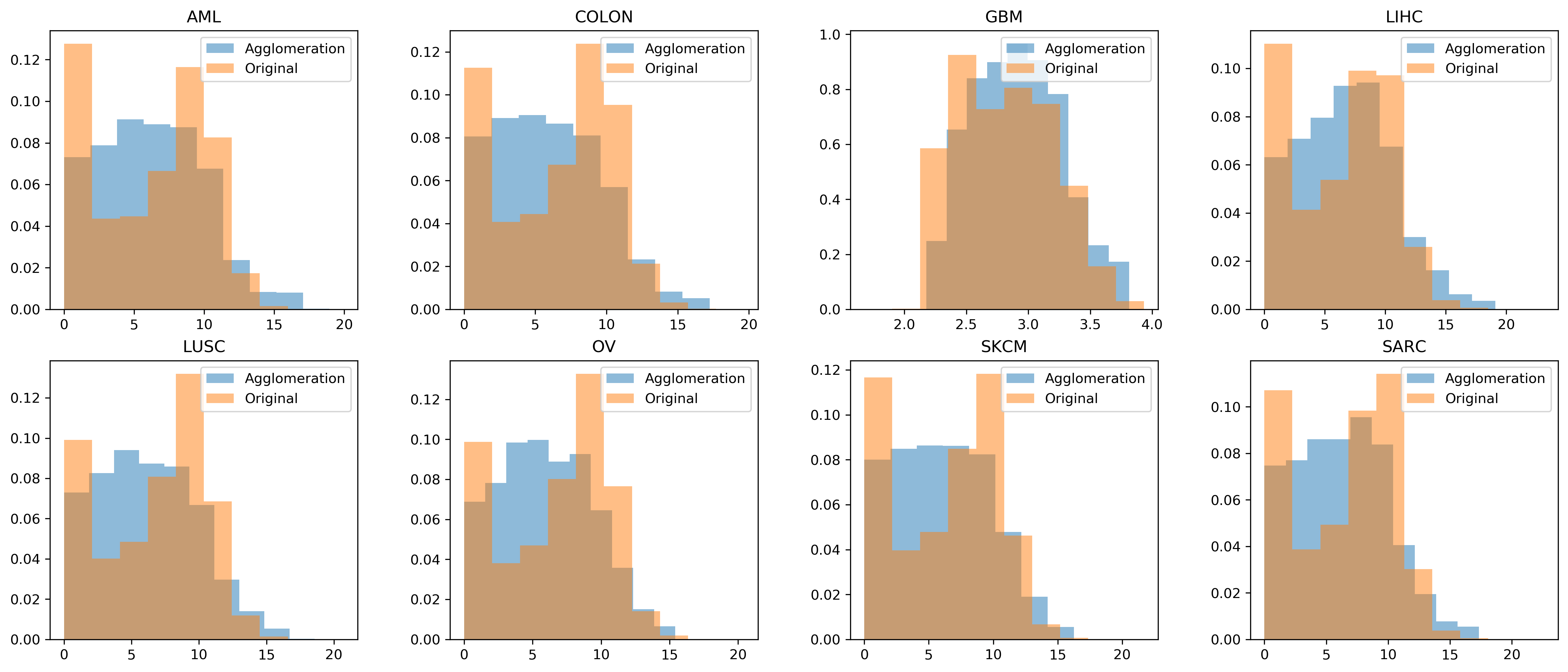}
    \caption{Distribution of original and feature \\ agglomeration data.}
    \label{datasets_agglom}
\end{figure}

In Supplementary Figure S~\ref{bic_pear_spear}, we present Pearson and Spearman correlation results on \texttt{BIC} data.
\begin{figure*}
        \centering
        \begin{subfigure}[b]{0.475\textwidth}
            \centering
            \includegraphics[width=\textwidth]{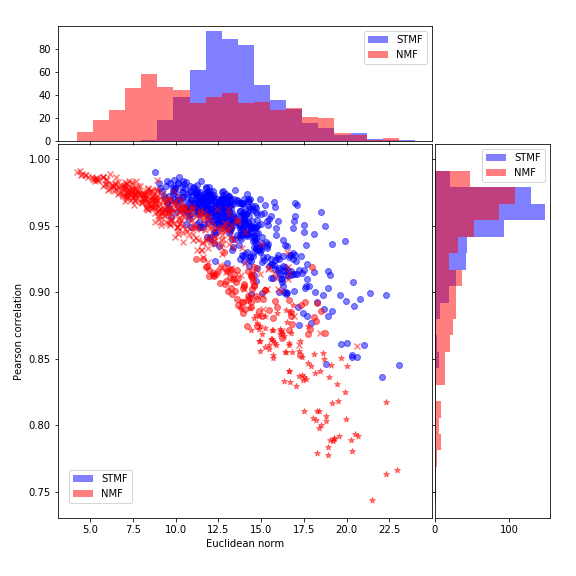}
            \caption{Pearson correlation}
        \end{subfigure}
        \hfill
        \begin{subfigure}[b]{0.475\textwidth}  
            \centering 
            \includegraphics[width=\textwidth]{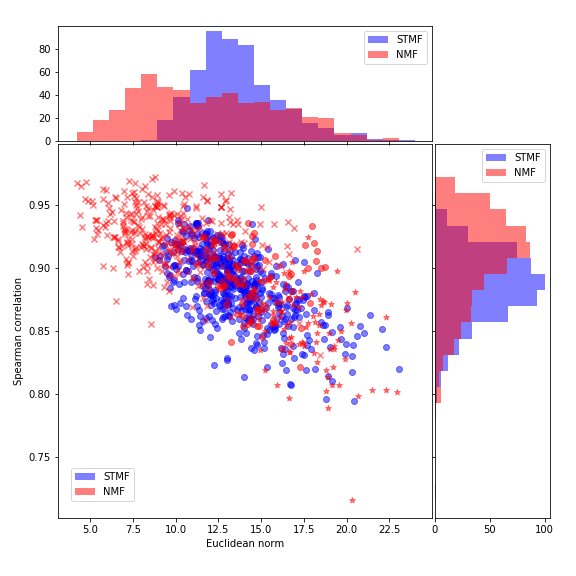}
            \caption{Spearman correlation}
        \end{subfigure}
        \caption{Pearson and Spearman correlation on \texttt{BIC} data.}
        \label{bic_pear_spear}
\end{figure*}

\clearpage
\subsection{\texttt{AML}}
\begin{figure}[!htb]
\captionsetup{justification=centering}
\begin{subfigure}{0.66\textwidth}
\centering
\includegraphics[scale=0.27]{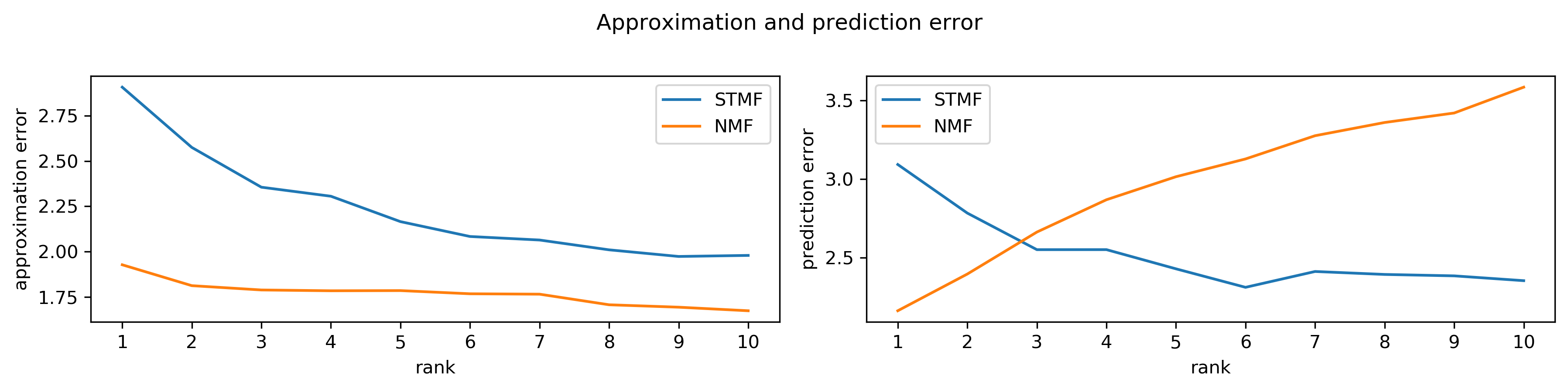}
\end{subfigure}
\begin{subfigure}{0.33\textwidth}
\centering
\includegraphics[scale=0.27]{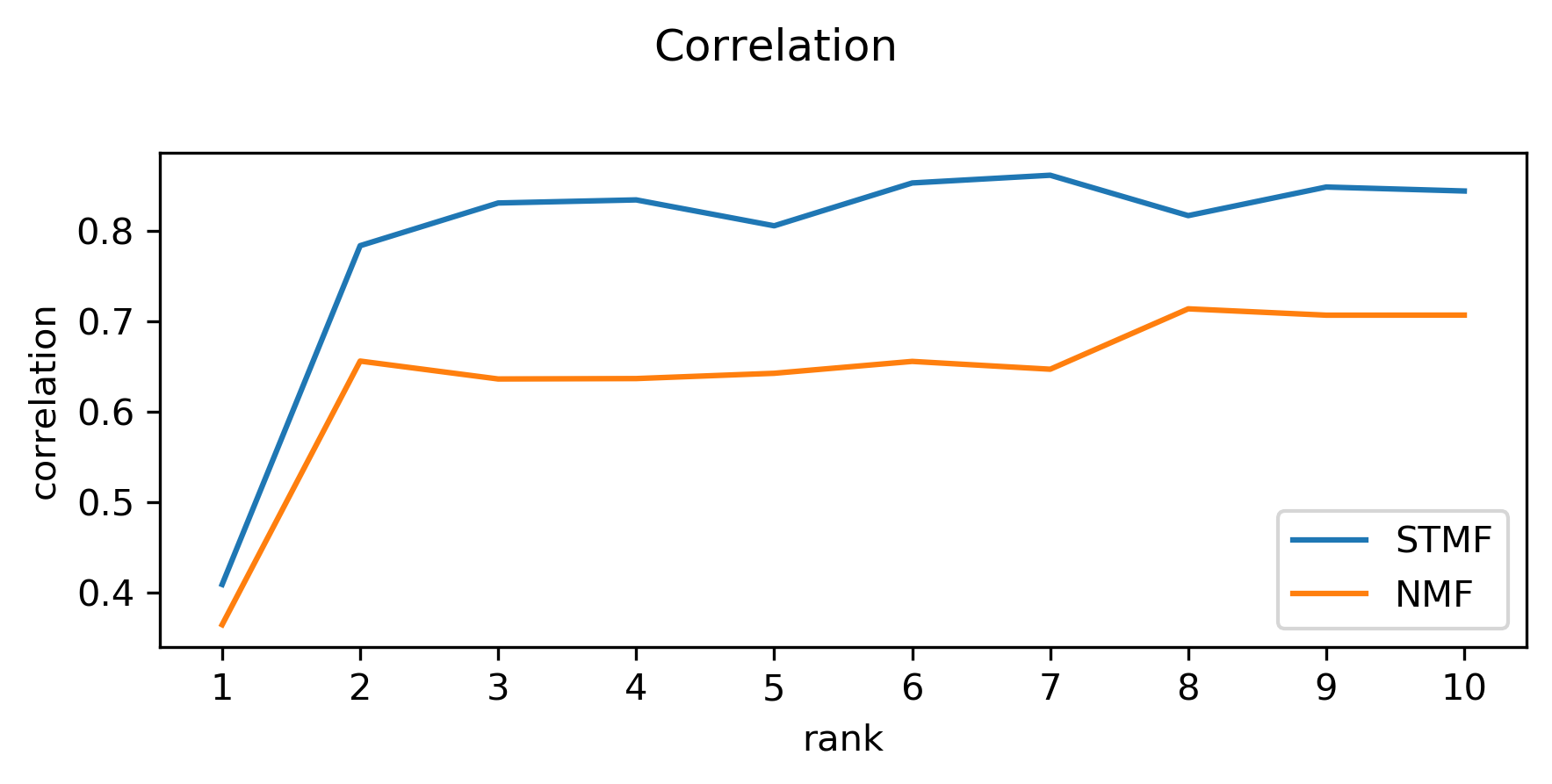}
\end{subfigure}
\caption{Difference between approximation and \\prediction RMSE and distance correlation of \texttt{STMF} \\ and \texttt{NMF} on \texttt{AML} data.}
\label{aml_error_and_corr}
\end{figure}

\begin{figure}[!htb]
    \centering
    \captionsetup{justification=centering}
    \includegraphics[scale=0.35]{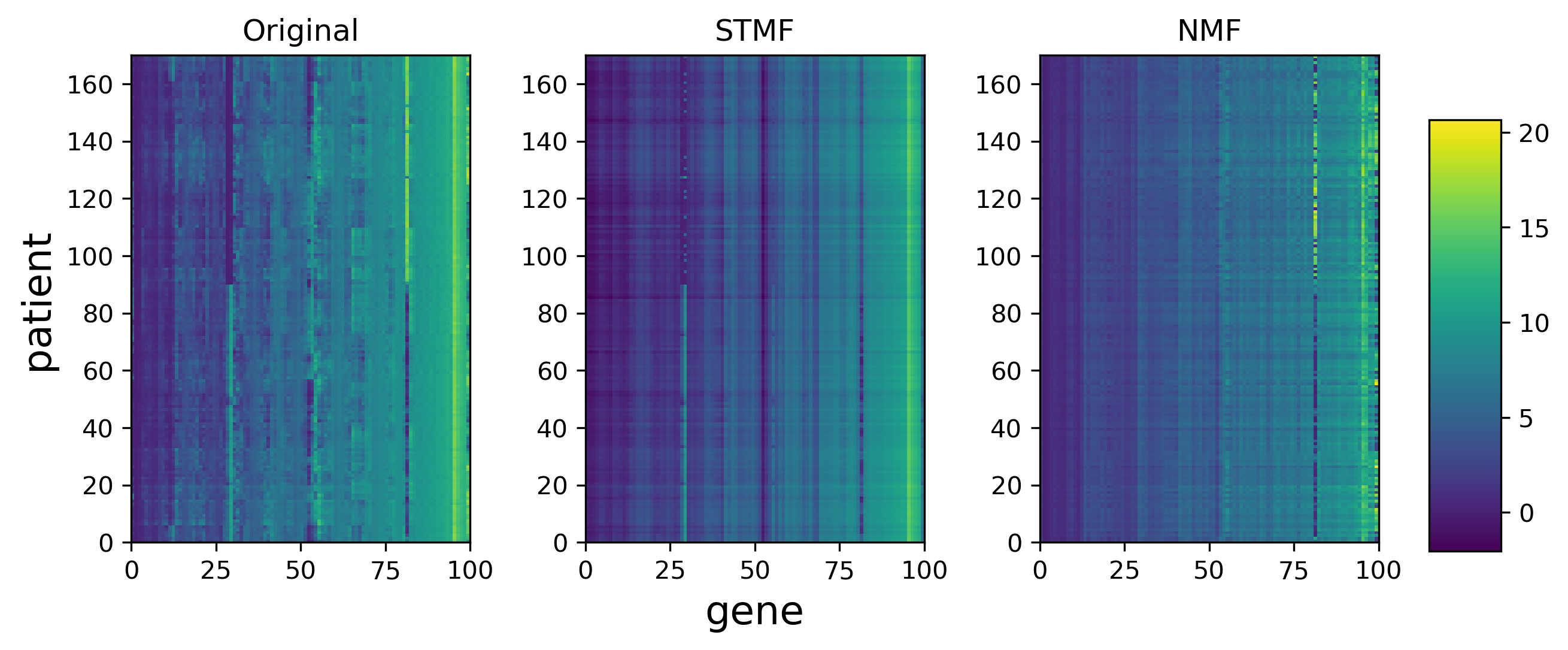}
    \caption{A comparison between \texttt{STMF}'s and \texttt{NMF}'s \\ predictions of rank 3 approximations on  \texttt{AML} data with $20\%$ missing values.}
    \label{aml_orig_approx}
\end{figure}

\begin{figure}[!htb] 
    \centering
    \captionsetup{justification=centering}
    \begin{subfigure}{.5\textwidth}
    \includegraphics[height=0.42\textwidth, width=\textwidth]{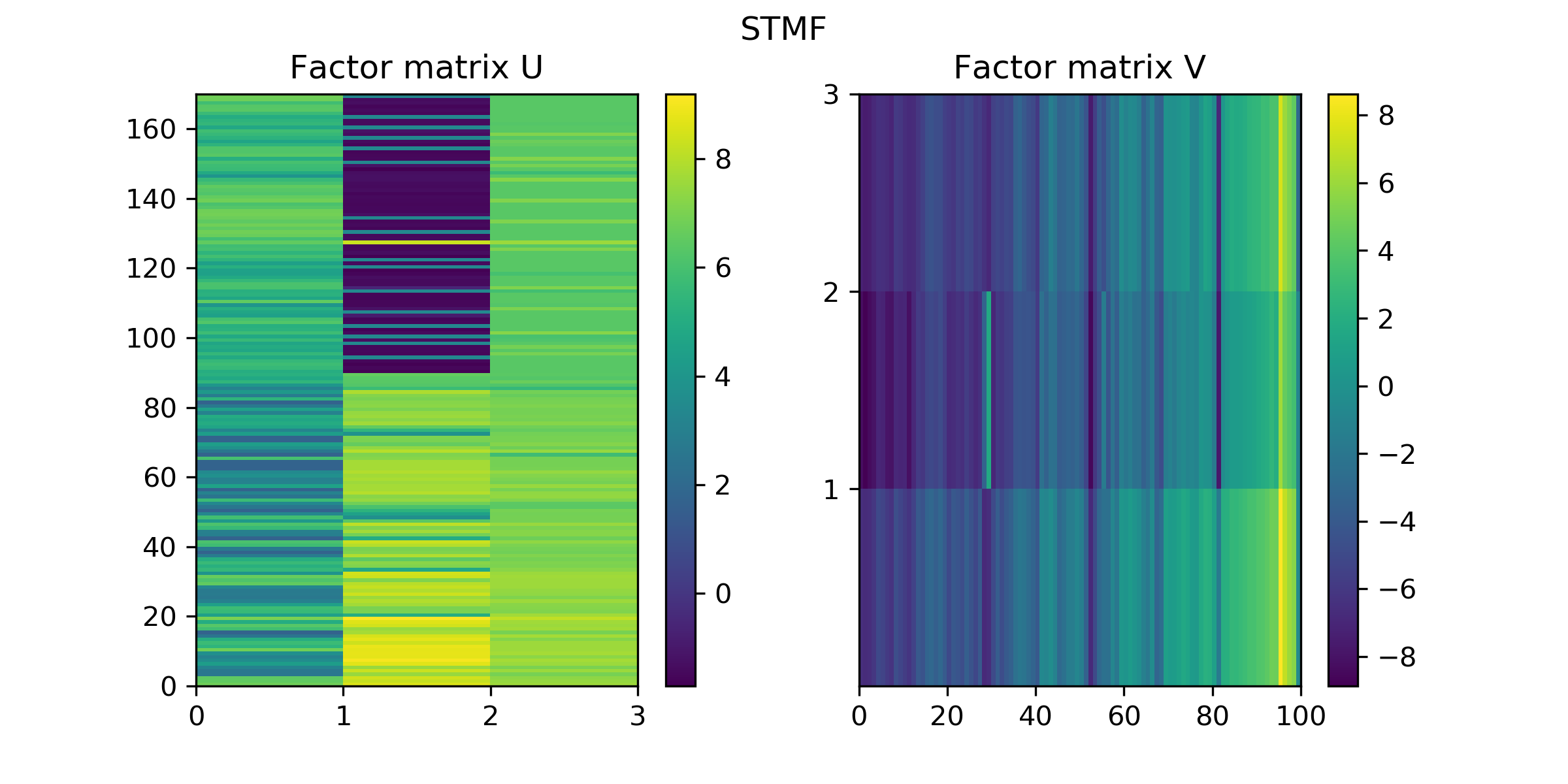}
    \caption{Factor matrices $U_{\texttt{STMF}}, V_{\texttt{STMF}}$ from \texttt{STMF}.}
\end{subfigure}\hfill%
\begin{subfigure}{.5\textwidth}
    \includegraphics[height=0.42\textwidth, width=\textwidth]{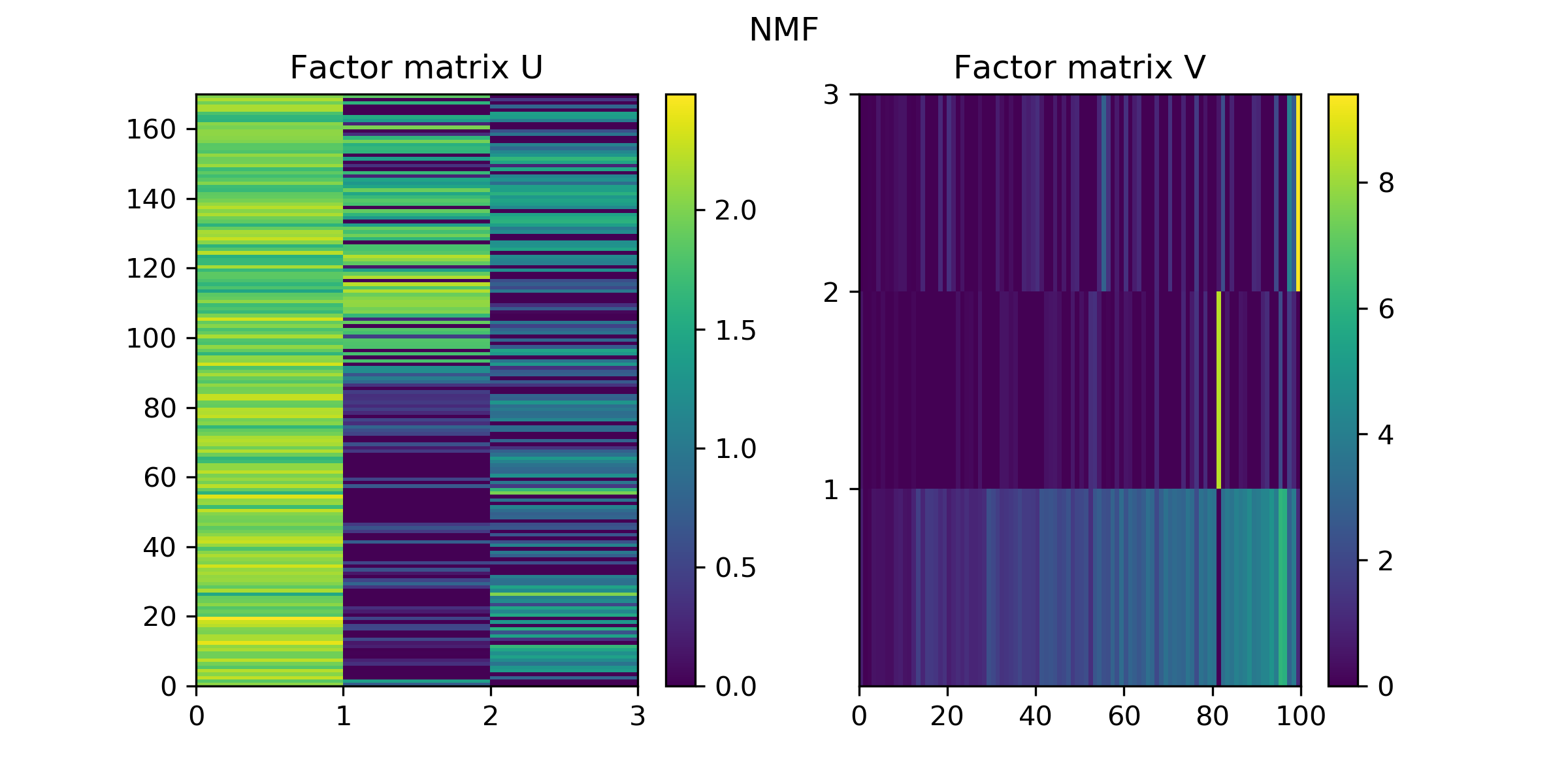}
    \caption{Factor matrices $U_{\texttt{NMF}}, V_{\texttt{NMF}}$ from \texttt{NMF}.}
\end{subfigure}
    \caption{Factor matrices $U_{\texttt{STMF}}, V_{\texttt{STMF}}$ and  $U_{\texttt{NMF}}, V_{\texttt{NMF}}$ from \texttt{STMF} and \texttt{NMF} on \texttt{AML} data,  respectively.}
    \label{aml_factor_matrices}
\end{figure}

\begin{figure}[!htb] 
    \centering
    \begin{subfigure}[t]{.5\textwidth}
        \includegraphics[height=0.42\textwidth, width=\textwidth]{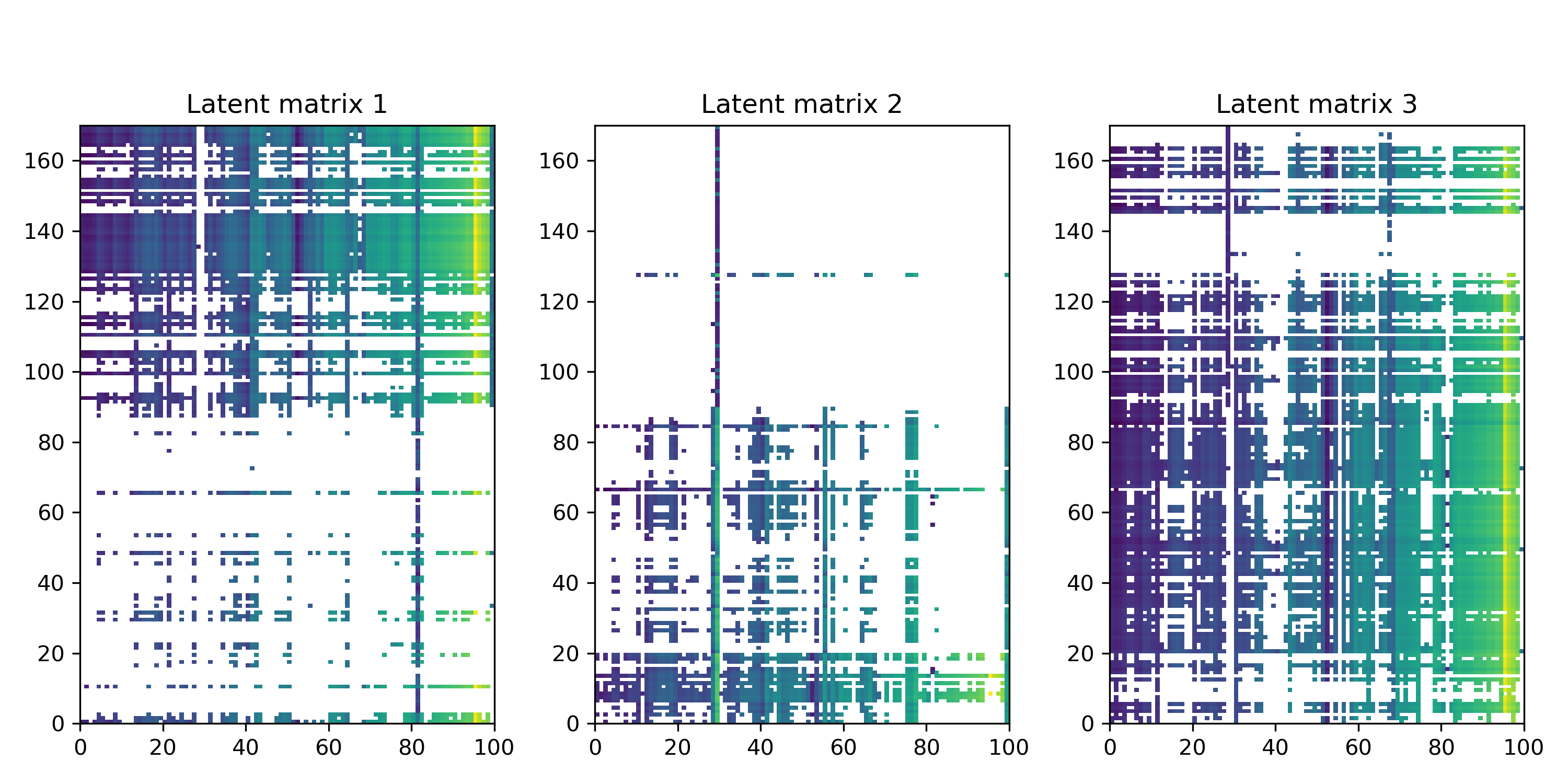}
        \caption{Latent matrices $R_{\texttt{STMF}}^{(i)}$, $i \in \{1, 3\}$, where \\ white represents the element which does not \\ contribute to the approximation $R_{\texttt{STMF}}$.}
    \end{subfigure}\hfill%
    \begin{subfigure}[t]{.5\textwidth}
        \includegraphics[height=0.42\textwidth, width=\textwidth]{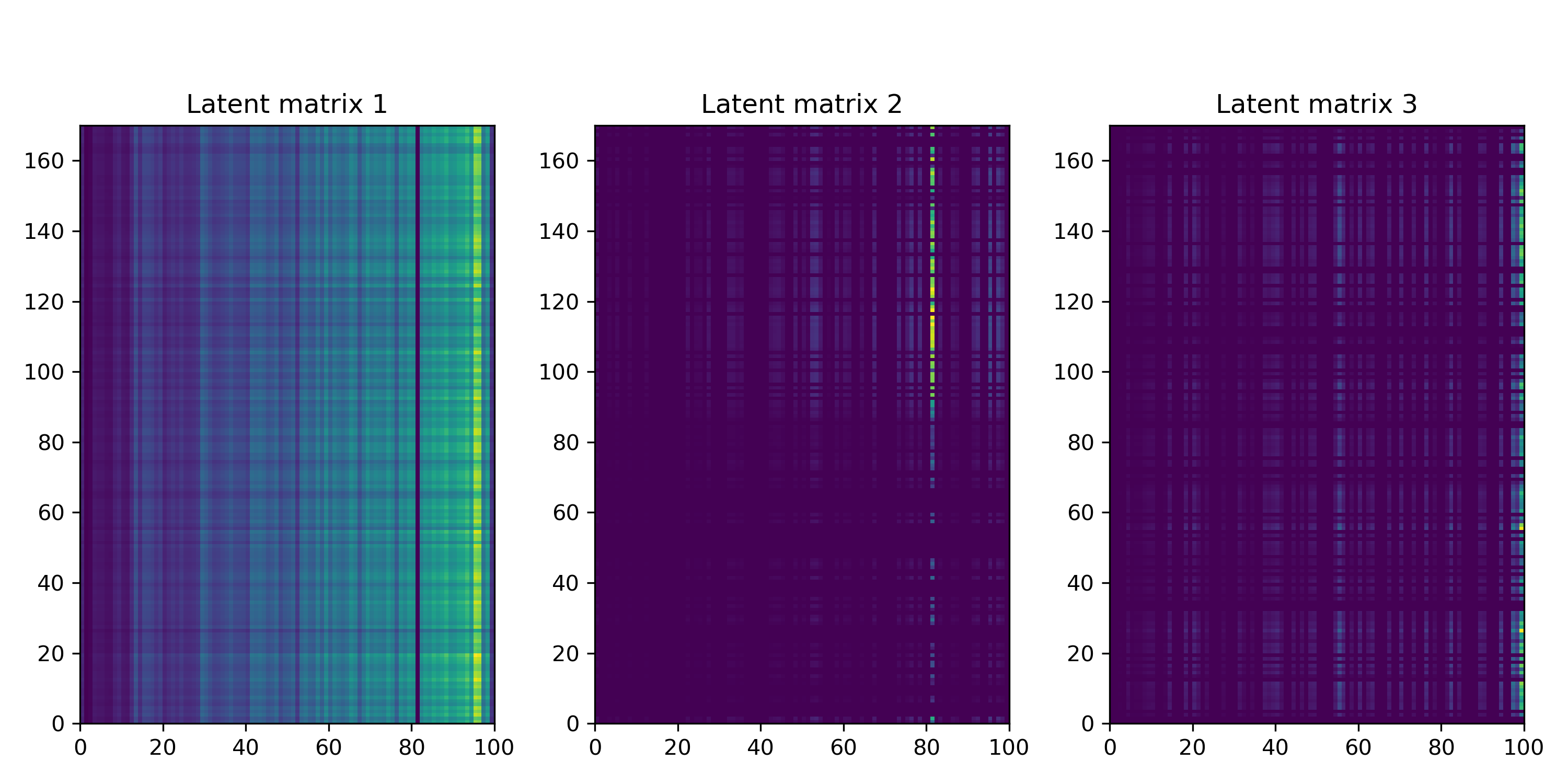}
        \caption{Latent matrices $R_{\texttt{NMF}}^{(i)}$, $i \in \{1, 3\}$.}
    \end{subfigure}
    \caption{\texttt{STMF}'s and \texttt{NMF}'s latent matrices on \texttt{AML} data.}
    \label{aml_latent_matrices}
\end{figure}

\begin{figure*}[!htb]
        \centering
        \begin{subfigure}[t]{0.32\textwidth}
            \centering
            \includegraphics[width=\textwidth]{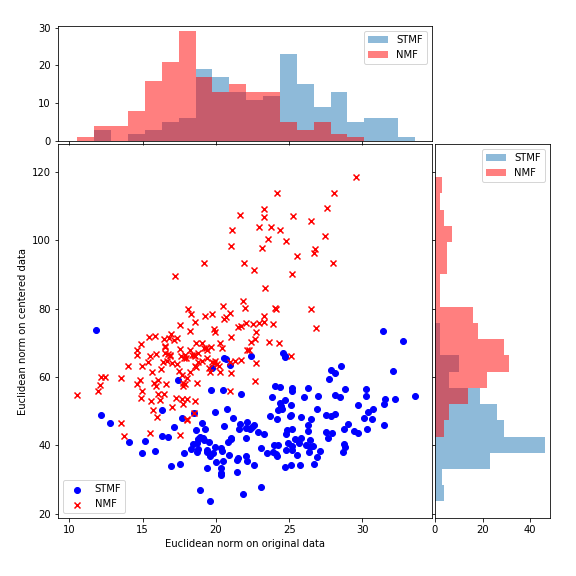}
            \caption{Euclidean norm on centered data}
        \end{subfigure}
        \hfill
        \begin{subfigure}[t]{0.32\textwidth}
            \centering 
            \includegraphics[width=\textwidth]{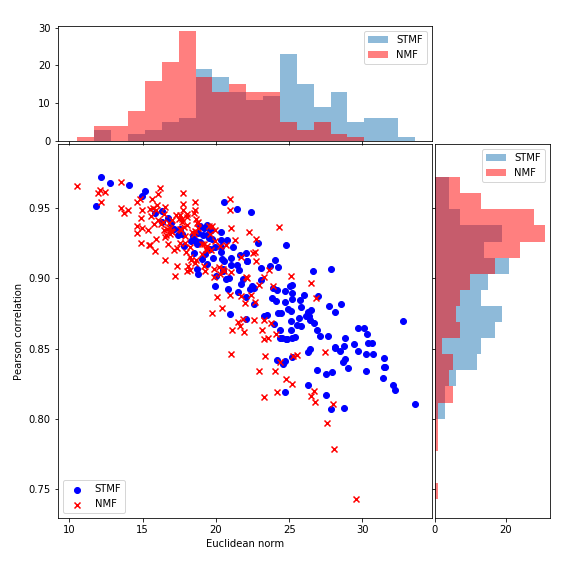}
            \caption{Pearson correlation}
        \end{subfigure}
        \hfill
        \begin{subfigure}[t]{0.32\textwidth}  
            \centering 
            \includegraphics[width=\textwidth]{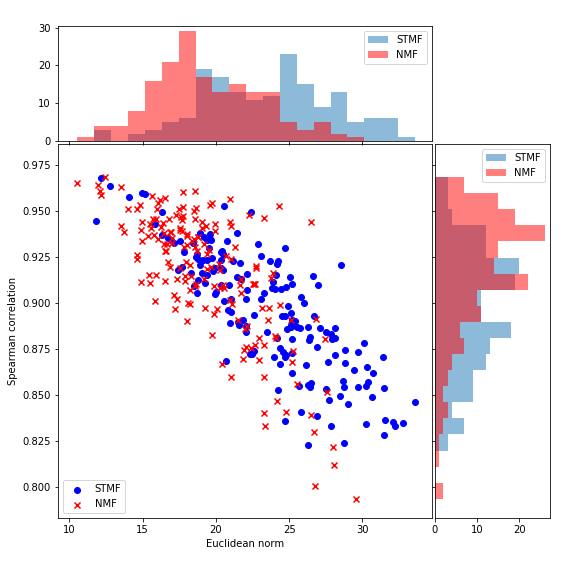}
            \caption{Spearman correlation}
        \end{subfigure}
        \caption{Euclidean norm on centered data, Pearson and Spearman correlation on \texttt{AML} data.}
        \label{aml_corrs}
\end{figure*}

\clearpage
\subsection{\texttt{COLON}}
\begin{figure}[!htb]
\captionsetup{justification=centering}
\begin{subfigure}{.66\textwidth}
\centering
\includegraphics[scale=0.27]{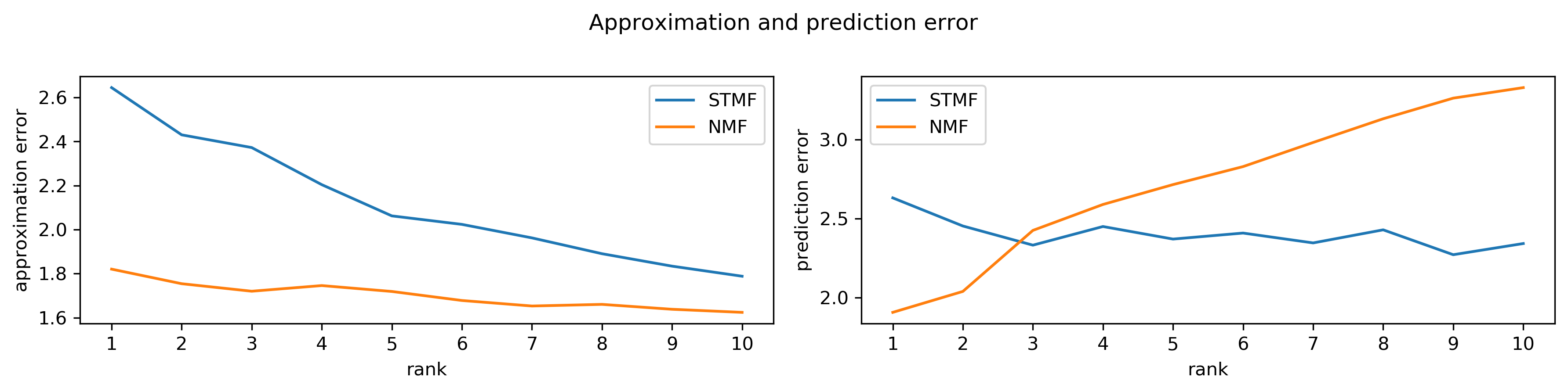}
\end{subfigure}
\begin{subfigure}{.33\textwidth}
\centering
\includegraphics[scale=0.27]{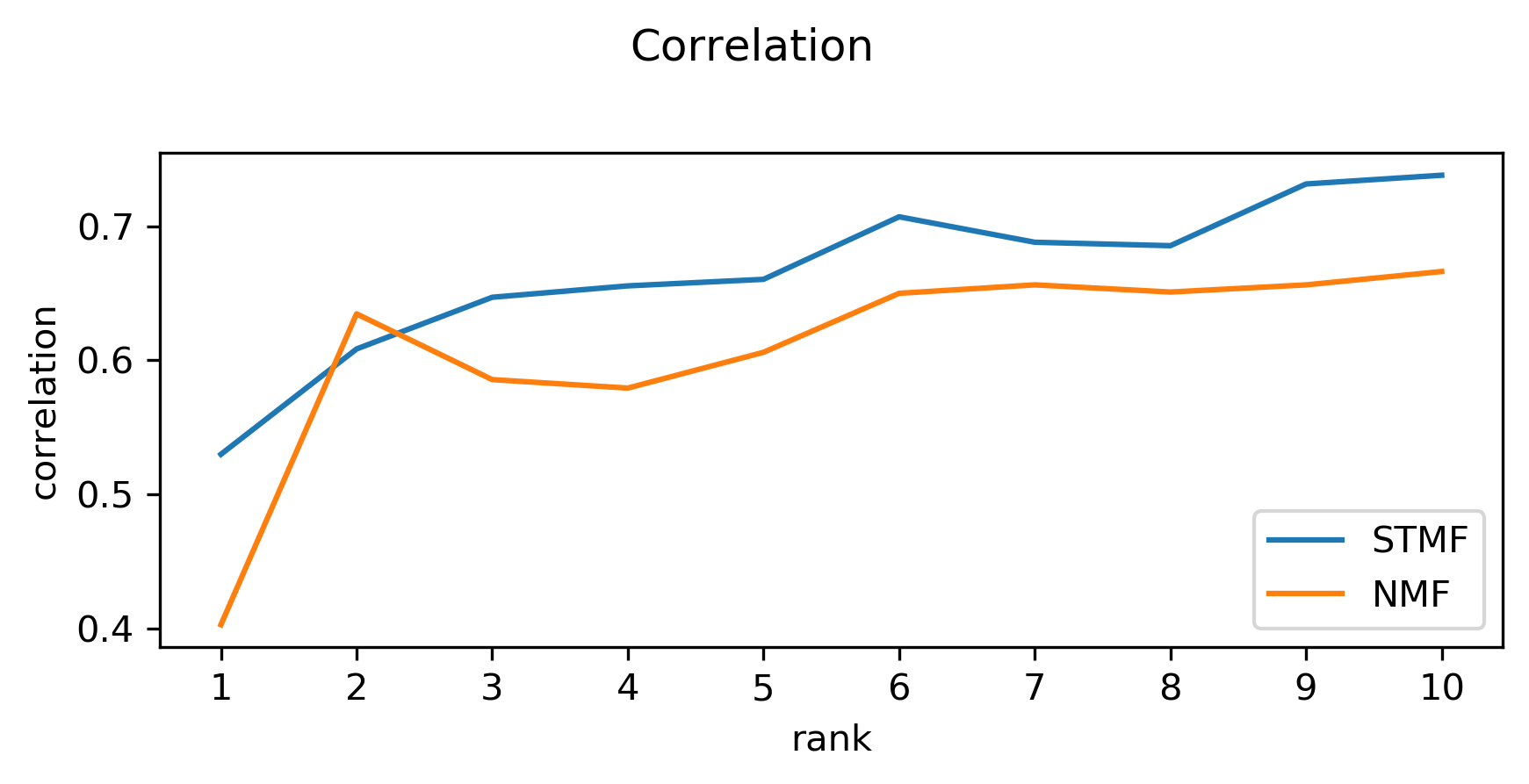}
\end{subfigure}
\caption{Difference between approximation and \\prediction RMSE and distance correlation of \texttt{STMF} \\ and \texttt{NMF} on \texttt{COLON} data.}
\label{colon_error_and_corr}
\end{figure}

\begin{figure}[!htb]
    \centering
    \captionsetup{justification=centering}
    \includegraphics[scale=0.35]{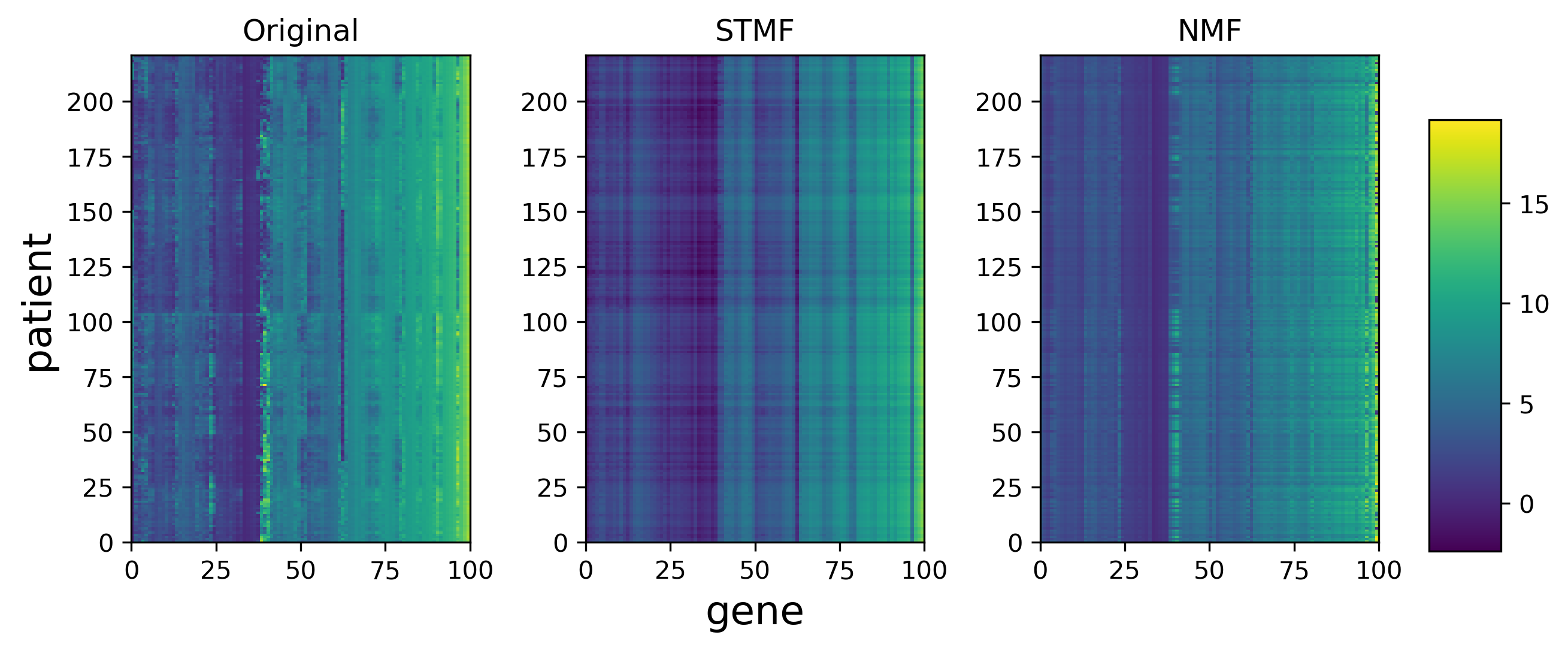}
    \caption{A comparison between \texttt{STMF}'s and \texttt{NMF}'s \\ predictions of rank 3 approximations on  \texttt{COLON} data with $20\%$ missing values.}
    \label{colon_orig_approx}
\end{figure}

\begin{figure}[!htb] 
    \centering
    \captionsetup{justification=centering}
    \begin{subfigure}{.5\textwidth}
    \includegraphics[height=0.42\textwidth, width=\textwidth]{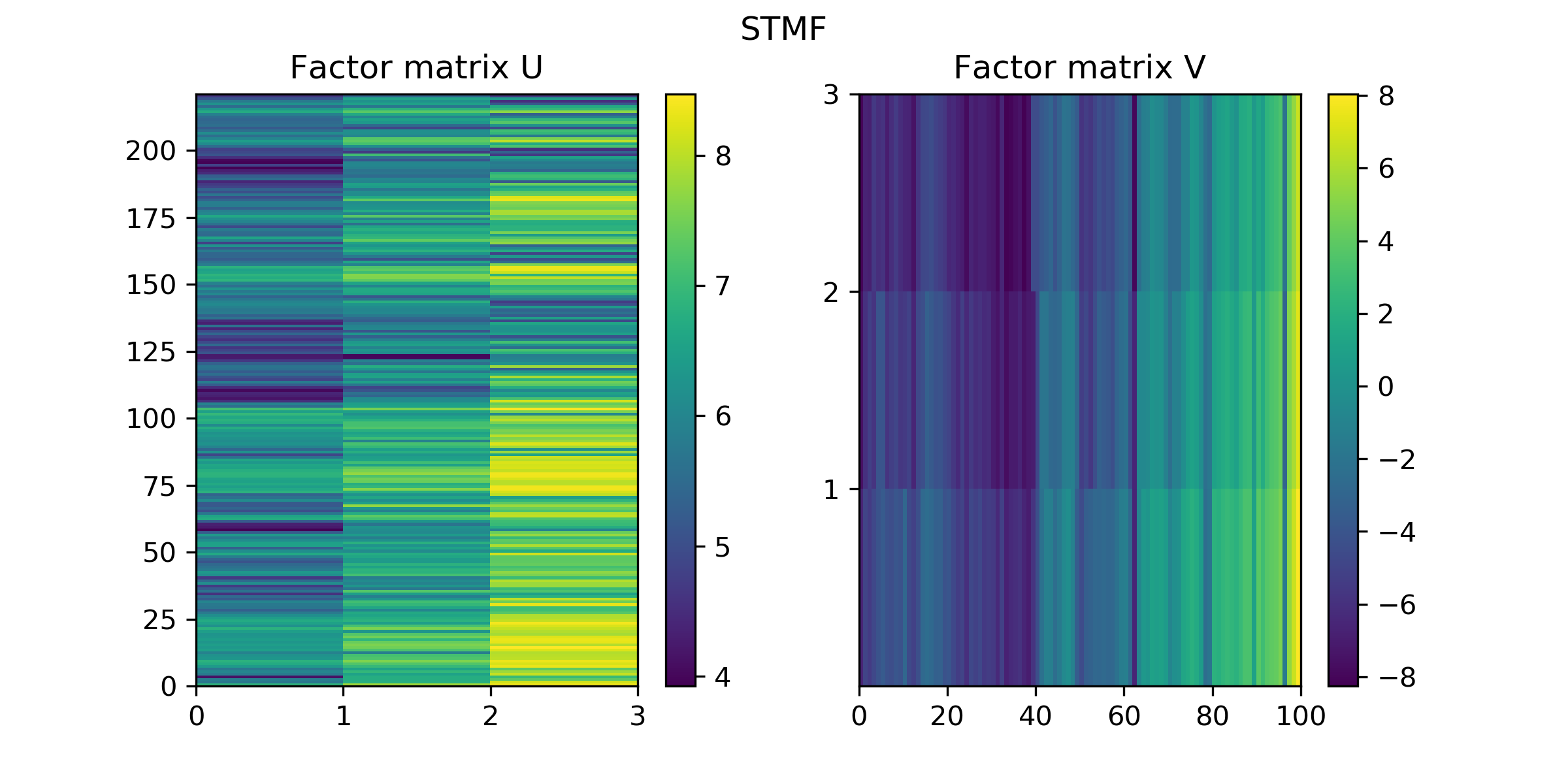}
    \caption{Factor matrices $U_{\texttt{STMF}}, V_{\texttt{STMF}}$ from \texttt{STMF}.}
\end{subfigure}\hfill%
\begin{subfigure}{.5\textwidth}
    \includegraphics[height=0.42\textwidth, width=\textwidth]{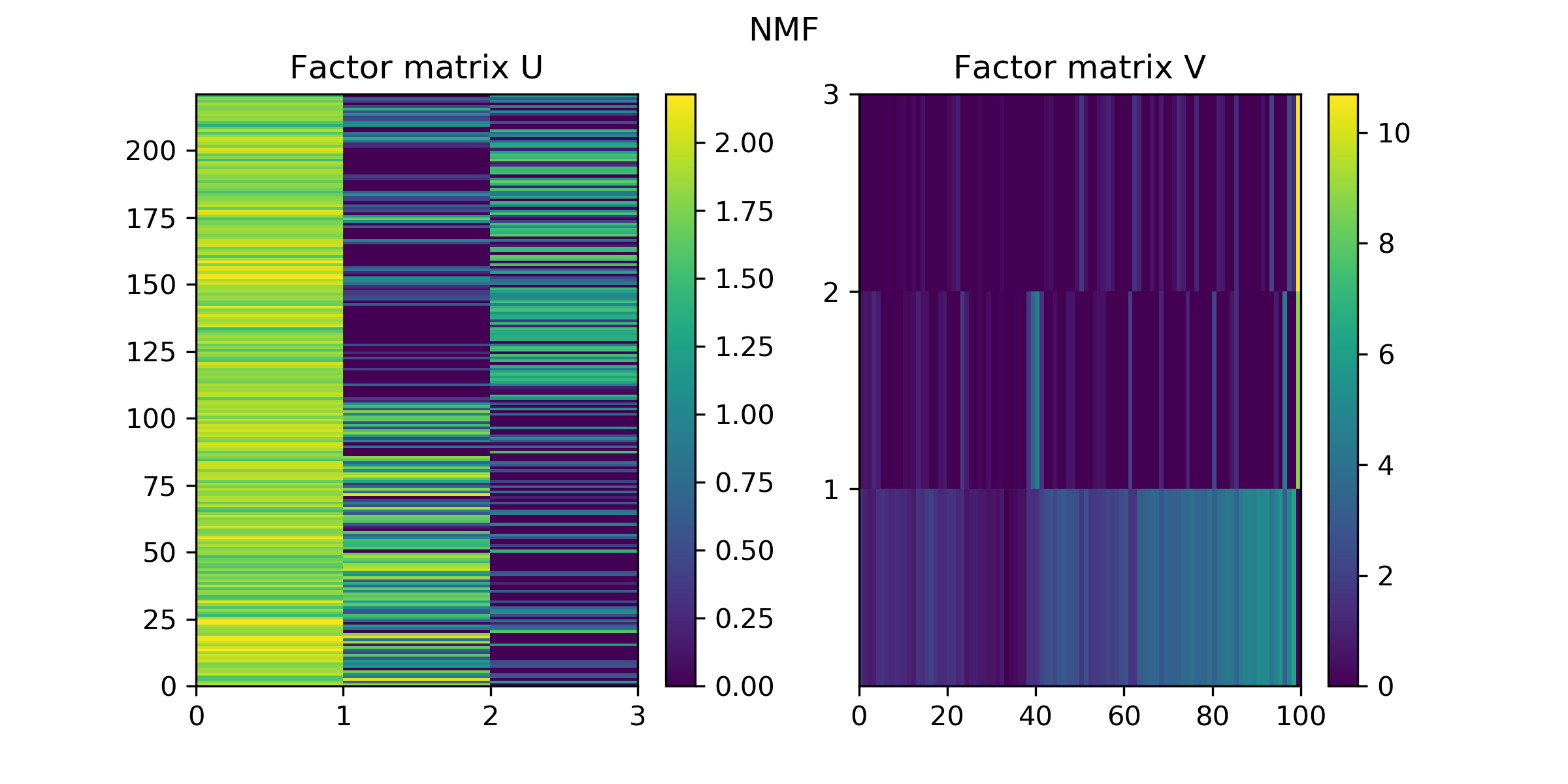}
    \caption{Factor matrices $U_{\texttt{NMF}}, V_{\texttt{NMF}}$ from \texttt{NMF}.}
\end{subfigure}
    \caption{Factor matrices $U_{\texttt{STMF}}, V_{\texttt{STMF}}$ and  $U_{\texttt{NMF}}, V_{\texttt{NMF}}$ from \texttt{STMF} and \texttt{NMF} on \texttt{COLON} data,  respectively.}
    \label{colon_factor_matrices}
\end{figure}

\begin{figure}[!htb] 
    \centering
    \begin{subfigure}[t]{.5\textwidth}
        \includegraphics[height=0.42\textwidth, width=\textwidth]{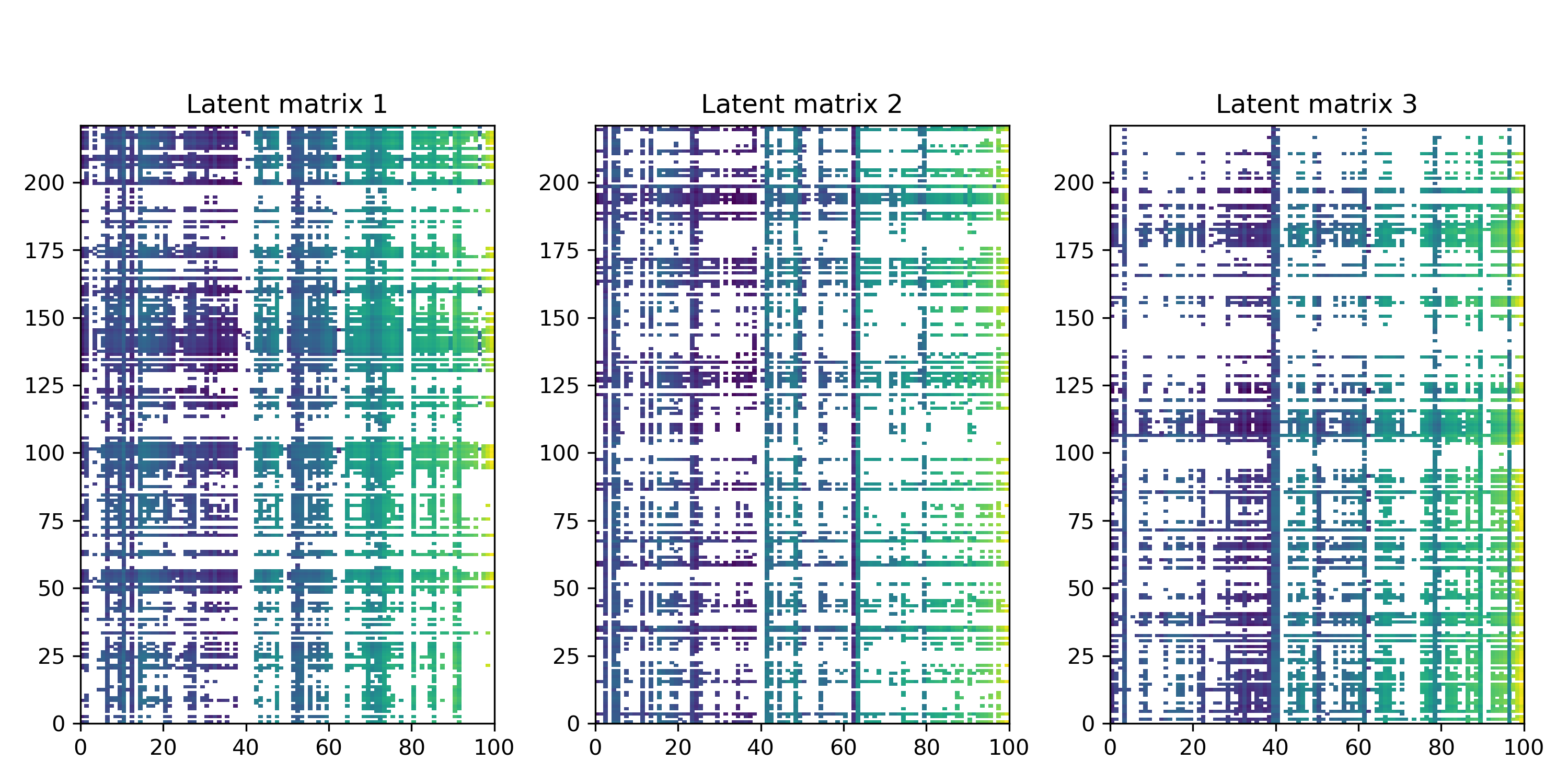}
        \caption{Latent matrices $R_{\texttt{STMF}}^{(i)}$, $i \in \{1, 3\}$, where \\ white represents the element which does not \\ contribute to the approximation $R_{\texttt{STMF}}$.}
    \end{subfigure}\hfill%
    \begin{subfigure}[t]{.5\textwidth}
        \includegraphics[height=0.42\textwidth, width=\textwidth]{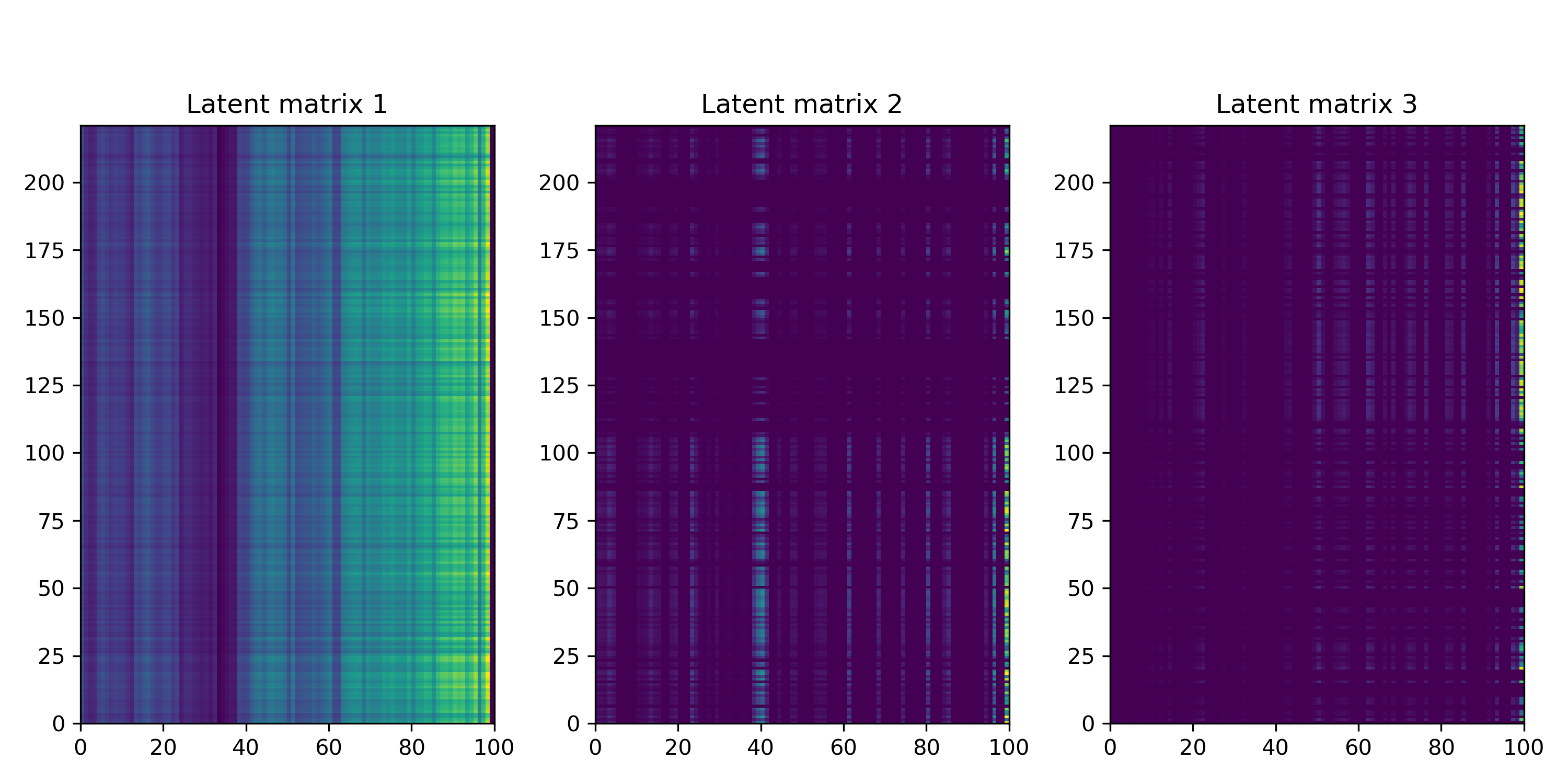}
        \caption{Latent matrices $R_{\texttt{NMF}}^{(i)}$, $i \in \{1, 3\}$.}
    \end{subfigure}
    \caption{\texttt{STMF}'s and \texttt{NMF}'s latent matrices on \texttt{COLON} data.}
    \label{colon_latent_matrices}
\end{figure}

\begin{figure*}[!htb]
        \centering
        \begin{subfigure}[t]{0.32\textwidth}
            \centering
            \includegraphics[width=\textwidth]{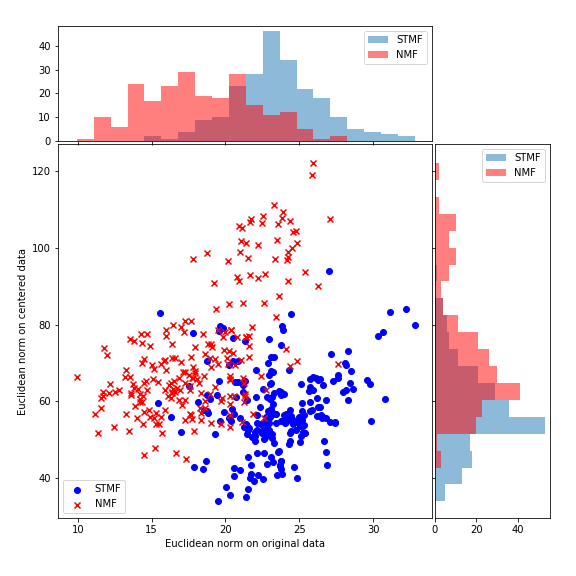}
            \caption{Euclidean norm on centered data}
        \end{subfigure}
        \hfill
        \begin{subfigure}[t]{0.32\textwidth}  
            \centering 
            \includegraphics[width=\textwidth]{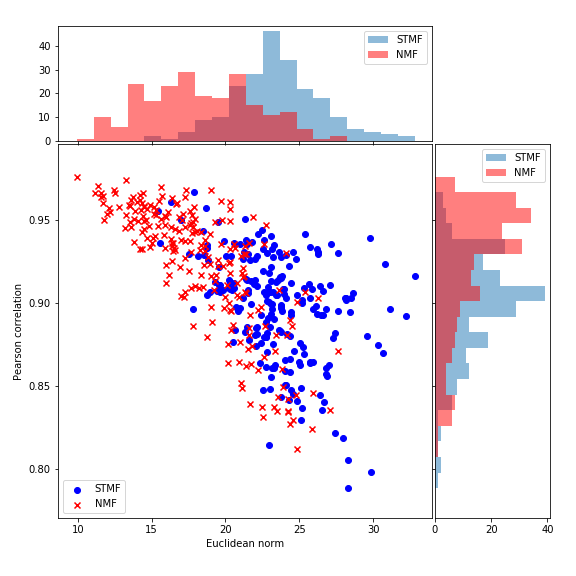}
            \caption{Pearson correlation}
        \end{subfigure}
        \hfill
        \begin{subfigure}[t]{0.32\textwidth}
            \centering 
            \includegraphics[width=\textwidth]{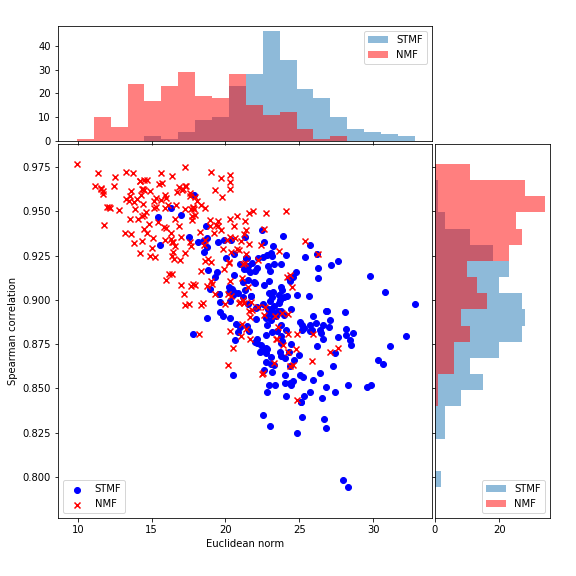}
            \caption{Spearman correlation}
        \end{subfigure}
        \caption{Euclidean norm on centered data, Pearson and Spearman correlation on \texttt{COLON} data.}
        \label{colon_corrs}
\end{figure*}

\clearpage
\subsection{\texttt{GBM}}
\begin{figure}[!htb]
\captionsetup{justification=centering}
\begin{subfigure}{.66\textwidth}
\centering
\includegraphics[scale=0.27]{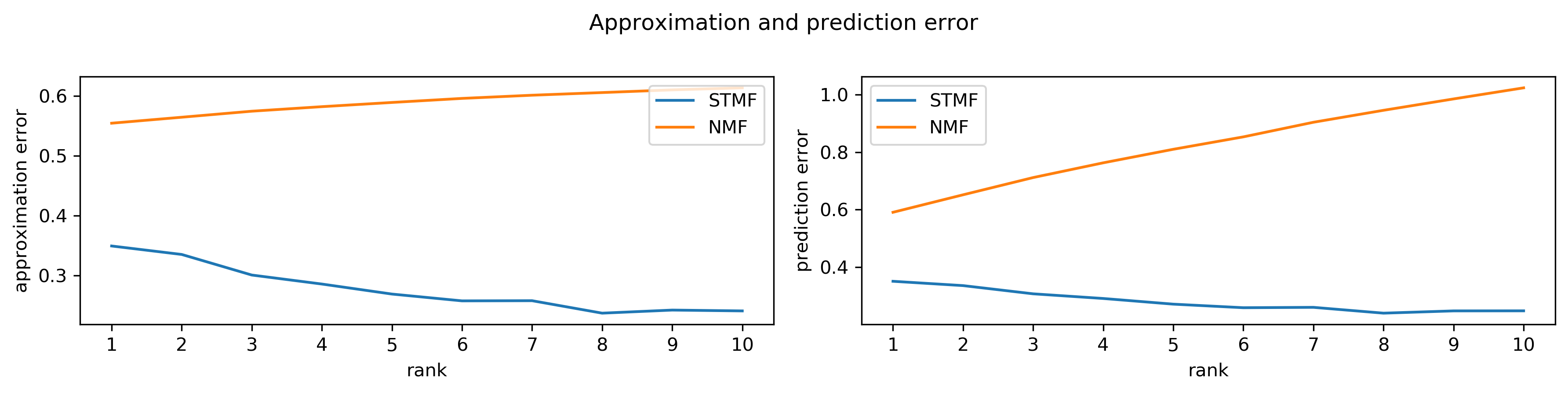}
\end{subfigure}
\begin{subfigure}{.33\textwidth}
\centering
\includegraphics[scale=0.27]{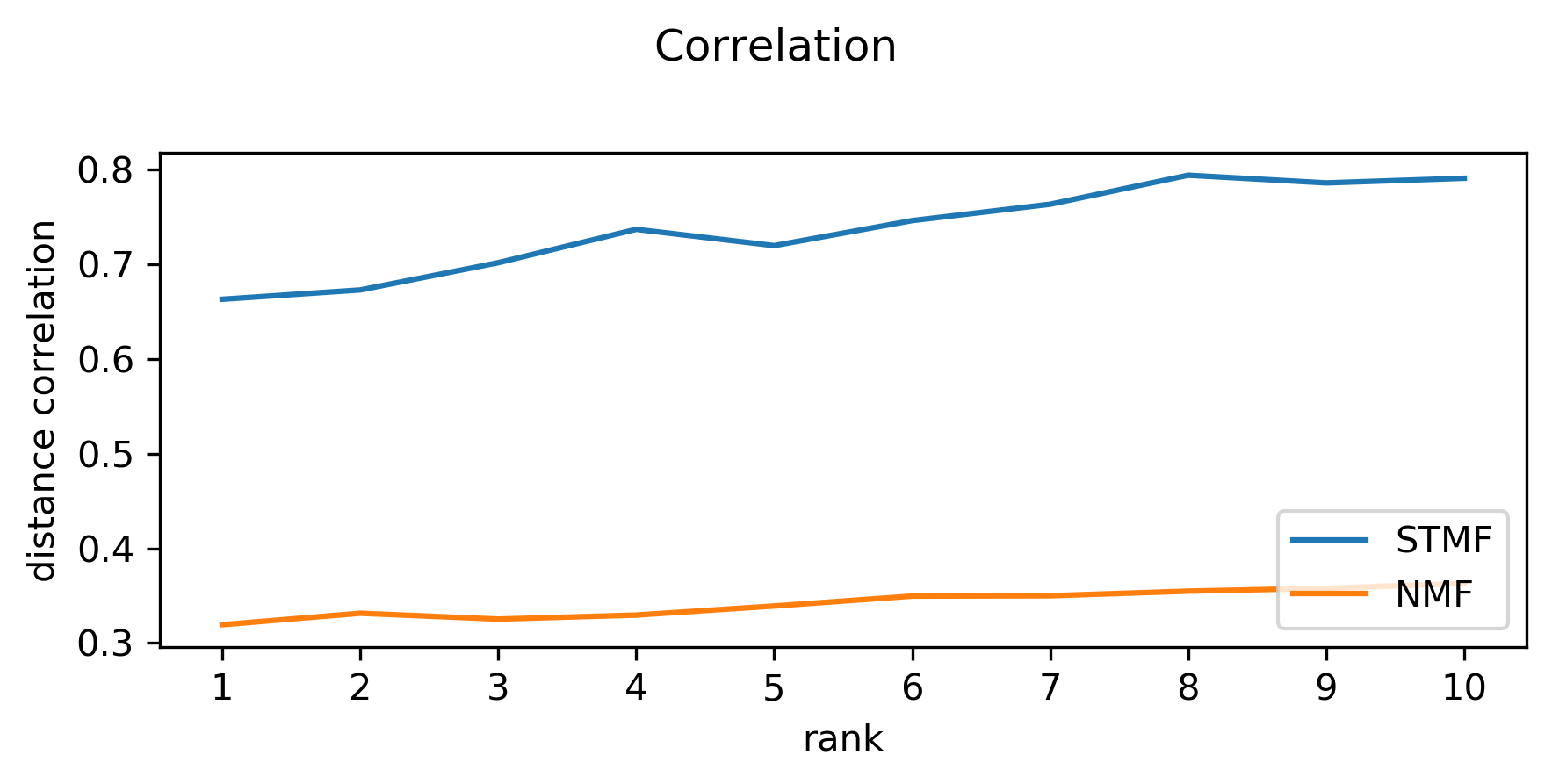}
\end{subfigure}
\caption{Difference between approximation and \\prediction RMSE and distance correlation of \texttt{STMF} \\ and \texttt{NMF} on \texttt{GBM} data.}
\label{gbm_error_and_corr}
\end{figure}

\begin{figure}[!htb]
    \centering
    \captionsetup{justification=centering}
    \includegraphics[scale=0.35]{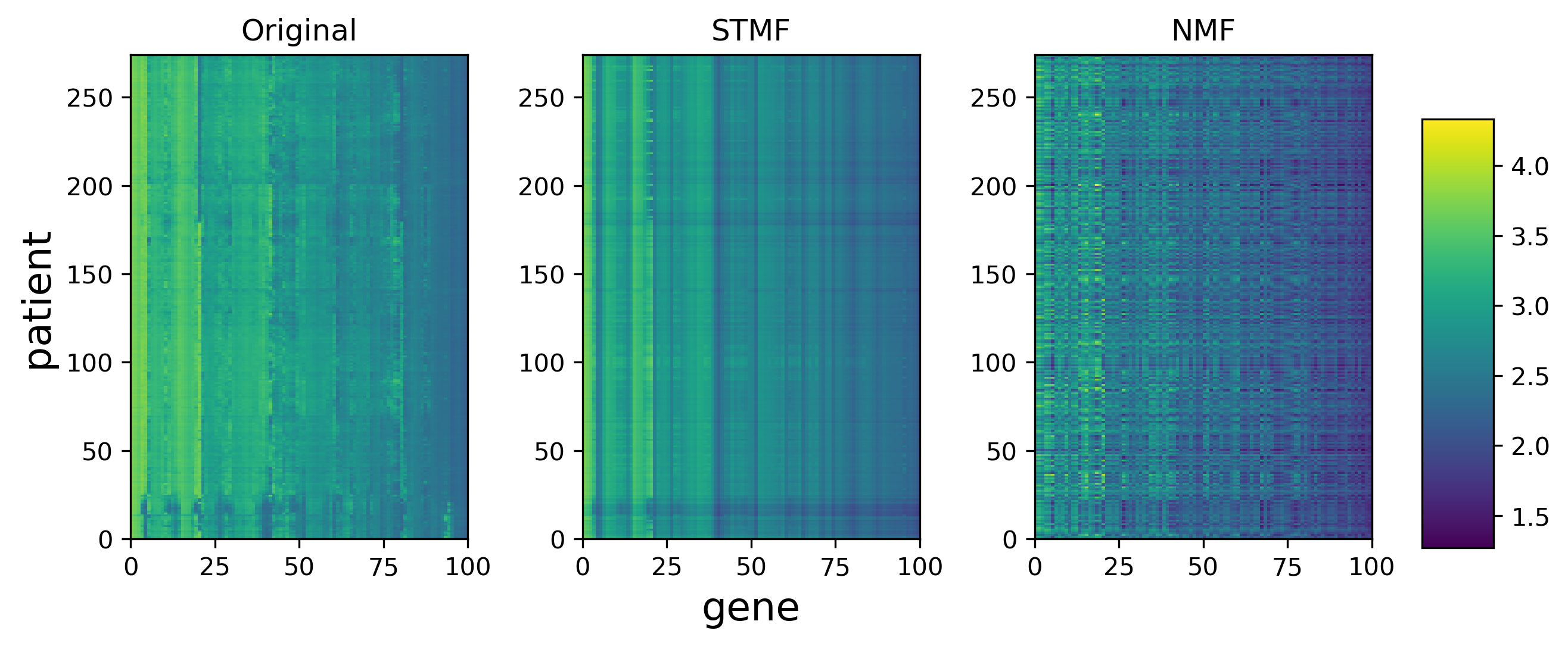}
    \caption{A comparison between \texttt{STMF}'s and \texttt{NMF}'s \\ predictions of rank 3 approximations on  \texttt{GBM} data with $20\%$ missing values.}
    \label{synth_orig_miss_approx}
\end{figure}

\begin{figure}[!htb] 
    \centering
    \captionsetup{justification=centering}
    \begin{subfigure}{.5\textwidth}
    \includegraphics[height=0.42\textwidth, width=\textwidth]{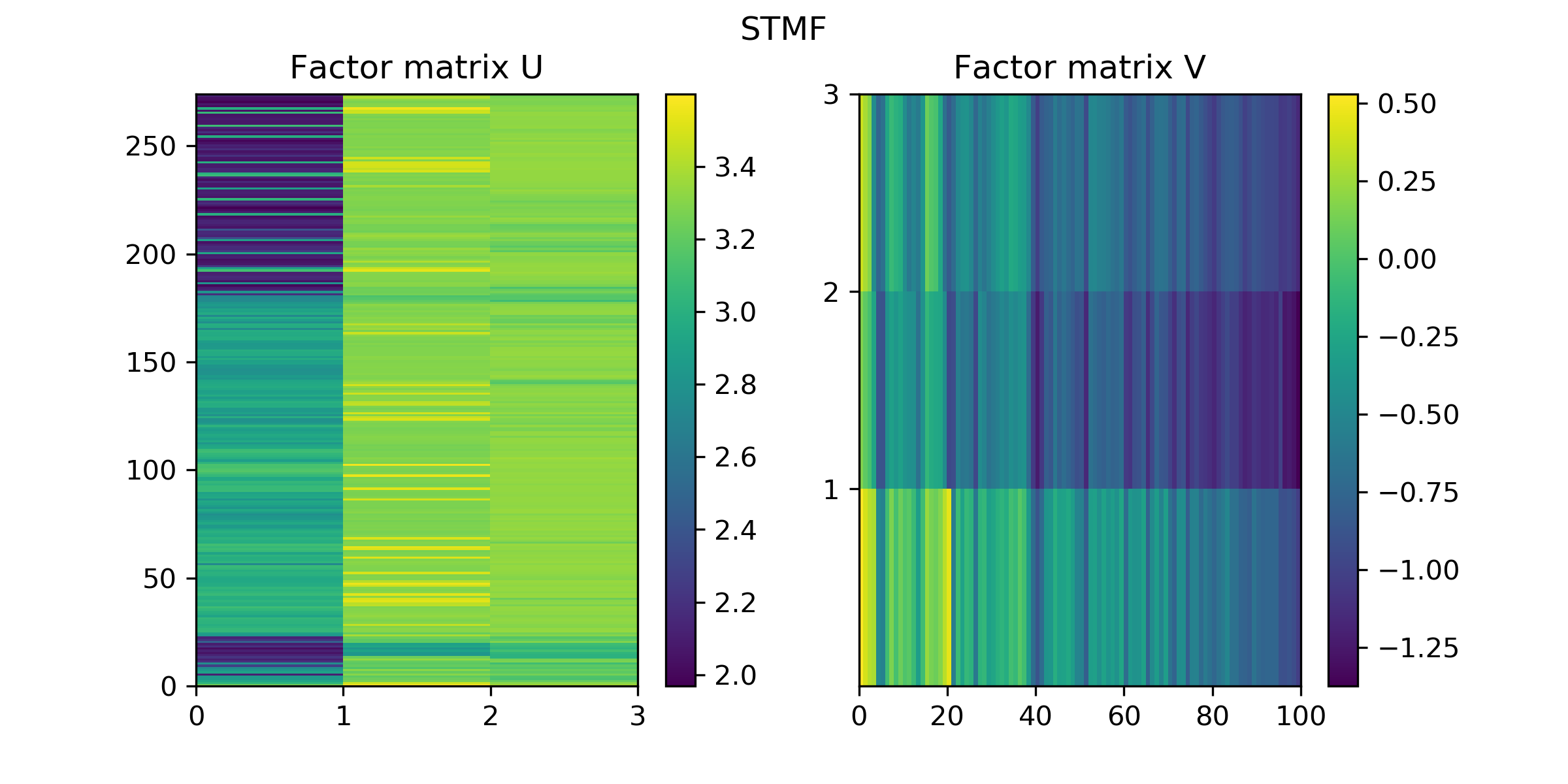}
    \caption{Factor matrices $U_{\texttt{STMF}}, V_{\texttt{STMF}}$ from \texttt{STMF}.}
\end{subfigure}\hfill%
\begin{subfigure}{.5\textwidth}
    \includegraphics[height=0.42\textwidth, width=\textwidth]{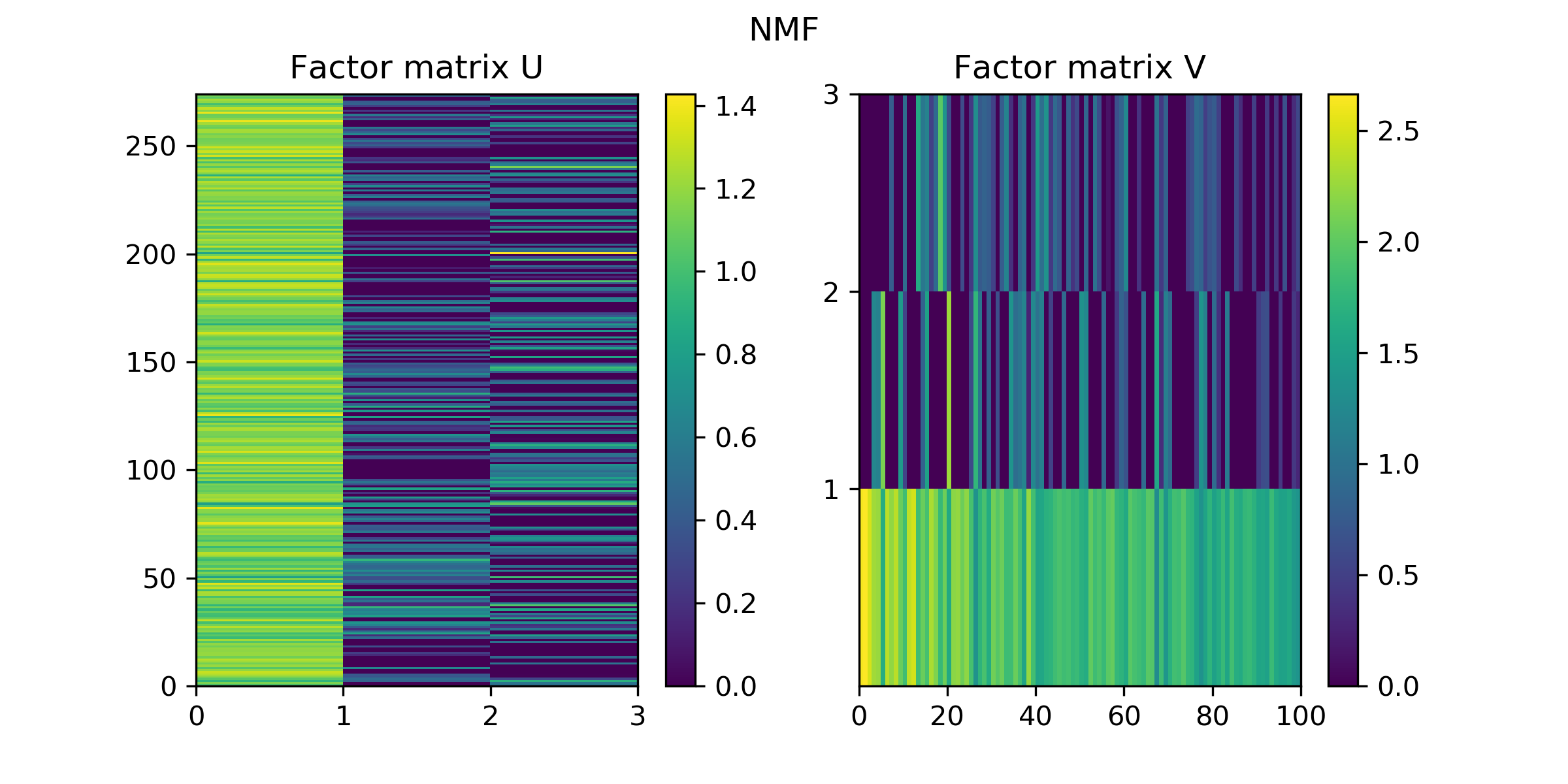}
    \caption{Factor matrices $U_{\texttt{NMF}}, V_{\texttt{NMF}}$ from \texttt{NMF}.}
\end{subfigure}
    \caption{Factor matrices $U_{\texttt{STMF}}, V_{\texttt{STMF}}$ and  $U_{\texttt{NMF}}, V_{\texttt{NMF}}$ from \texttt{STMF} and \texttt{NMF} on \texttt{GBM} data,  respectively.}
    \label{gbm_factor_matrices}
\end{figure}

\begin{figure}[!htb] 
    \centering
    \begin{subfigure}[t]{.5\textwidth}
        \includegraphics[height=0.42\textwidth, width=\textwidth]{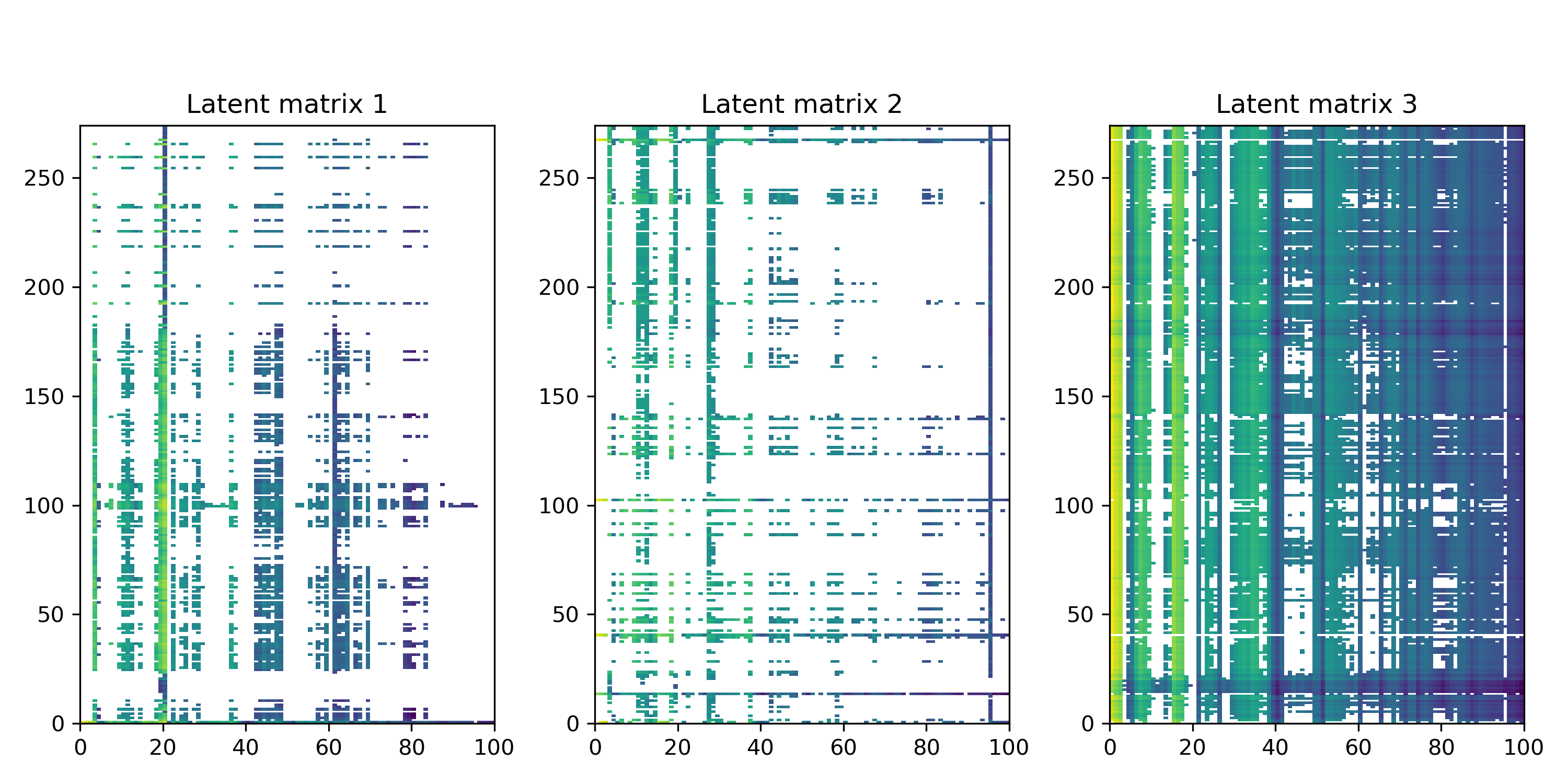}
        \caption{Latent matrices $R_{\texttt{STMF}}^{(i)}$, $i \in \{1, 3\}$, where \\ white represents the element which does not \\ contribute to the approximation $R_{\texttt{STMF}}$.}
    \end{subfigure}\hfill%
    \begin{subfigure}[t]{.5\textwidth}
        \includegraphics[height=0.42\textwidth, width=\textwidth]{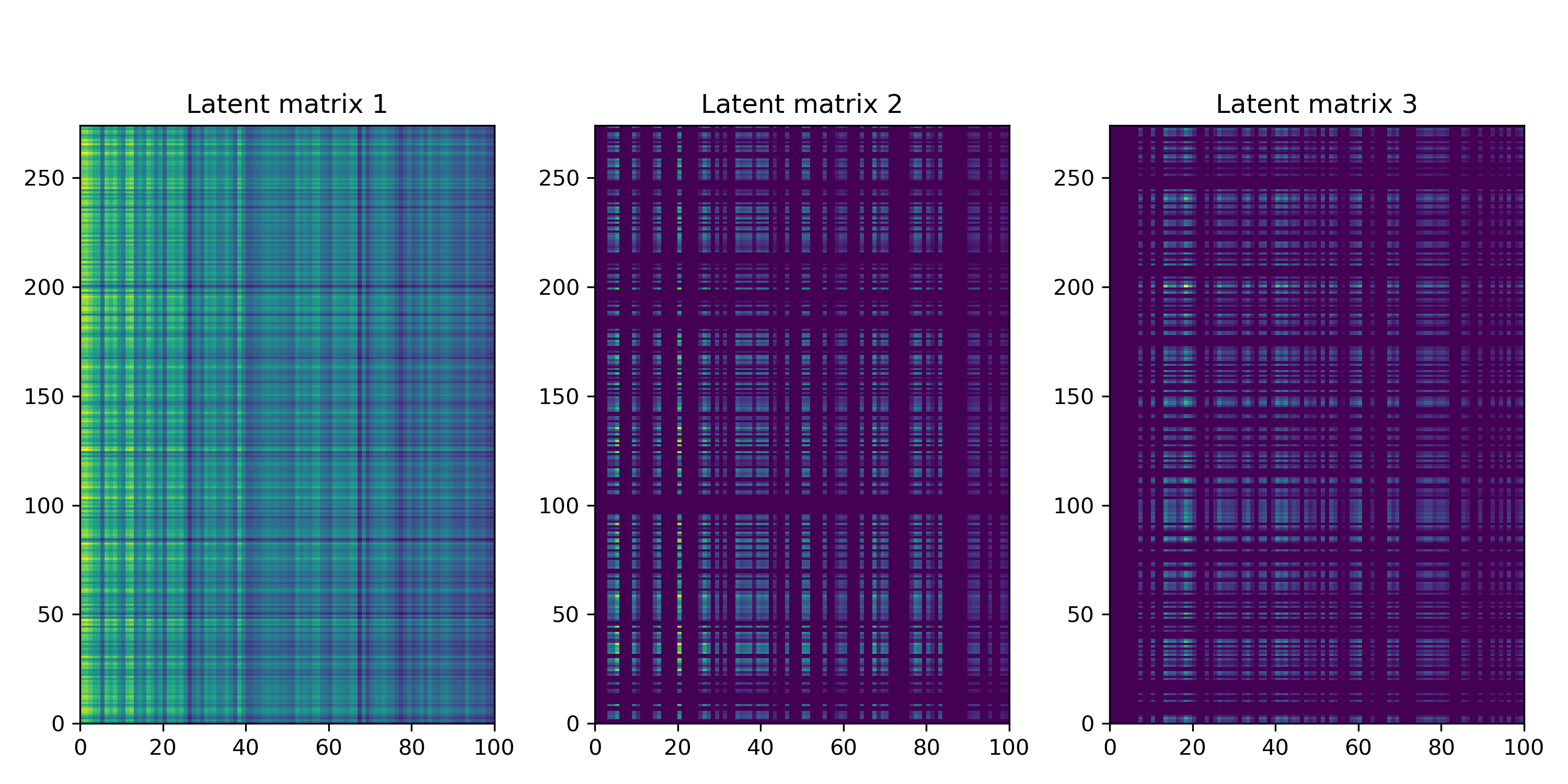}
        \caption{Latent matrices $R_{\texttt{NMF}}^{(i)}$, $i \in \{1, 3\}$.}
    \end{subfigure}
    \caption{\texttt{STMF}'s and \texttt{NMF}'s latent matrices on \texttt{GBM} data.}
    \label{gbm_latent_matrices}
\end{figure}

\begin{figure*}[!htb]
        \centering
        \begin{subfigure}[t]{0.32\textwidth}
            \centering
            \includegraphics[width=\textwidth]{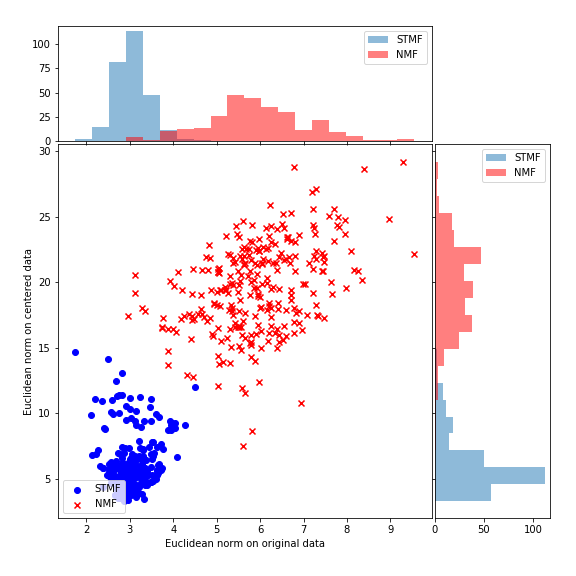}
            \caption{Euclidean norm on centered data}
        \end{subfigure}
        \hfill
        \begin{subfigure}[t]{0.32\textwidth}  
            \centering 
            \includegraphics[width=\textwidth]{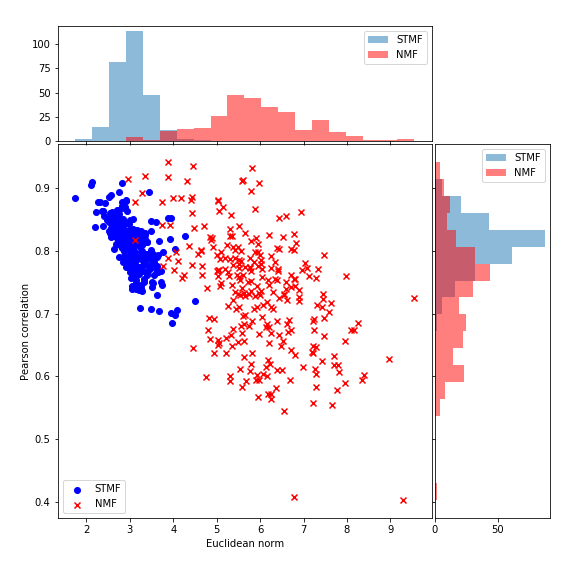}
            \caption{Pearson correlation}
        \end{subfigure}
        \hfill
        \begin{subfigure}[t]{0.32\textwidth}  
            \centering 
            \includegraphics[width=\textwidth]{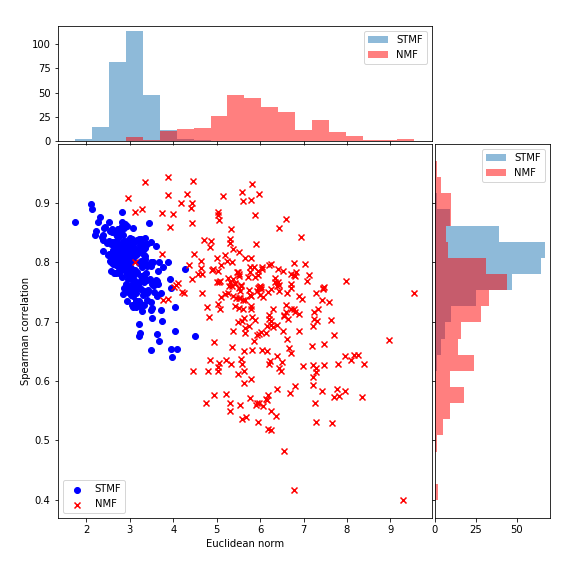}
            \caption{Spearman correlation}
        \end{subfigure}
        \caption{Euclidean norm on centered data, Pearson and Spearman correlation on \texttt{GBM} data.}
        \label{gbm_corrs}
\end{figure*}

\clearpage
\subsection{\texttt{LIHC}}
\begin{figure}[!htb]
\captionsetup{justification=centering}
\begin{subfigure}{.66\textwidth}
\centering
\includegraphics[scale=0.27]{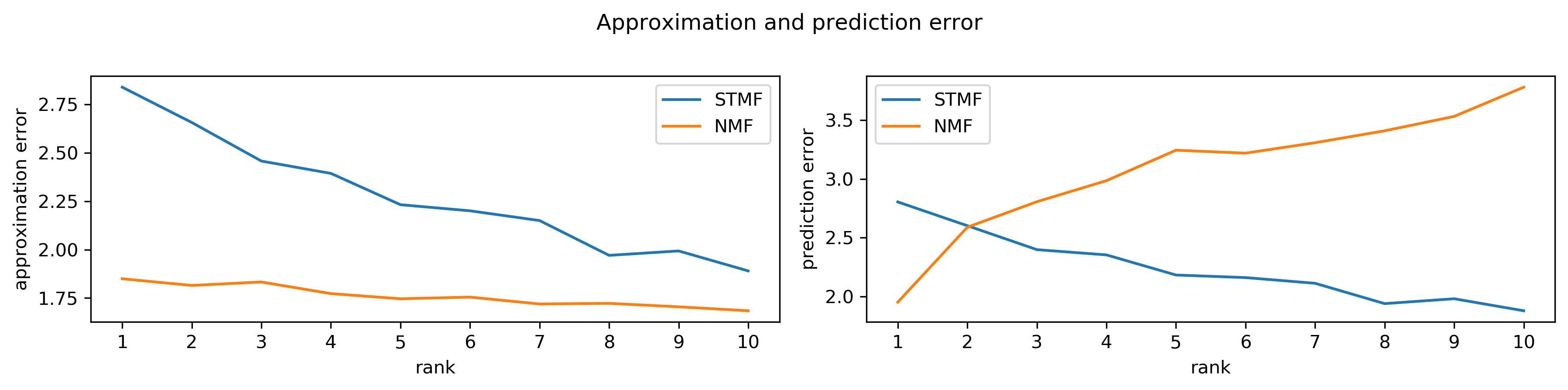}
\end{subfigure}
\begin{subfigure}{.33\textwidth}
\centering
\includegraphics[scale=0.27]{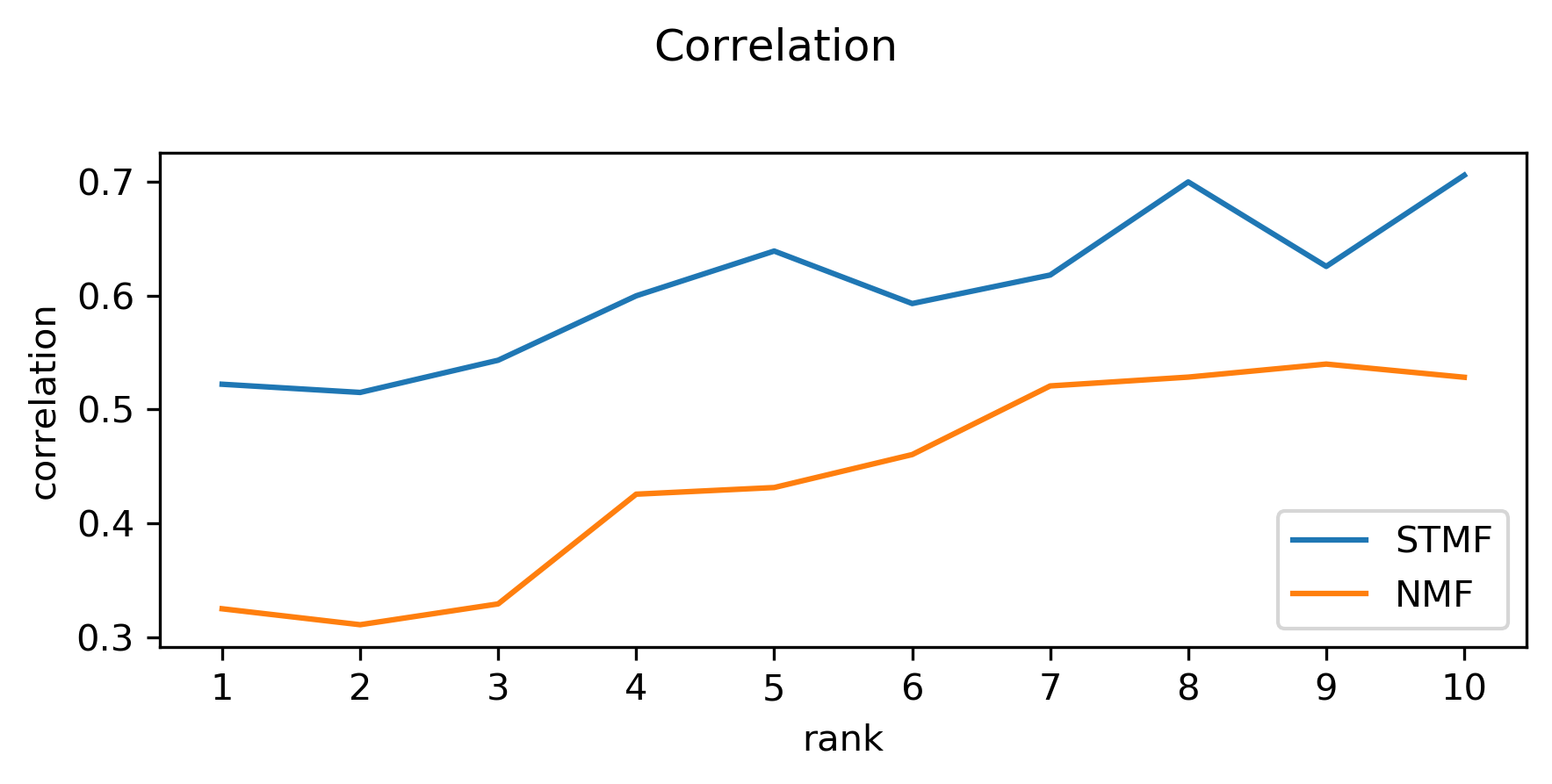}
\end{subfigure}
\caption{Difference between approximation and \\prediction RMSE and distance correlation of \texttt{STMF} \\ and \texttt{NMF} on \texttt{LIHC} data.}
\label{lihc_error_and_corr}
\end{figure}

\begin{figure}[!htb]
    \centering
    \captionsetup{justification=centering}
    \includegraphics[scale=0.35]{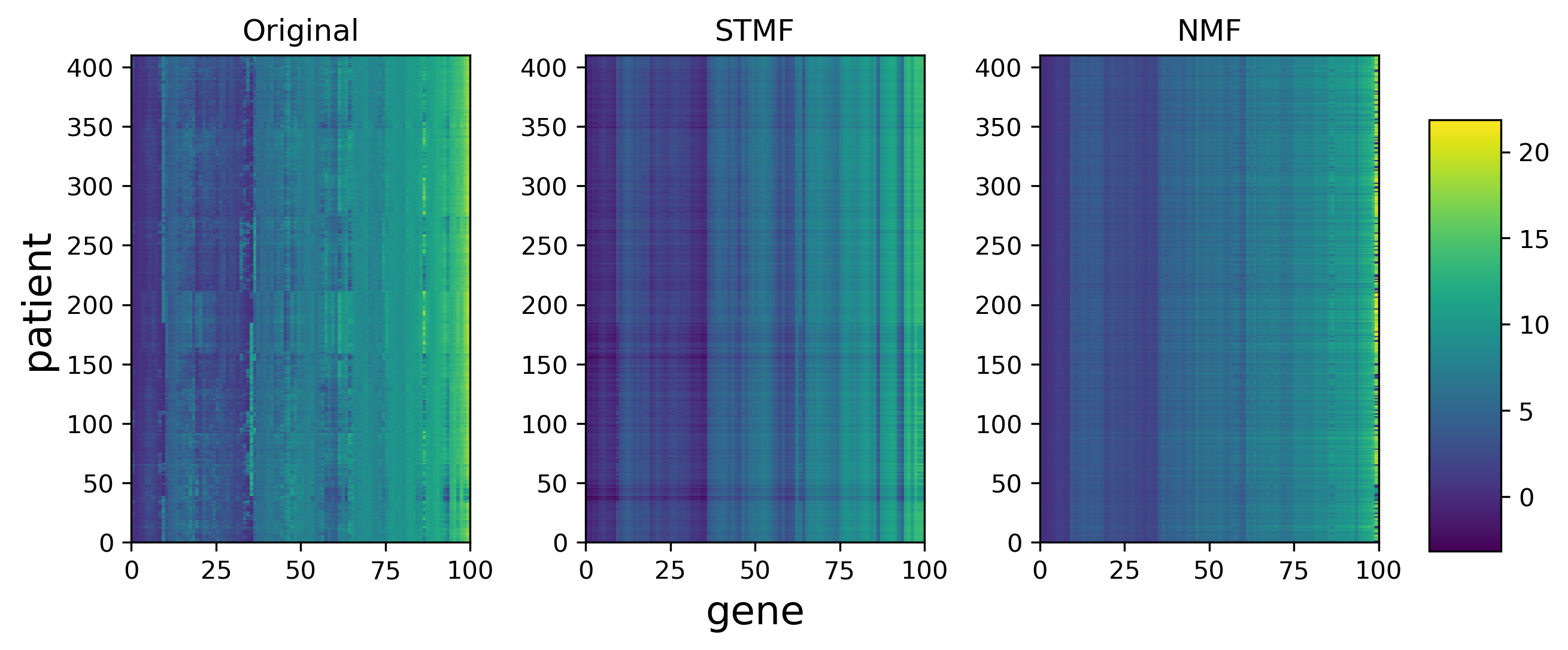}
    \caption{A comparison between \texttt{STMF}'s and \texttt{NMF}'s \\ predictions of rank 2 approximations on  \texttt{LIHC} data with $20\%$ missing values.}
    \label{lihc_orig_approx}
\end{figure}

\begin{figure}[!htb] 
    \centering
    \captionsetup{justification=centering}
    \begin{subfigure}{.5\textwidth}
    \includegraphics[height=0.42\textwidth, width=\textwidth]{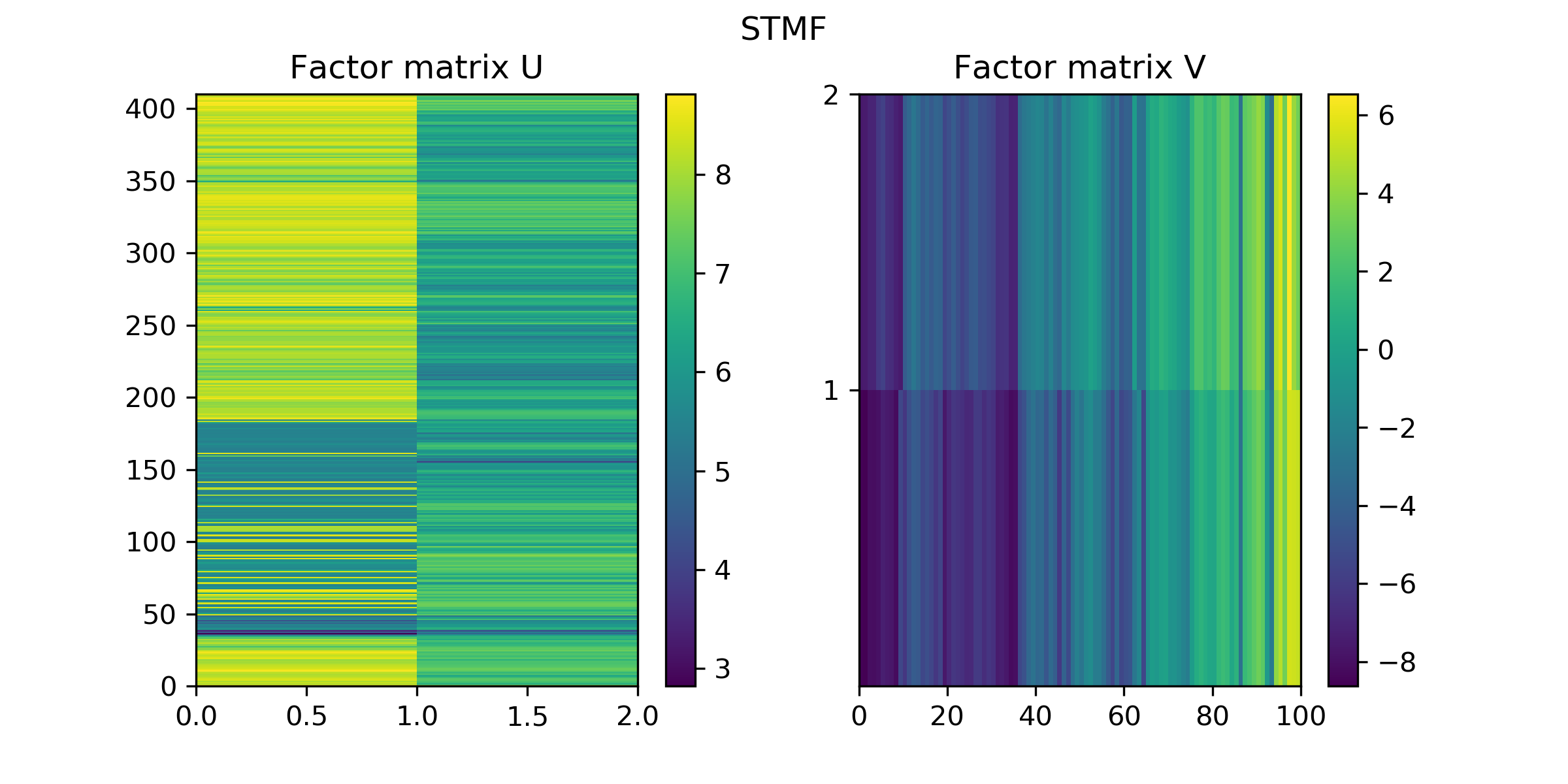}
    \caption{Factor matrices $U_{\texttt{STMF}}, V_{\texttt{STMF}}$ from \texttt{STMF}.}
\end{subfigure}\hfill%
\begin{subfigure}{.5\textwidth}
    \includegraphics[height=0.42\textwidth, width=\textwidth]{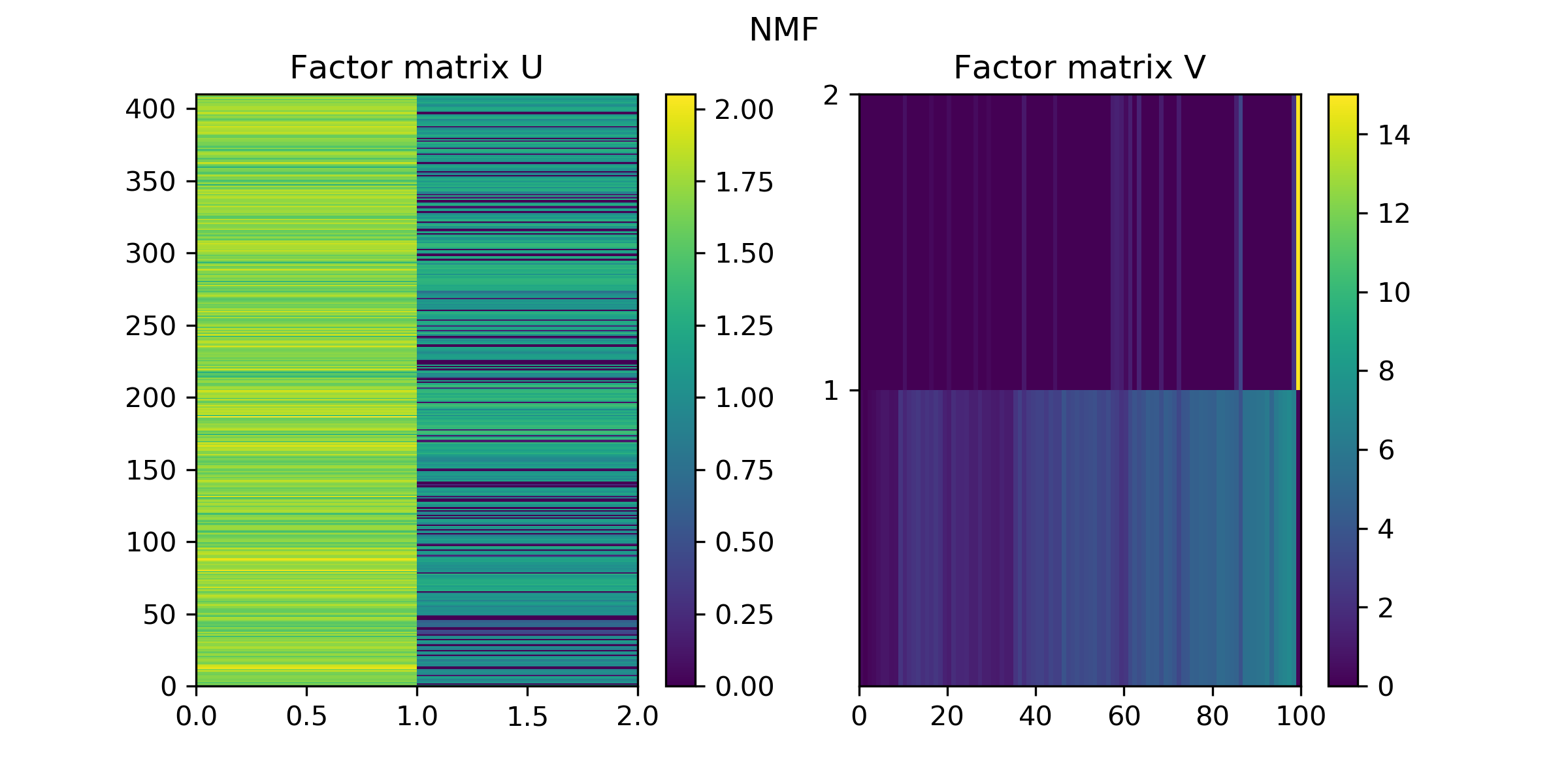}
    \caption{Factor matrices $U_{\texttt{NMF}}, V_{\texttt{NMF}}$ from \texttt{NMF}.}
\end{subfigure}
    \caption{Factor matrices $U_{\texttt{STMF}}, V_{\texttt{STMF}}$ and  $U_{\texttt{NMF}}, V_{\texttt{NMF}}$ from \texttt{STMF} and \texttt{NMF} on \texttt{LIHC} data,  respectively.}
    \label{lihc_factor_matrices}
\end{figure}

\begin{figure}[!htb] 
    \centering
    \begin{subfigure}[t]{.5\textwidth}
        \includegraphics[height=0.42\textwidth, width=\textwidth]{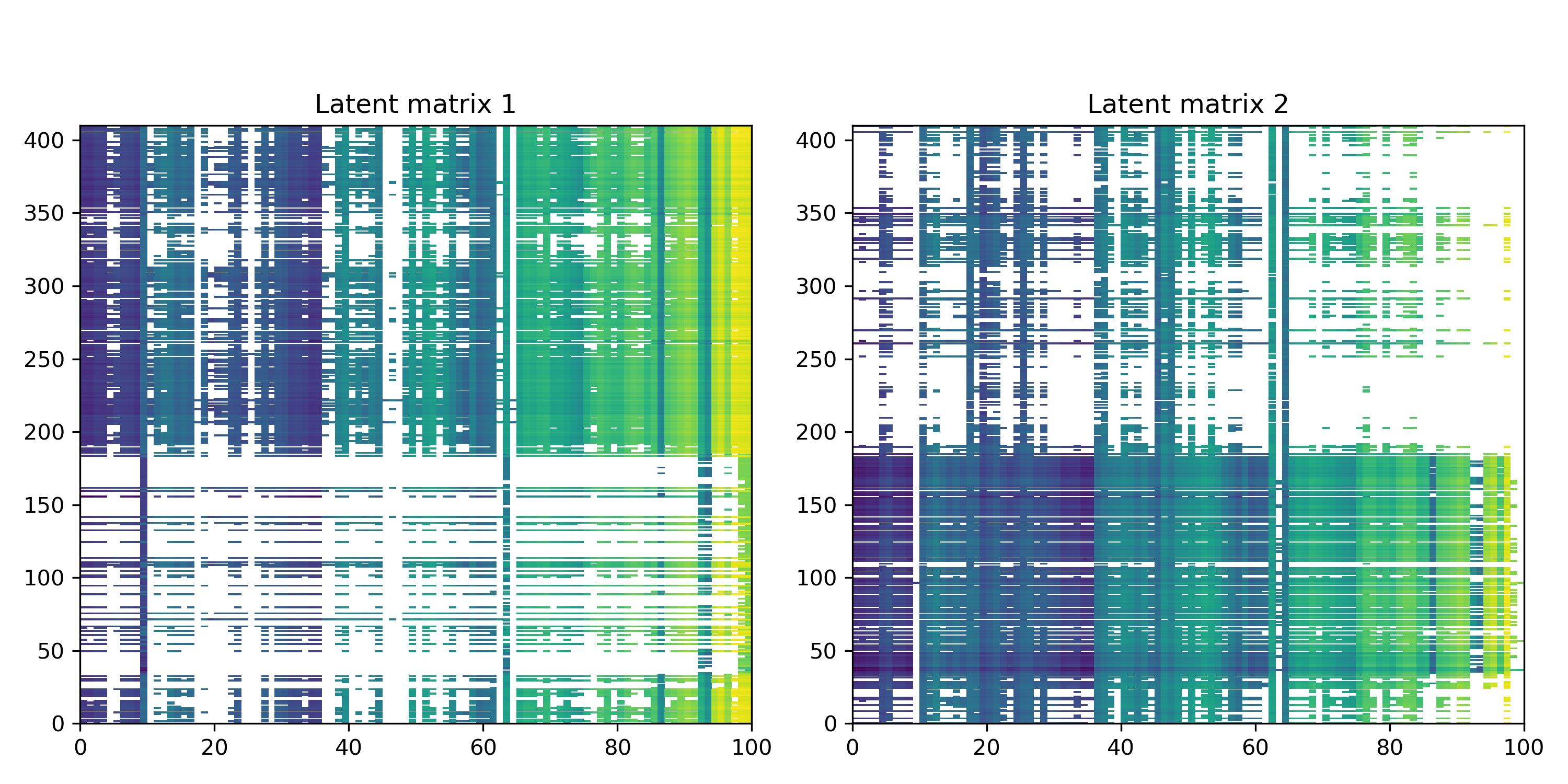}
        \caption{Latent matrices $R_{\texttt{STMF}}^{(i)}$, $i \in \{1, 2\}$, where \\ white represents the element which does not \\ contribute to the approximation $R_{\texttt{STMF}}$.}
    \end{subfigure}\hfill%
    \begin{subfigure}[t]{.5\textwidth}
        \includegraphics[height=0.42\textwidth, width=\textwidth]{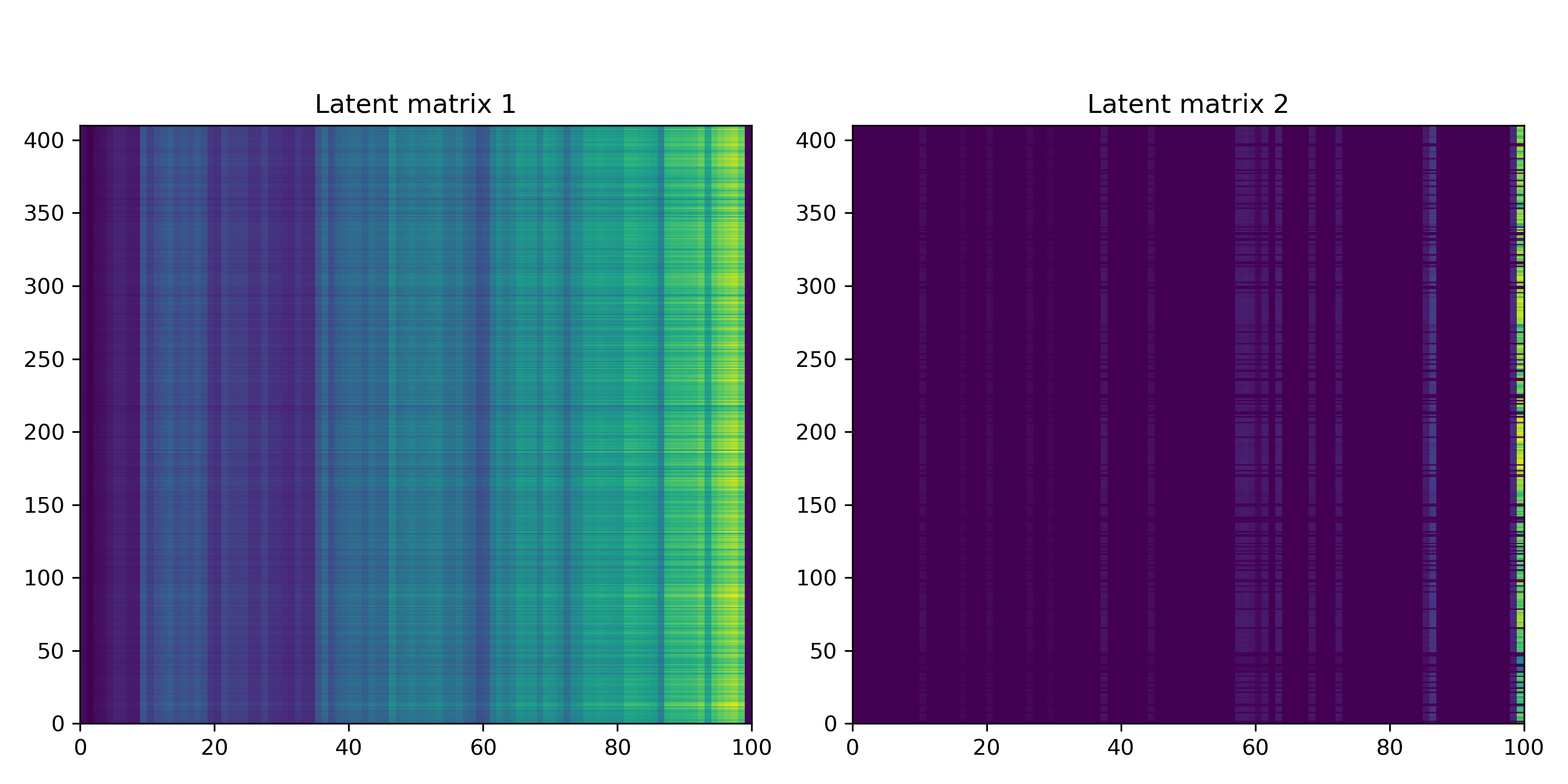}
        \caption{Latent matrices $R_{\texttt{NMF}}^{(i)}$, $i \in \{1, 2\}$.}
    \end{subfigure}
    \caption{\texttt{STMF}'s and \texttt{NMF}'s latent matrices on \texttt{LIHC} data.}
    \label{lihc_latent_matrices}
\end{figure}

\begin{figure*}[!htb]
        \centering
        \begin{subfigure}[t]{0.32\textwidth}
            \centering
            \includegraphics[width=\textwidth]{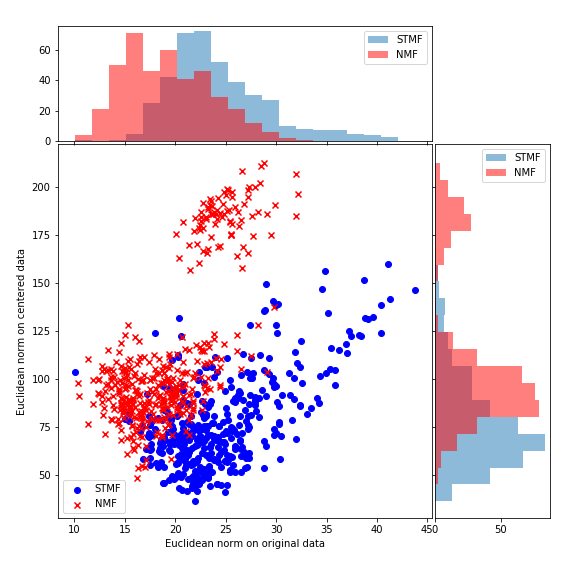}
            \caption{Euclidean norm on centered data}
        \end{subfigure}
        \hfill
        \begin{subfigure}[t]{0.32\textwidth}
            \centering 
            \includegraphics[width=\textwidth]{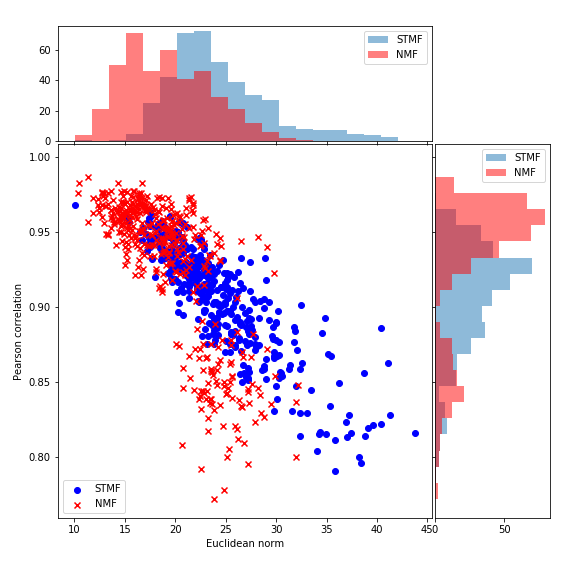}
            \caption{Pearson correlation}
        \end{subfigure}
        \hfill
        \begin{subfigure}[t]{0.32\textwidth}
            \centering 
            \includegraphics[width=\textwidth]{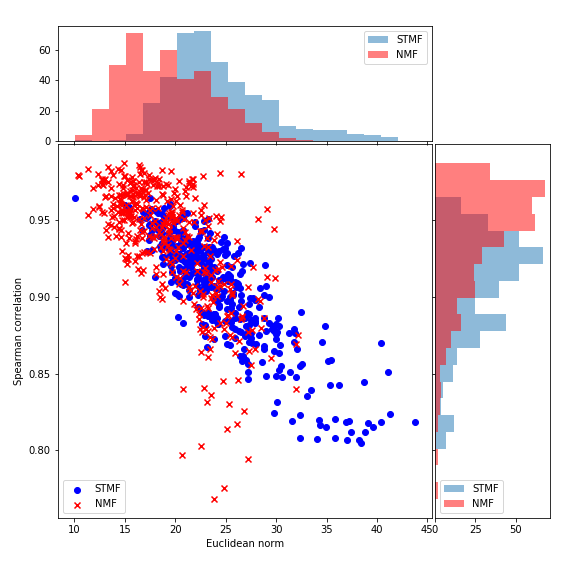}
            \caption{Spearman correlation}
        \end{subfigure}
        \caption{Euclidean norm on centered data, Pearson and Spearman correlation on \texttt{LIHC} data.}
        \label{lihc_corrs}
\end{figure*}

\clearpage
\subsection{\texttt{LUSC}}
\begin{figure}[!htb]
\captionsetup{justification=centering}
\begin{subfigure}{.66\textwidth}
\centering
\includegraphics[scale=0.27]{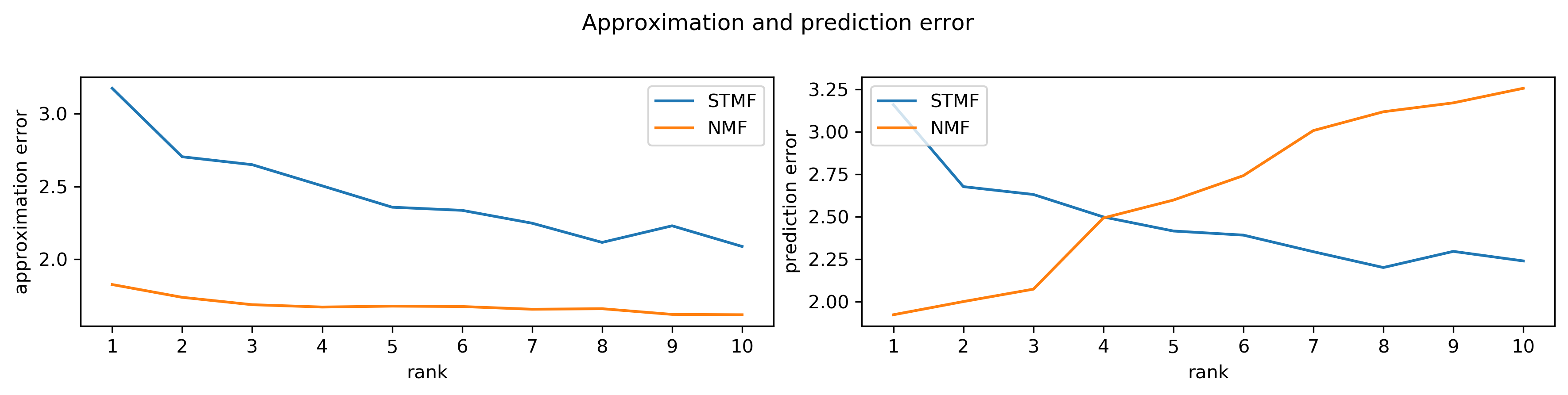}
\end{subfigure}
\begin{subfigure}{.33\textwidth}
\centering
\includegraphics[scale=0.27]{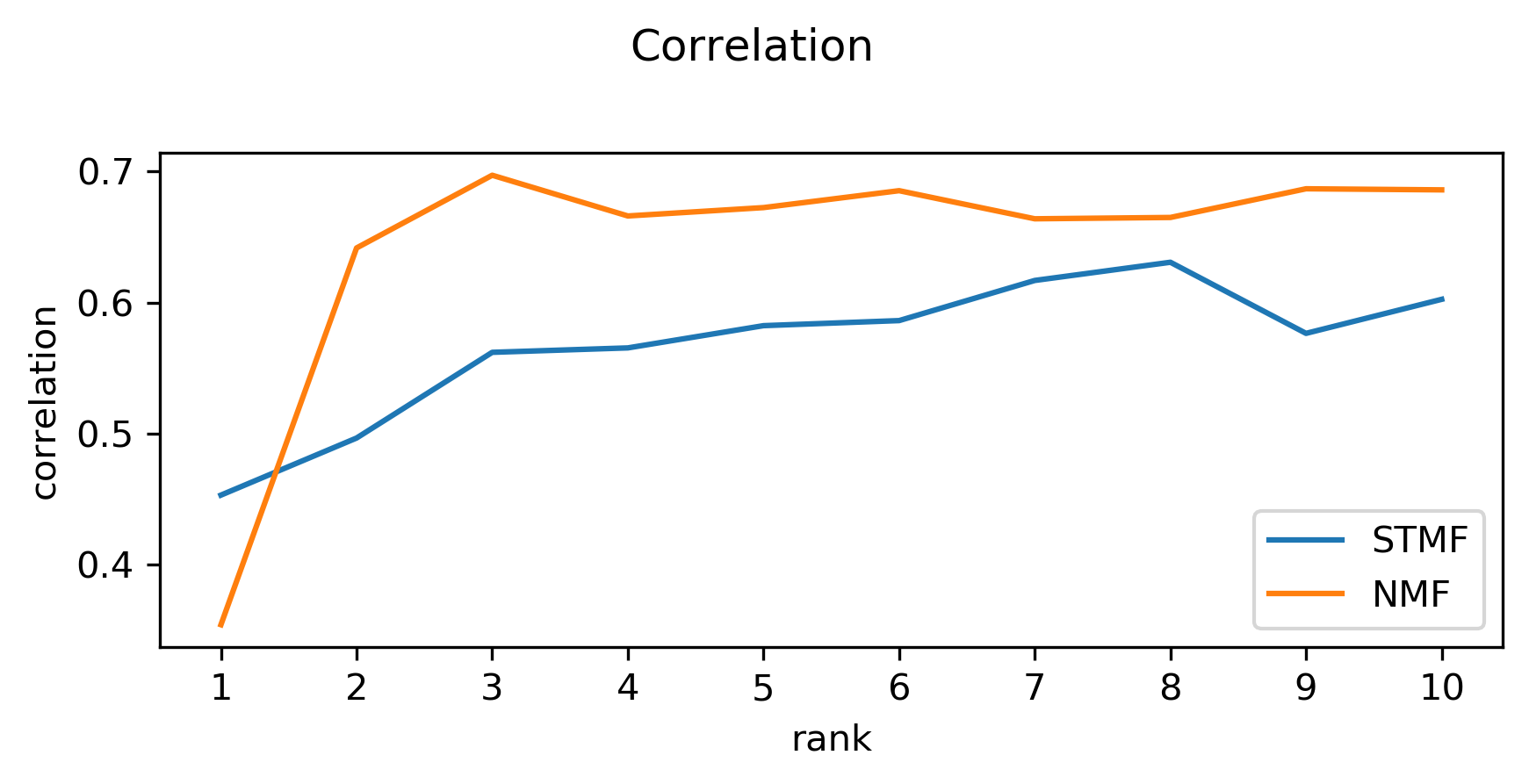}
\end{subfigure}
\caption{Difference between approximation and \\prediction RMSE and distance correlation of \texttt{STMF} \\ and \texttt{NMF} on \texttt{LUSC} data.}
\label{lusc_error_and_corr}
\end{figure}

\begin{figure}[!htb]
    \centering
    \captionsetup{justification=centering}
    \includegraphics[scale=0.35]{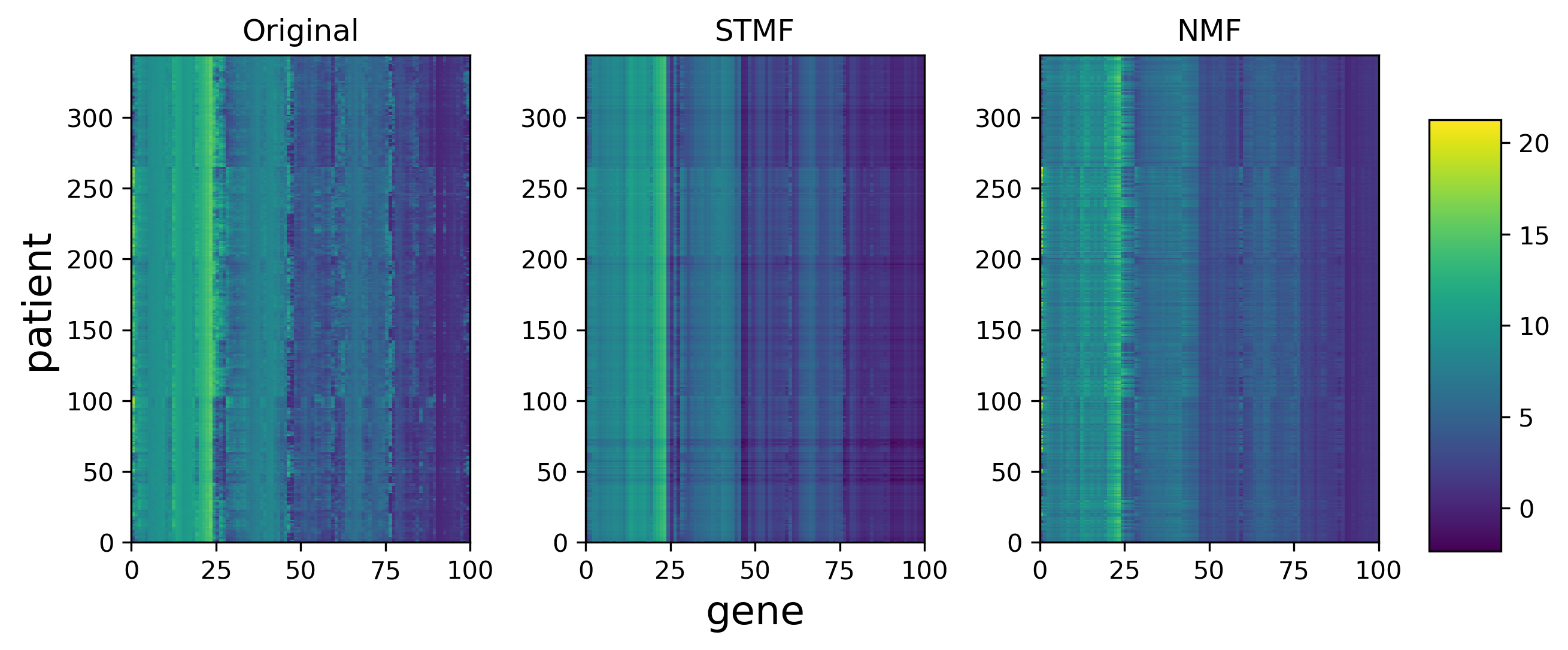}
    \caption{A comparison between \texttt{STMF}'s and \texttt{NMF}'s \\ predictions of rank 3 approximations on  \texttt{LUSC} data with $20\%$ missing values.}
    \label{lusc_synth_orig_approx}
\end{figure}

\begin{figure}[!htb] 
    \centering
    \captionsetup{justification=centering}
    \begin{subfigure}{.5\textwidth}
    \includegraphics[height=0.42\textwidth, width=\textwidth]{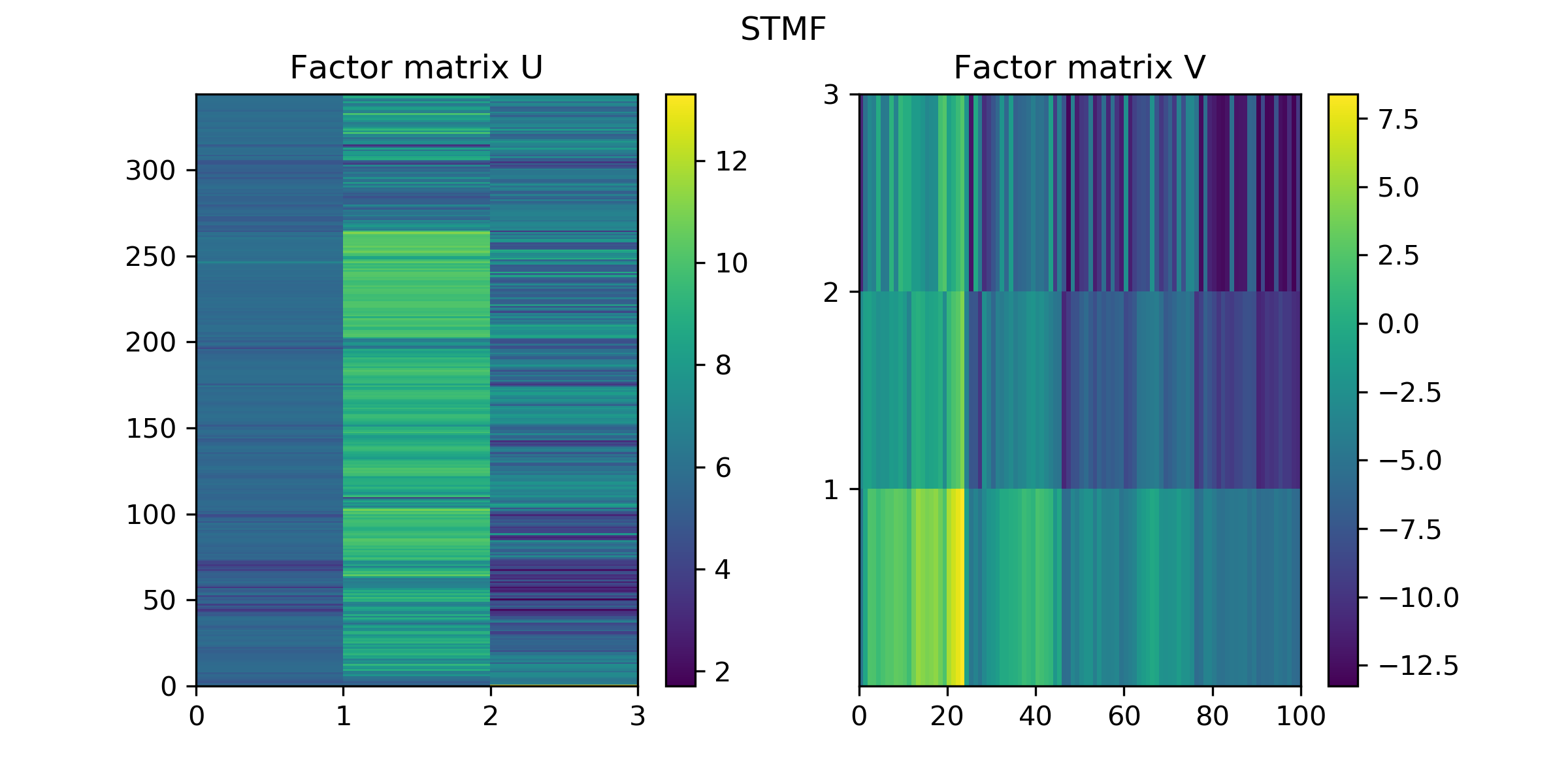}
    \caption{Factor matrices $U_{\texttt{STMF}}, V_{\texttt{STMF}}$ from \texttt{STMF}.}
\end{subfigure}\hfill%
\begin{subfigure}{.5\textwidth}
    \includegraphics[height=0.42\textwidth, width=\textwidth]{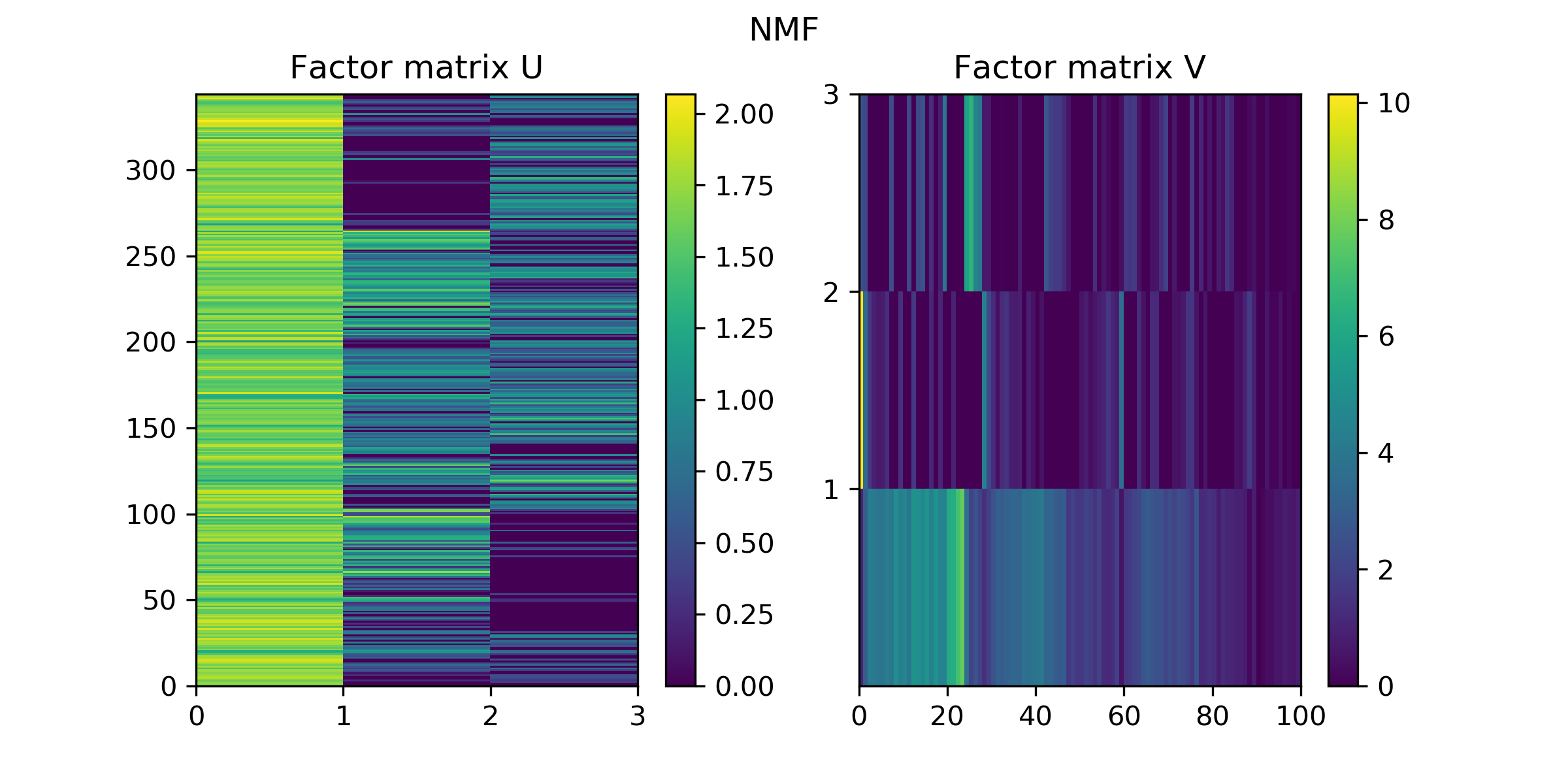}
    \caption{Factor matrices $U_{\texttt{NMF}}, V_{\texttt{NMF}}$ from \texttt{NMF}.}
\end{subfigure}
    \caption{Factor matrices $U_{\texttt{STMF}}, V_{\texttt{STMF}}$ and  $U_{\texttt{NMF}}, V_{\texttt{NMF}}$ from \texttt{STMF} and \texttt{NMF} on \texttt{LUSC} data,  respectively.}
    \label{lusc_factor_matrices}
\end{figure}

\begin{figure}[!htb] 
    \centering
    \begin{subfigure}[t]{.5\textwidth}
        \includegraphics[height=0.42\textwidth, width=\textwidth]{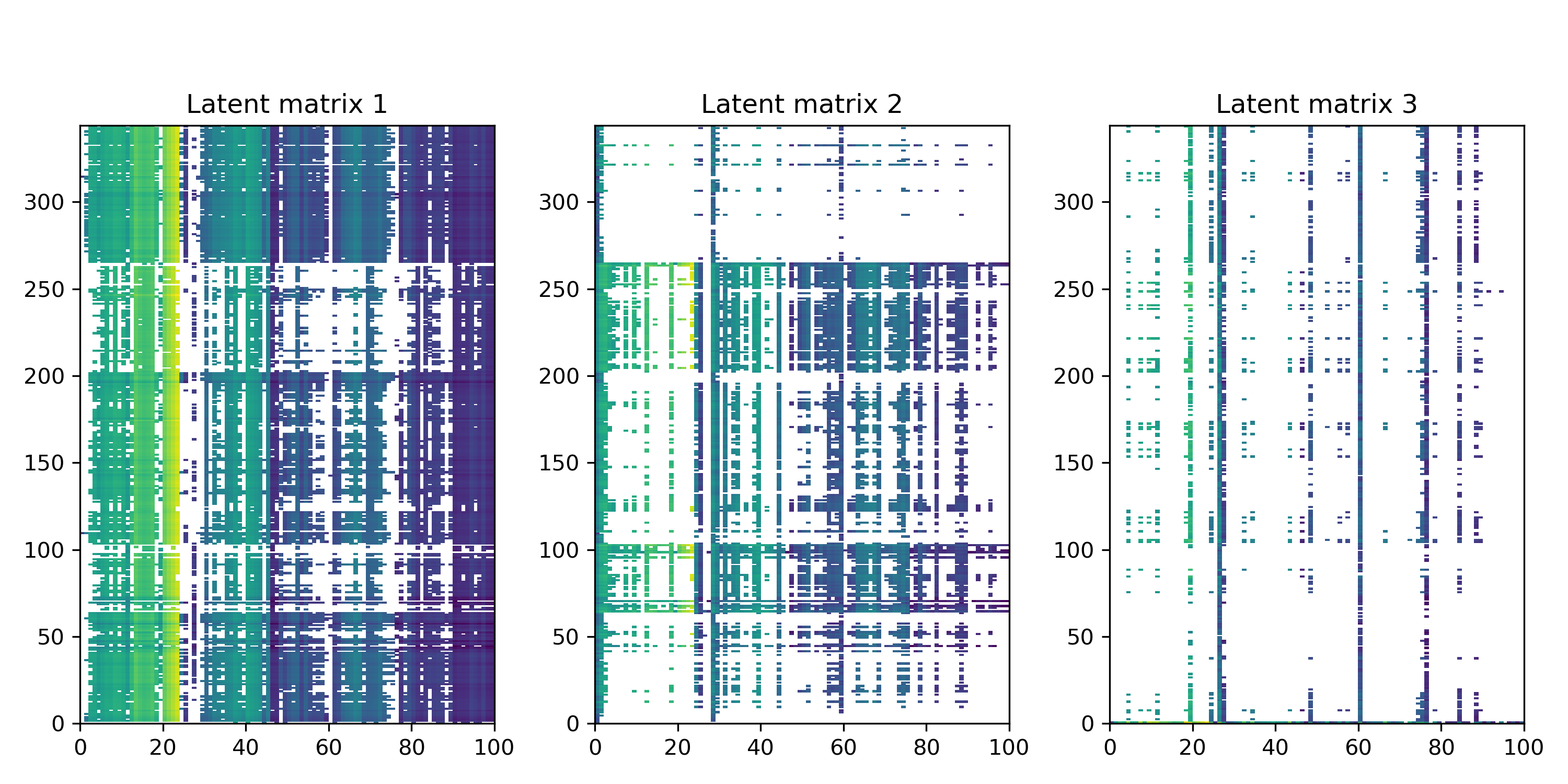}
        \caption{Latent matrices $R_{\texttt{STMF}}^{(i)}$, $i \in \{1, 3\}$, where \\ white represents the element which does not \\ contribute to the approximation $R_{\texttt{STMF}}$.}
    \end{subfigure}\hfill%
    \begin{subfigure}[t]{.5\textwidth}
        \includegraphics[height=0.42\textwidth, width=\textwidth]{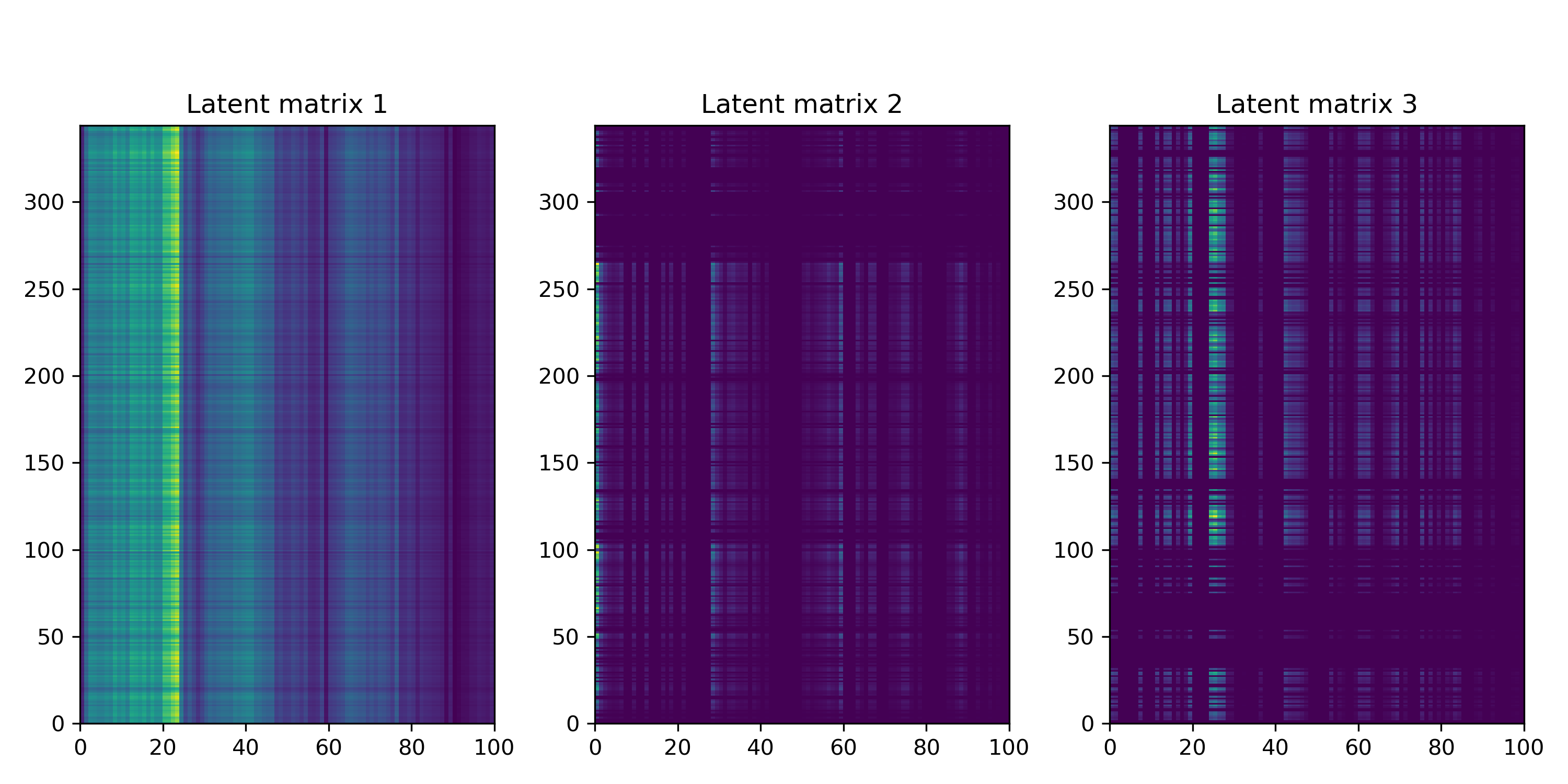}
        \caption{Latent matrices $R_{\texttt{NMF}}^{(i)}$, $i \in \{1, 3\}$.}
    \end{subfigure}
    \caption{\texttt{STMF}'s and \texttt{NMF}'s latent matrices on \texttt{LUSC} data.}
    \label{lusc_latent_matrices}
\end{figure}

\begin{figure*}[!htb]
        \centering
        \begin{subfigure}[t]{0.32\textwidth}
            \centering
            \includegraphics[width=\textwidth]{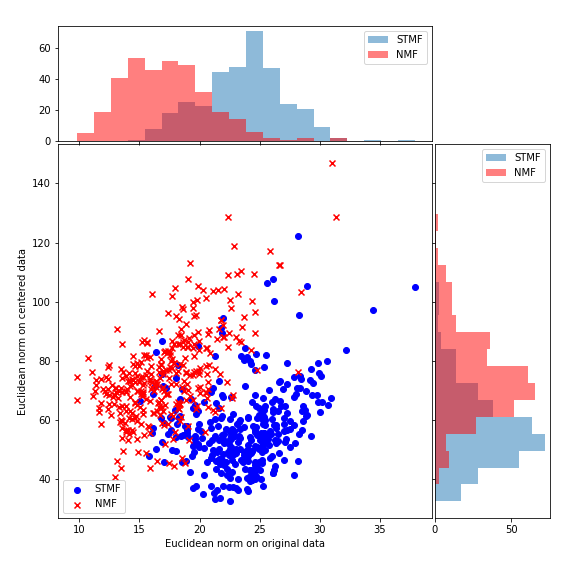}
            \caption{Euclidean norm on centered data}
        \end{subfigure}
        \hfill
        \begin{subfigure}[t]{0.32\textwidth}
            \centering 
            \includegraphics[width=\textwidth]{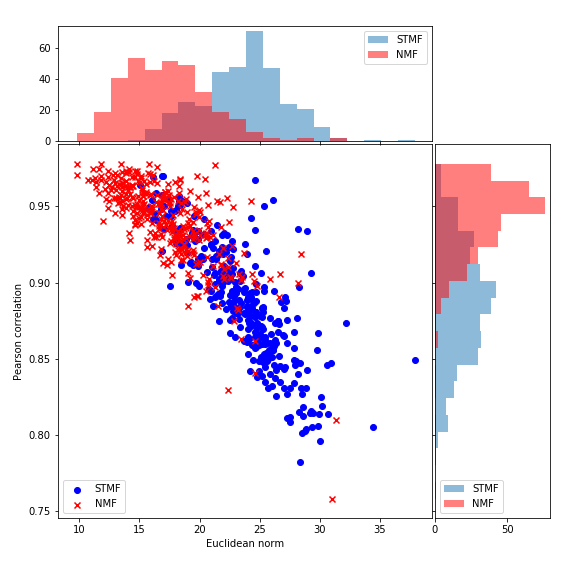}
            \caption{Pearson correlation}
        \end{subfigure}
        \hfill
        \begin{subfigure}[t]{0.32\textwidth}
            \centering 
            \includegraphics[width=\textwidth]{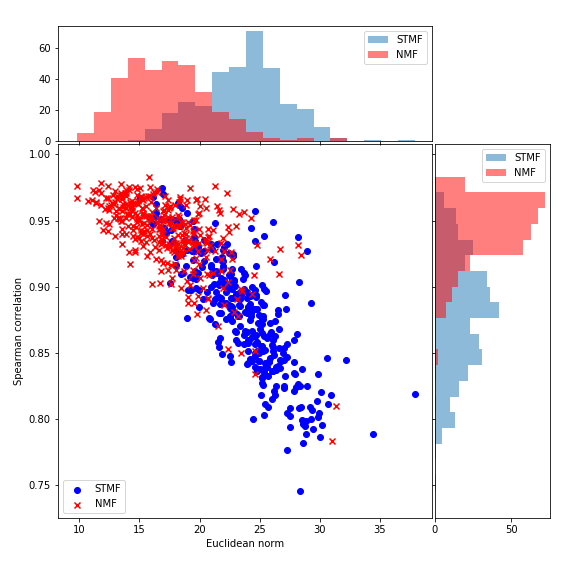}
            \caption{Spearman correlation}
        \end{subfigure}
        \caption{Euclidean norm on centered data, Pearson and Spearman correlation on \texttt{LUSC} data.}
        \label{lusc_corrs}
\end{figure*}

\clearpage
\subsection{\texttt{OV}}
\begin{figure}[!htb]
\captionsetup{justification=centering}
\begin{subfigure}{.66\textwidth}
\centering
\includegraphics[scale=0.27]{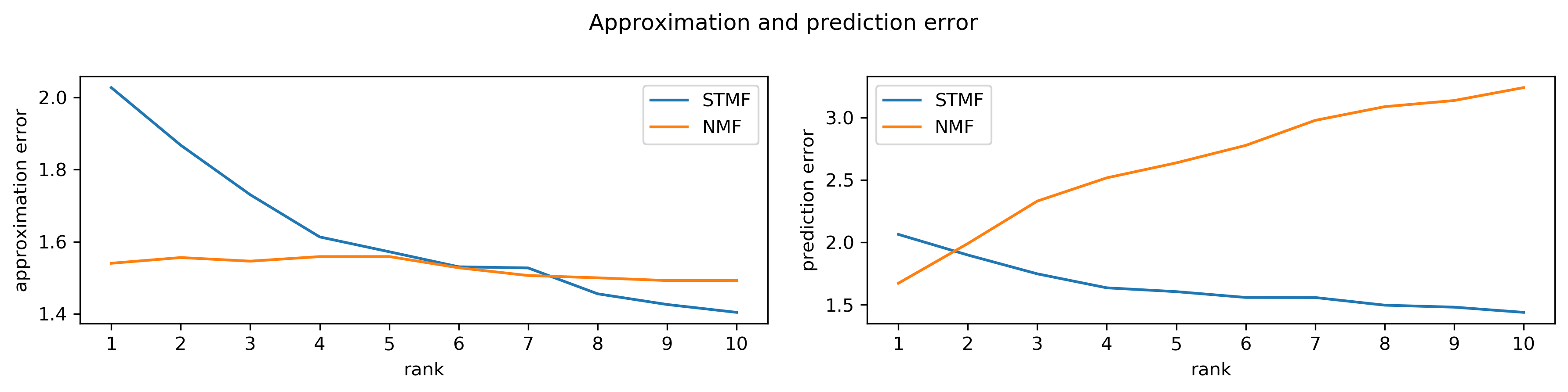}
\end{subfigure}
\begin{subfigure}{.33\textwidth}
\centering
\includegraphics[scale=0.27]{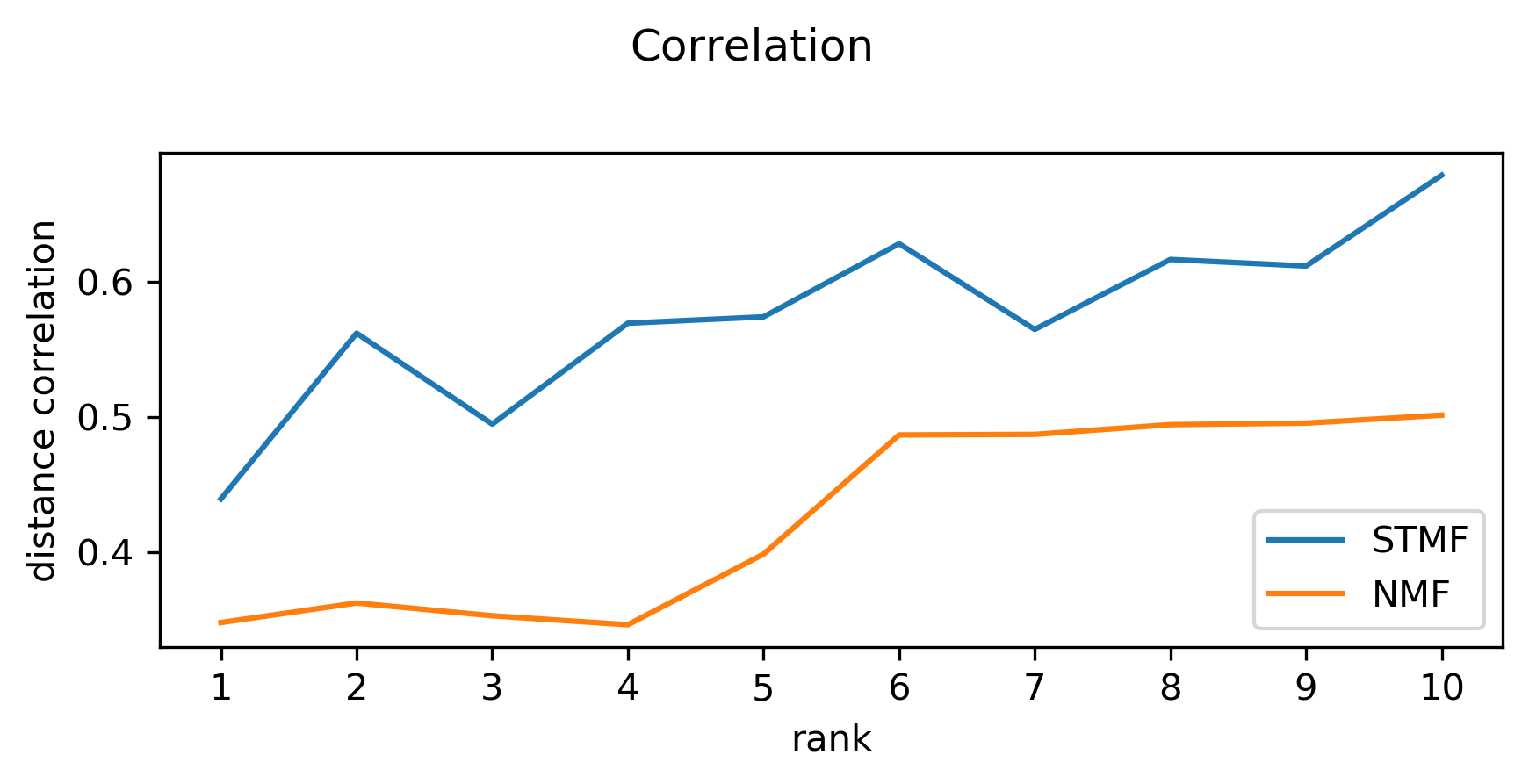}
\end{subfigure}
\caption{Difference between approximation and \\prediction RMSE and distance correlation of \texttt{STMF} \\ and \texttt{NMF} on \texttt{OV} data.}
\label{ov_error_and_corr}
\end{figure}

\begin{figure}[!htb]
    \centering
    \captionsetup{justification=centering}
    \includegraphics[scale=0.35]{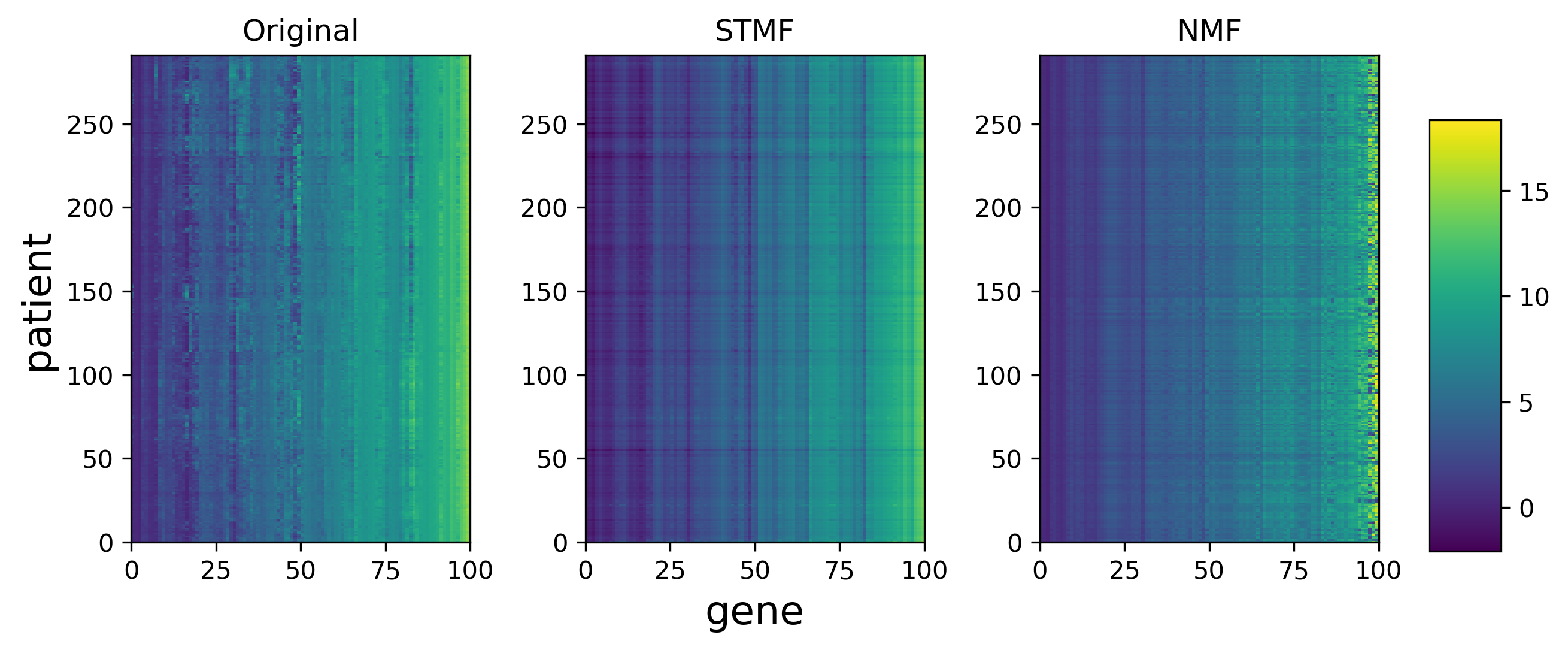}
    \caption{A comparison between \texttt{STMF}'s and \texttt{NMF}'s \\ predictions of rank 4 approximations on  \texttt{OV} data with $20\%$ missing values.}
    \label{ov_synth_orig_approx}
\end{figure}

\begin{figure}[!htb] 
    \centering
    \captionsetup{justification=centering}
    \begin{subfigure}{.5\textwidth}
    \includegraphics[height=0.42\textwidth, width=\textwidth]{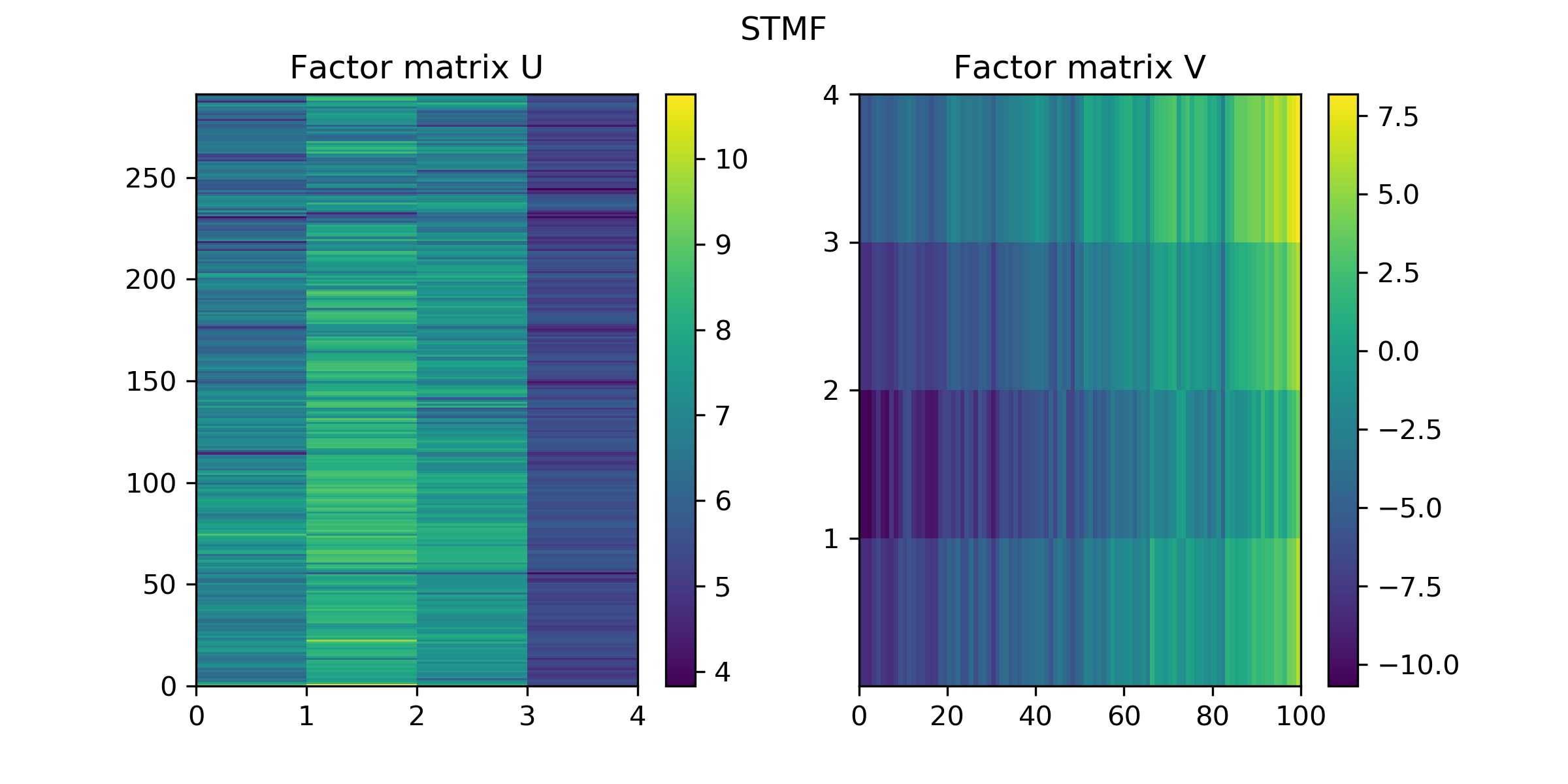}
    \caption{Factor matrices $U_{\texttt{STMF}}, V_{\texttt{STMF}}$ from \texttt{STMF}.}
\end{subfigure}\hfill%
\begin{subfigure}{.5\textwidth}
    \includegraphics[height=0.42\textwidth, width=\textwidth]{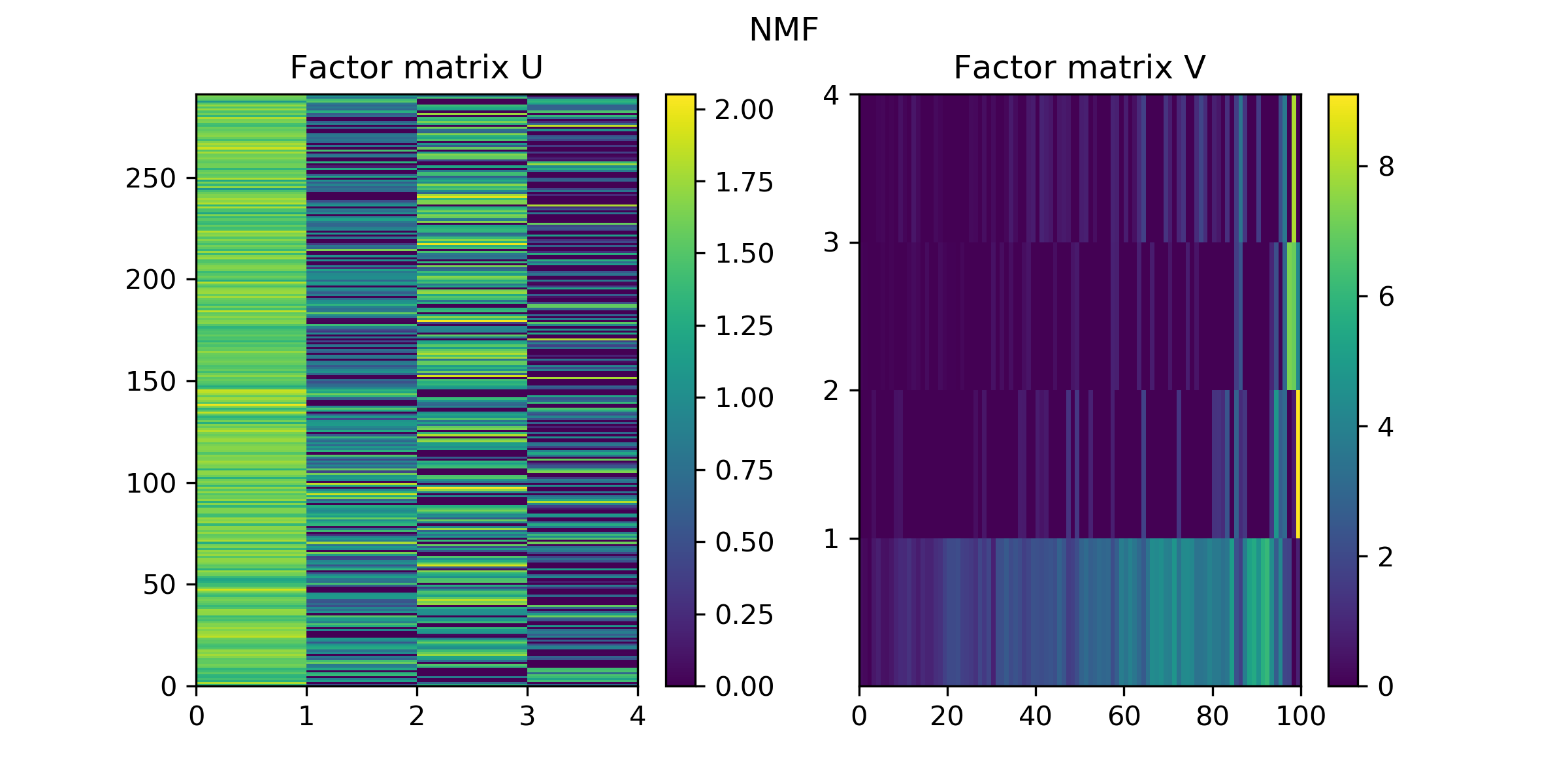}
    \caption{Factor matrices $U_{\texttt{NMF}}, V_{\texttt{NMF}}$ from \texttt{NMF}.}
\end{subfigure}
    \caption{Factor matrices $U_{\texttt{STMF}}, V_{\texttt{STMF}}$ and  $U_{\texttt{NMF}}, V_{\texttt{NMF}}$ from \texttt{STMF} and \texttt{NMF} on \texttt{OV} data,  respectively.}
    \label{ov_factor_matrices}
\end{figure}

\begin{figure}[!htb] 
    \centering
    \begin{subfigure}[t]{.5\textwidth}
        \includegraphics[height=0.42\textwidth, width=\textwidth]{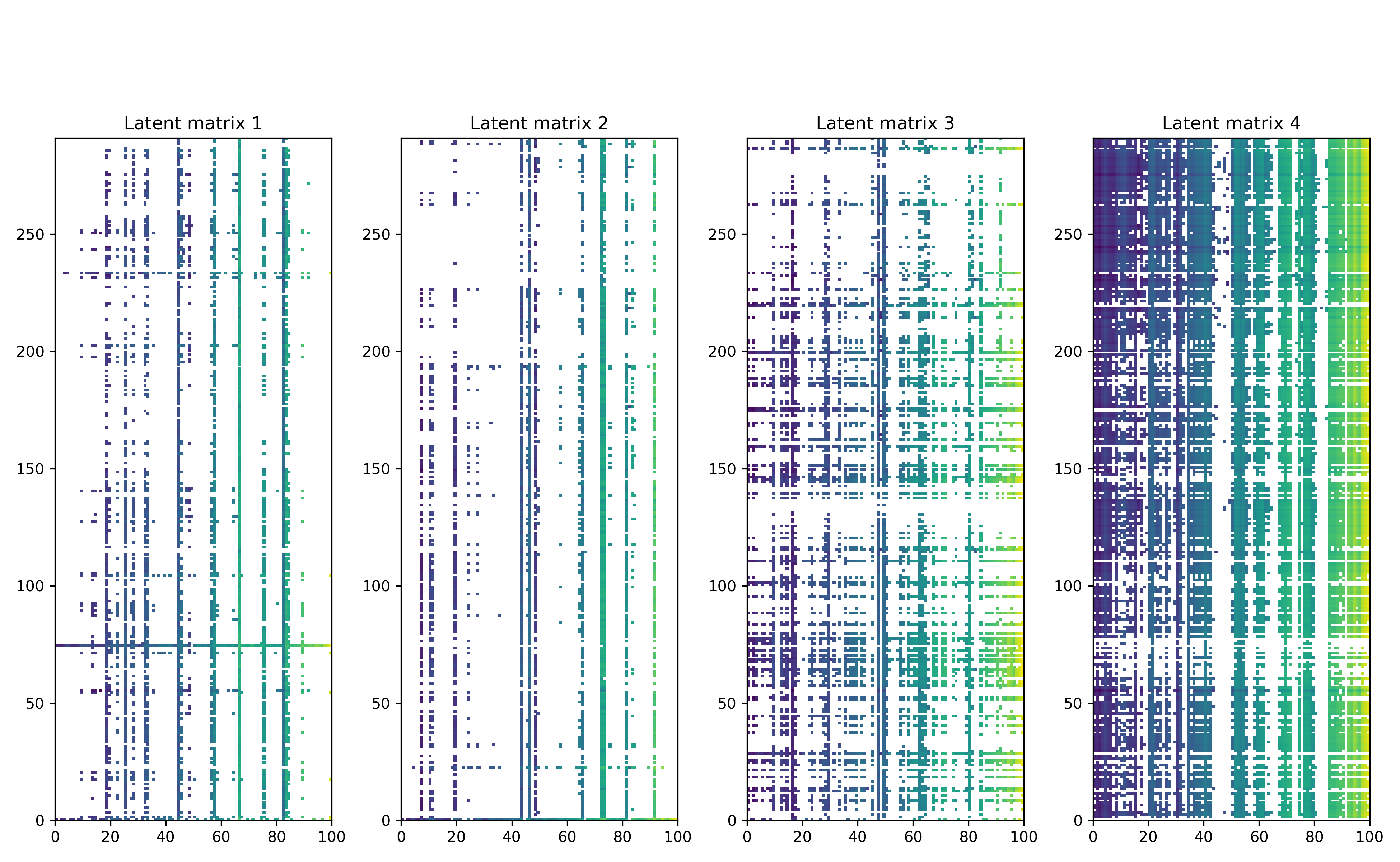}
        \caption{Latent matrices $R_{\texttt{STMF}}^{(i)}$, $i \in \{1, 4\}$, where \\ white represents the element which does not \\ contribute to the approximation $R_{\texttt{STMF}}$.}
    \end{subfigure}\hfill%
    \begin{subfigure}[t]{.5\textwidth}
        \includegraphics[height=0.42\textwidth, width=\textwidth]{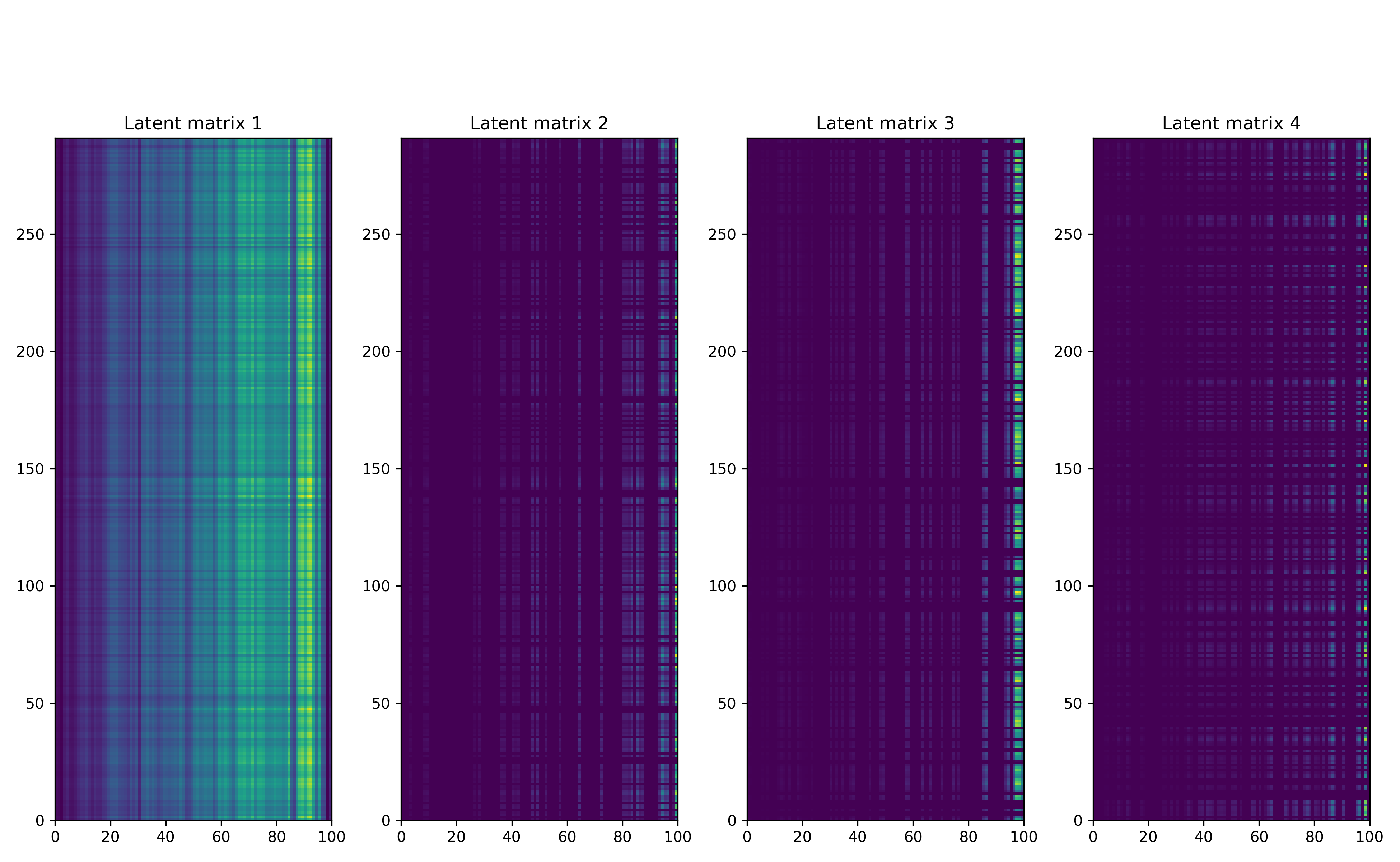}
        \caption{Latent matrices $R_{\texttt{NMF}}^{(i)}$, $i \in \{1, 4\}$.}
    \end{subfigure}
    \caption{\texttt{STMF}'s and \texttt{NMF}'s latent matrices on \texttt{OV} data.}
    \label{ov_latent_matrices}
\end{figure}

\begin{figure*}[!htb]
    \centering
    \begin{subfigure}[t]{0.32\textwidth}
        \centering
        \includegraphics[width=\textwidth]{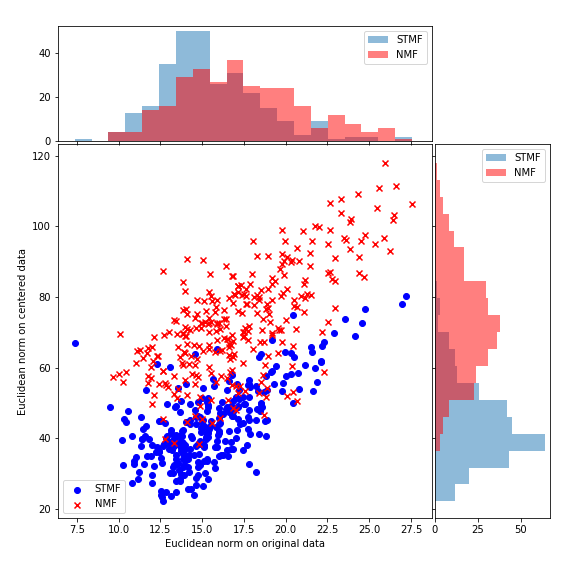}
        \caption{Euclidean norm on centered data}
    \end{subfigure}
    \hfill
    \begin{subfigure}[t]{0.32\textwidth}  
        \centering 
        \includegraphics[width=\textwidth]{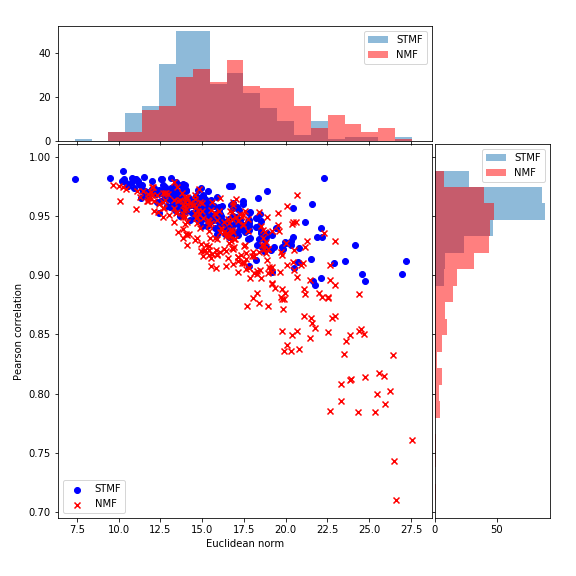}
        \caption{Pearson correlation}
    \end{subfigure}
    \hfill
    \begin{subfigure}[t]{0.32\textwidth}  
        \centering 
        \includegraphics[width=\textwidth]{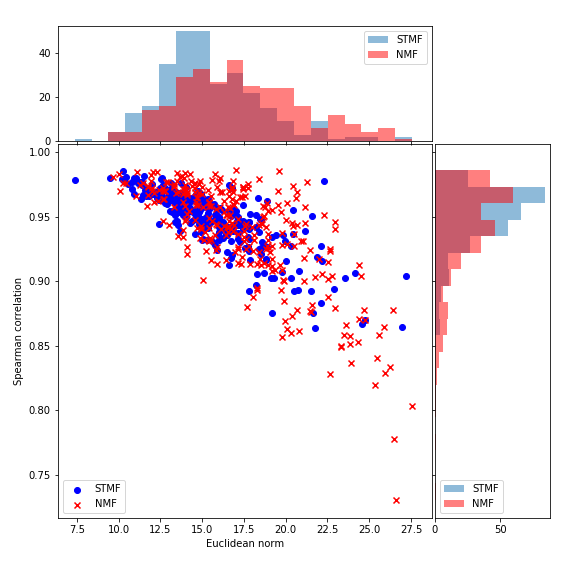}
        \caption{Spearman correlation}
    \end{subfigure}
    \caption{Euclidean norm on centered data, Pearson and Spearman correlation on \texttt{OV} data.}
    \label{ov_corrs}
\end{figure*}

\clearpage
\subsection{\texttt{SKCM}}
\begin{figure}[!htb]
\captionsetup{justification=centering}
\begin{subfigure}{.66\textwidth}
\centering
\includegraphics[scale=0.27]{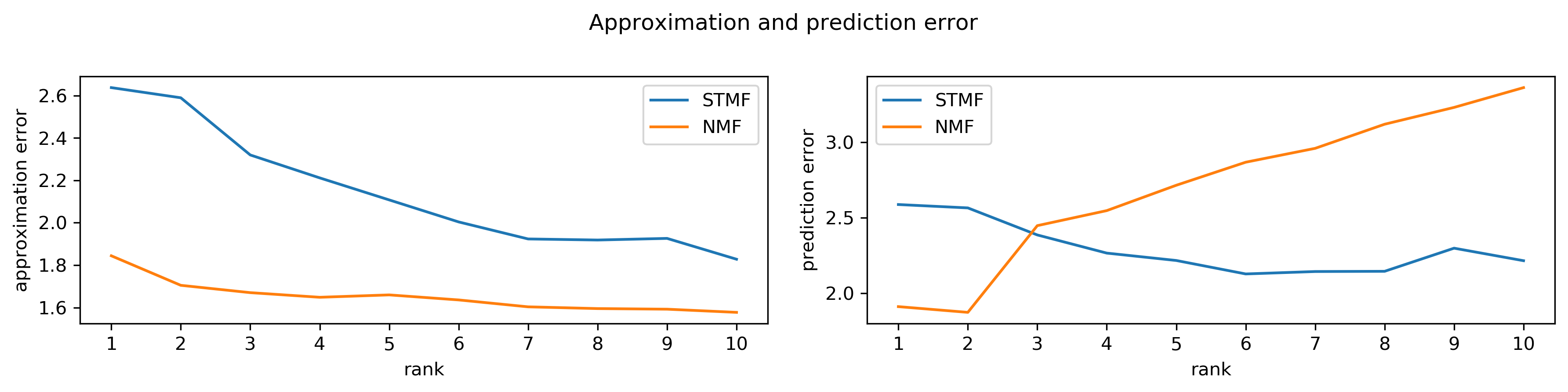}
\end{subfigure}
\begin{subfigure}{.33\textwidth}
\centering
\includegraphics[scale=0.27]{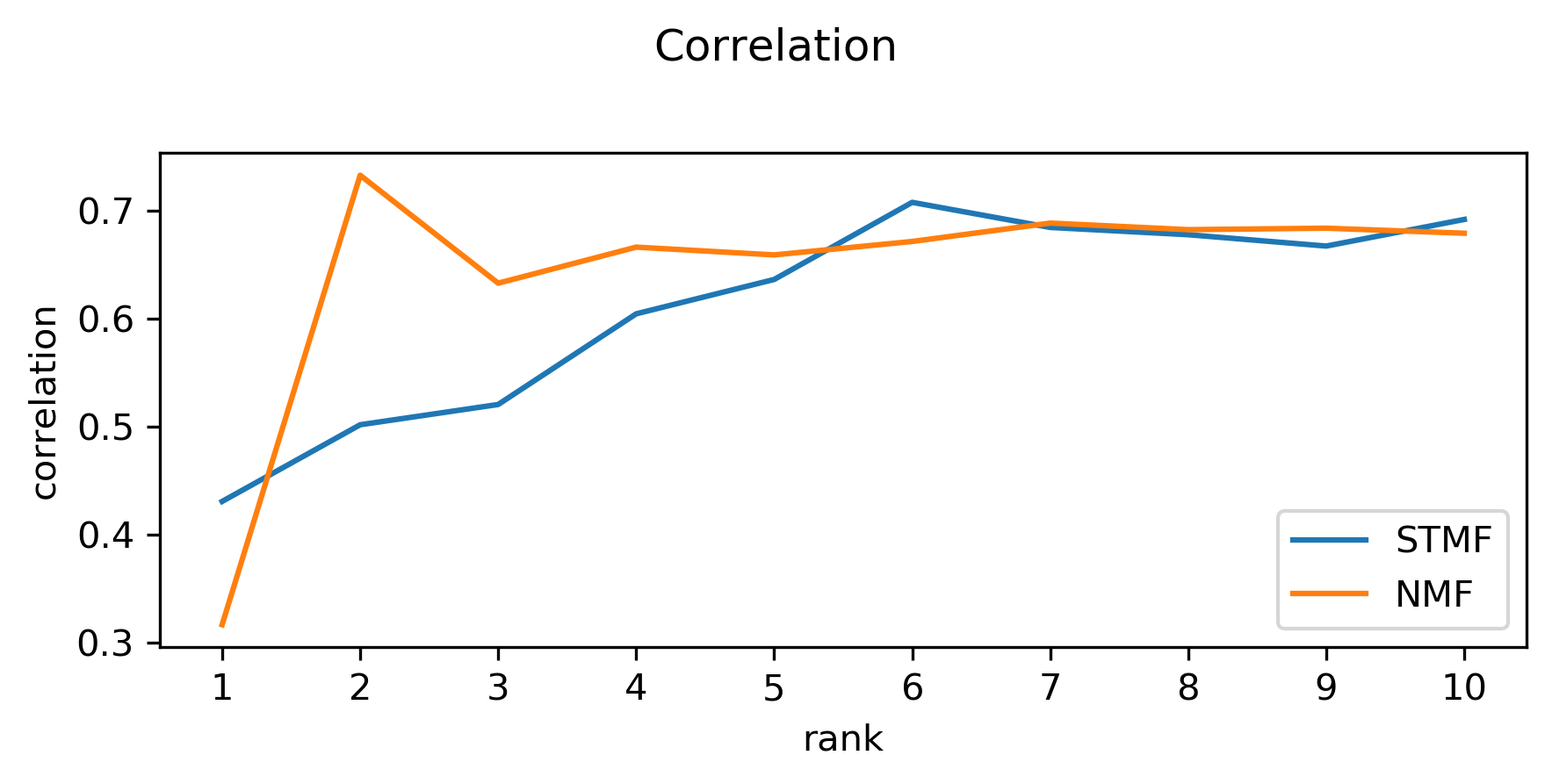}
\end{subfigure}
\caption{Difference between approximation and \\prediction RMSE and distance correlation of \texttt{STMF} \\ and \texttt{NMF} on \texttt{SKCM} data.}
\label{skcm_error_and_corr}
\end{figure}

\begin{figure}[!htb]
    \centering
    \captionsetup{justification=centering}
    \includegraphics[scale=0.35]{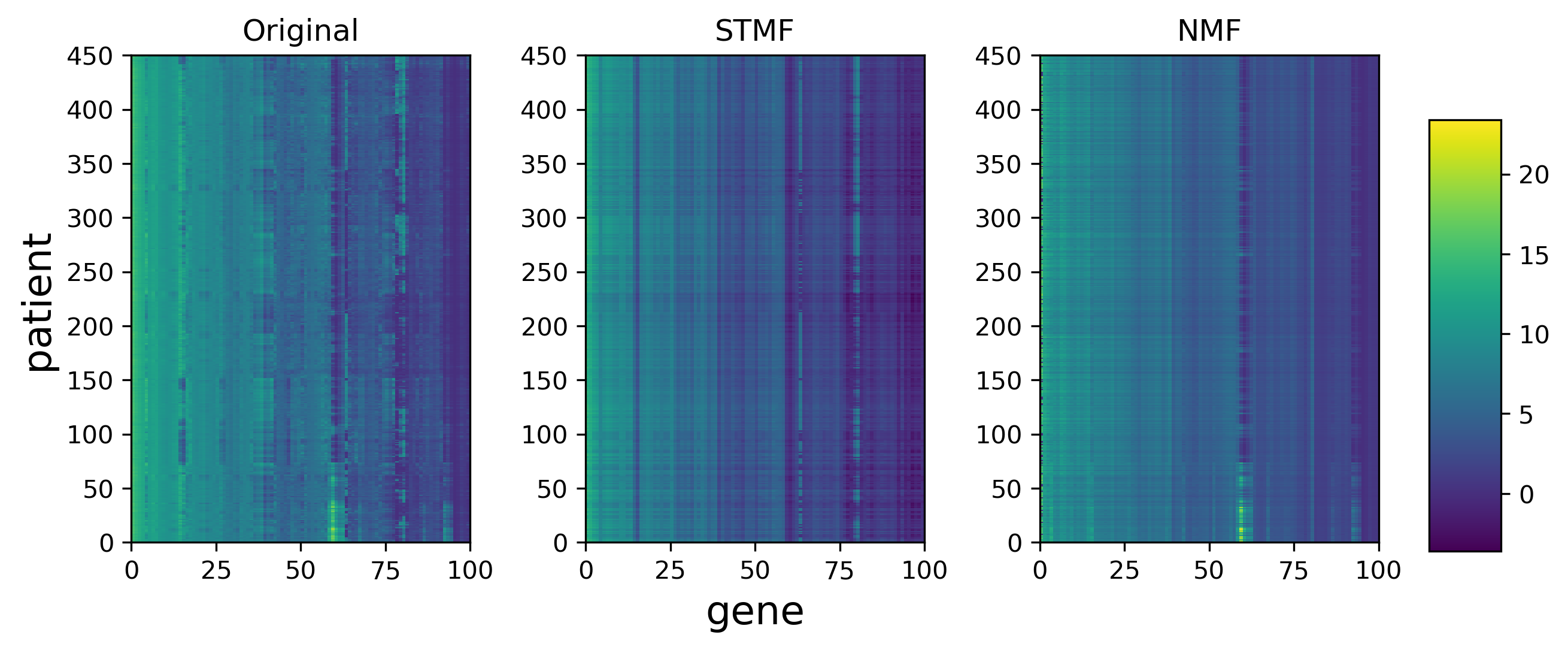}
    \caption{A comparison between \texttt{STMF}'s and \texttt{NMF}'s \\ predictions of rank 3 approximations on  \texttt{SKCM} data with $20\%$ missing values.}
    \label{skcm_orig_approx}
\end{figure}

\begin{figure}[!htb] 
    \centering
    \captionsetup{justification=centering}
    \begin{subfigure}{.5\textwidth}
    \includegraphics[height=0.42\textwidth, width=\textwidth]{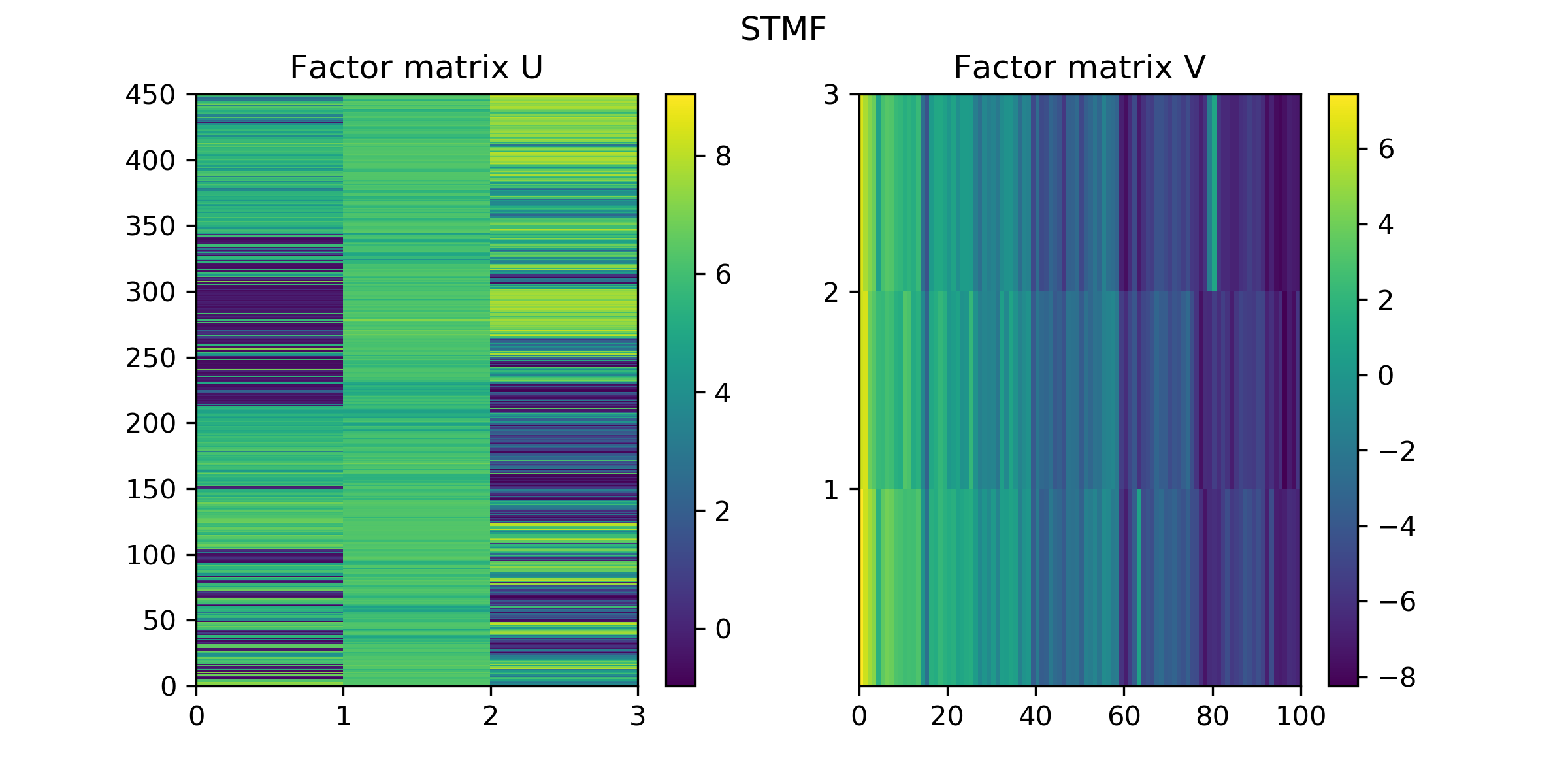}
    \caption{Factor matrices $U_{\texttt{STMF}}, V_{\texttt{STMF}}$ from \texttt{STMF}.}
\end{subfigure}\hfill%
\begin{subfigure}{.5\textwidth}
    \includegraphics[height=0.42\textwidth, width=\textwidth]{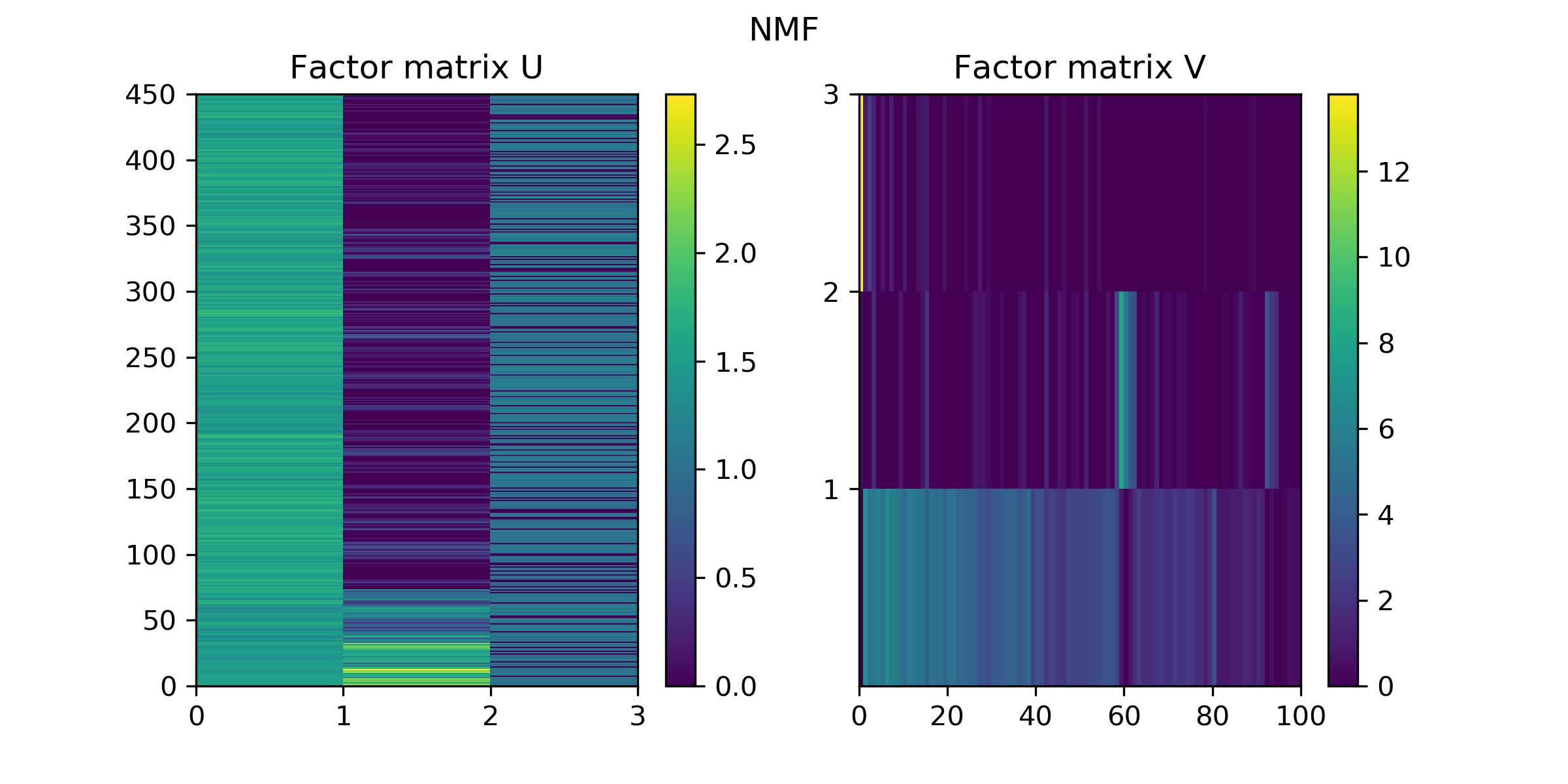}
    \caption{Factor matrices $U_{\texttt{NMF}}, V_{\texttt{NMF}}$ from \texttt{NMF}.}
\end{subfigure}
    \caption{Factor matrices $U_{\texttt{STMF}}, V_{\texttt{STMF}}$ and  $U_{\texttt{NMF}}, V_{\texttt{NMF}}$ from \texttt{STMF} and \texttt{NMF} on \texttt{SKCM} data,  respectively.}
    \label{skcm_factor_matrices}
\end{figure}

\begin{figure}[!htb] 
    \centering
    \begin{subfigure}[t]{.5\textwidth}
        \includegraphics[height=0.42\textwidth, width=\textwidth]{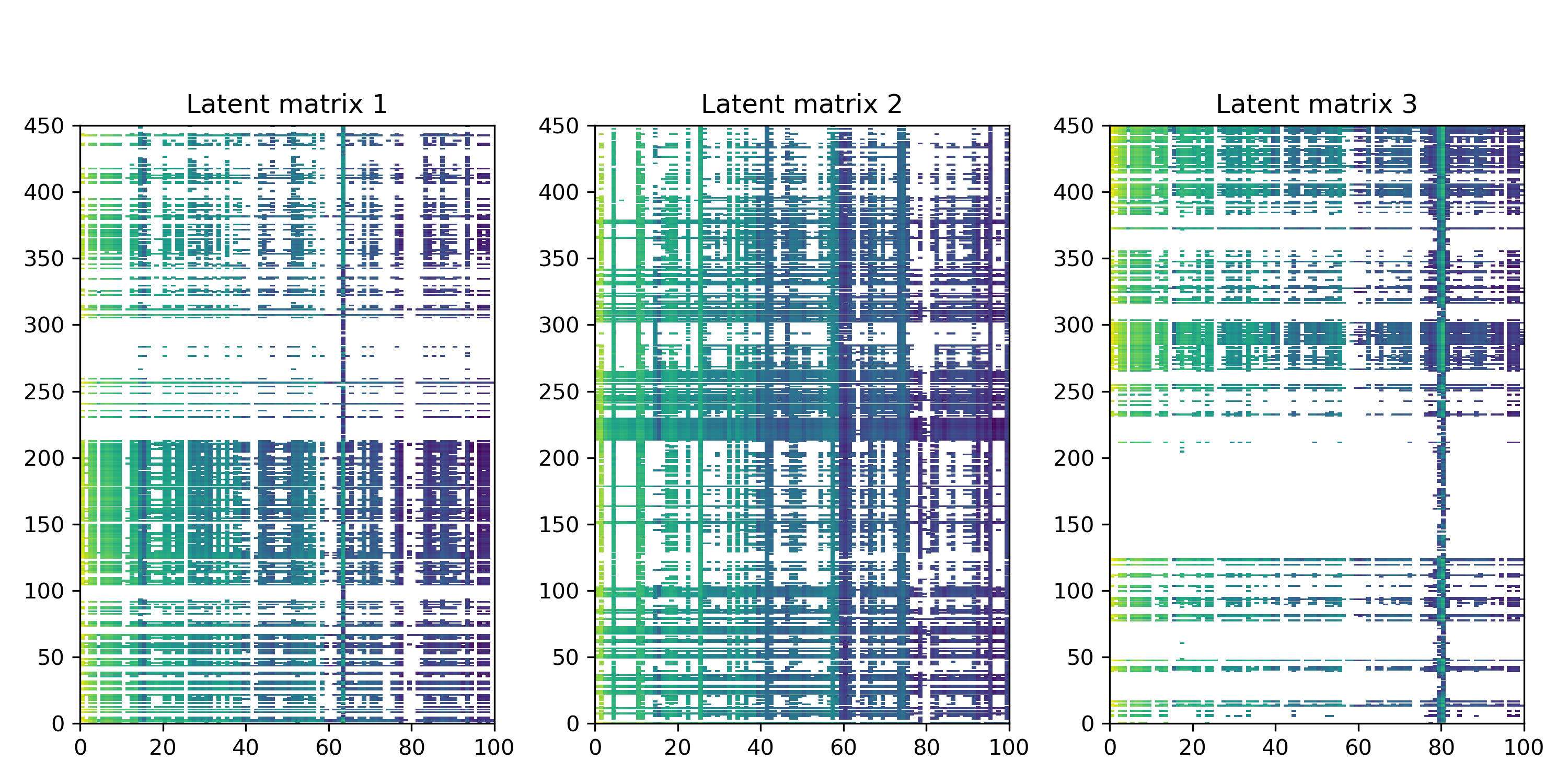}
        \caption{Latent matrices $R_{\texttt{STMF}}^{(i)}$, $i \in \{1, 3\}$, where \\ white represents the element which does not \\ contribute to the approximation $R_{\texttt{STMF}}$.}
    \end{subfigure}\hfill%
    \begin{subfigure}[t]{.5\textwidth}
        \includegraphics[height=0.42\textwidth, width=\textwidth]{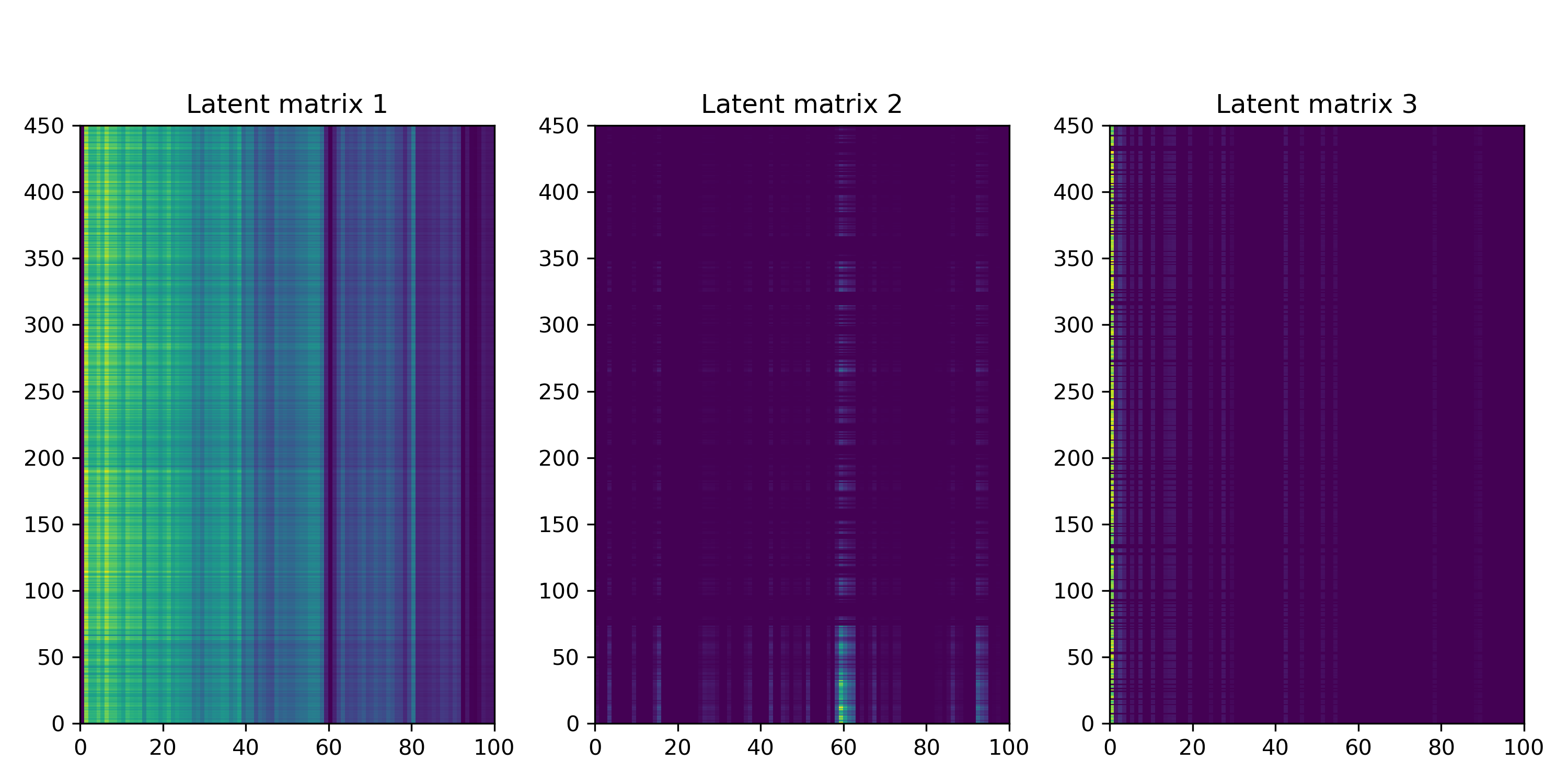}
        \caption{Latent matrices $R_{\texttt{NMF}}^{(i)}$, $i \in \{1, 3\}$.}
    \end{subfigure}
    \caption{\texttt{STMF}'s and \texttt{NMF}'s latent matrices on \texttt{SKCM} data.}
    \label{skcm_latent_matrices}
\end{figure}

\begin{figure*}[!htb]
        \centering
        \begin{subfigure}[t]{0.32\textwidth}
            \centering
            \includegraphics[width=\textwidth]{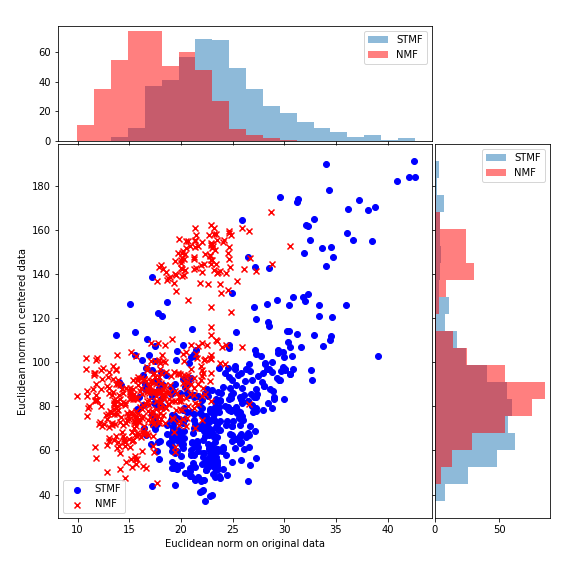}
            \caption{Euclidean norm on centered data}
        \end{subfigure}
        \hfill
        \begin{subfigure}[t]{0.32\textwidth}
            \centering 
            \includegraphics[width=\textwidth]{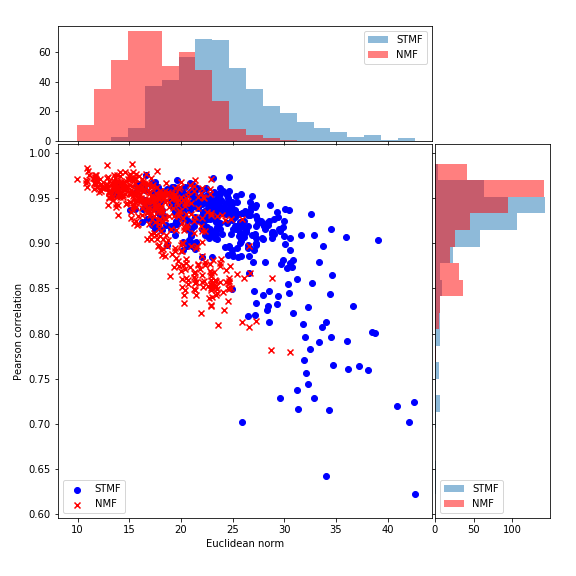}
            \caption{Pearson correlation}
        \end{subfigure}
        \hfill
        \begin{subfigure}[t]{0.32\textwidth}
            \centering 
            \includegraphics[width=\textwidth]{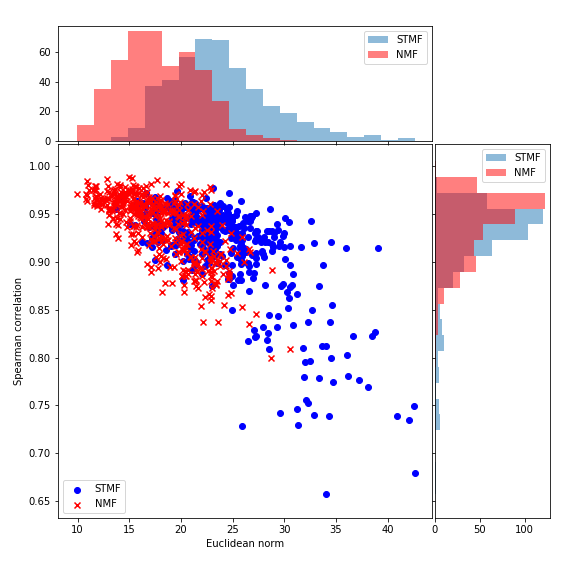}
            \caption{Spearman correlation}
        \end{subfigure}
        \caption{Euclidean norm on centered data, Pearson and Spearman correlation on \texttt{SKCM} data.}
        \label{skcm_corrs}
\end{figure*}

\clearpage
\subsection{\texttt{SARC}}
\begin{figure}[!htb]
\captionsetup{justification=centering}
\begin{subfigure}{.66\textwidth}
\centering
\includegraphics[scale=0.27]{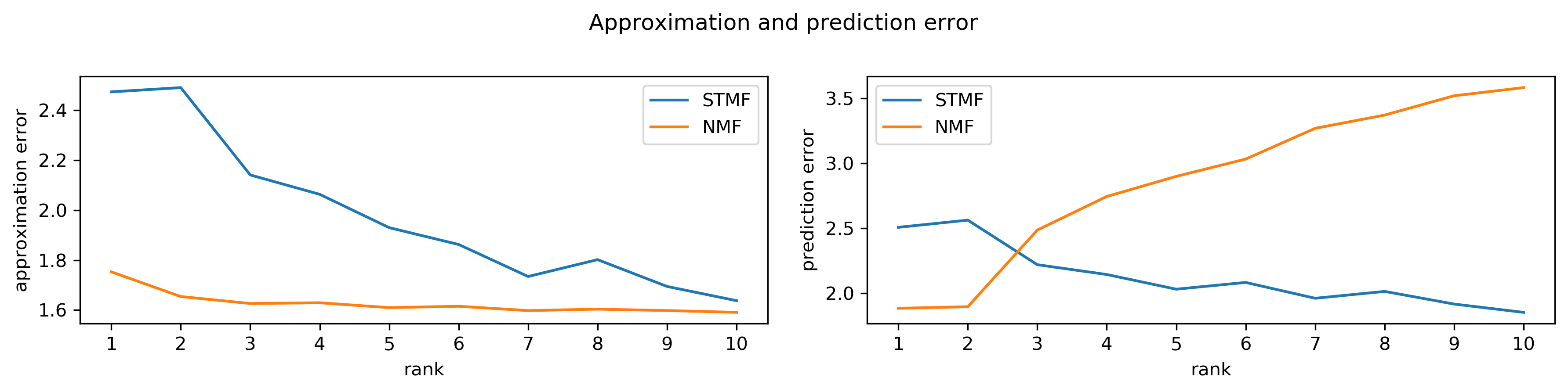}
\end{subfigure}
\begin{subfigure}{.33\textwidth}
\centering
\includegraphics[scale=0.27]{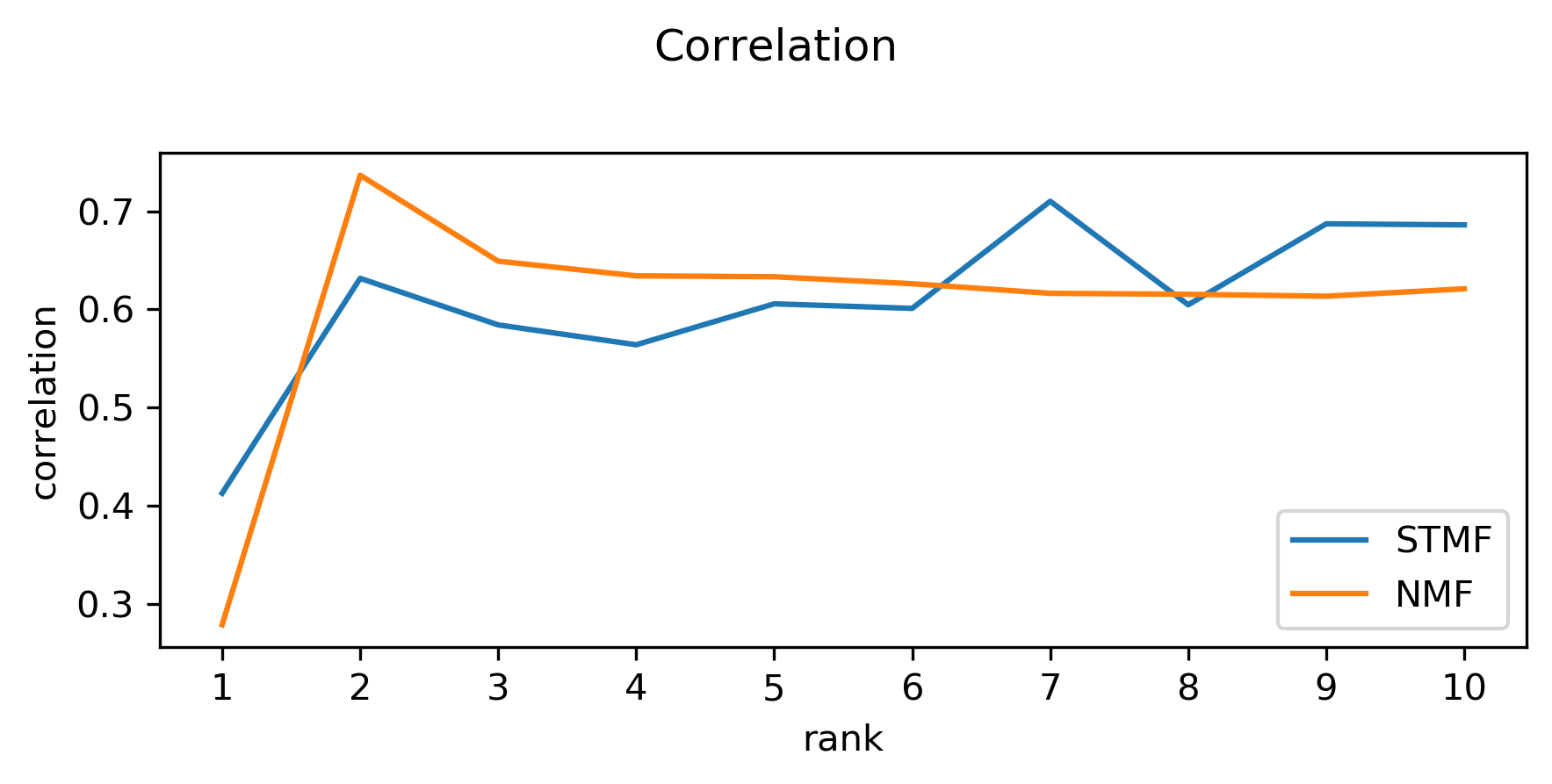}
\end{subfigure}
\caption{Difference between approximation and \\prediction RMSE and distance correlation of \texttt{STMF} \\ and \texttt{NMF} on \texttt{SARC} data.}
\label{sarc_error_and_corr}
\end{figure}

\begin{figure}[!htb]
    \centering
    \captionsetup{justification=centering}
    \includegraphics[scale=0.35]{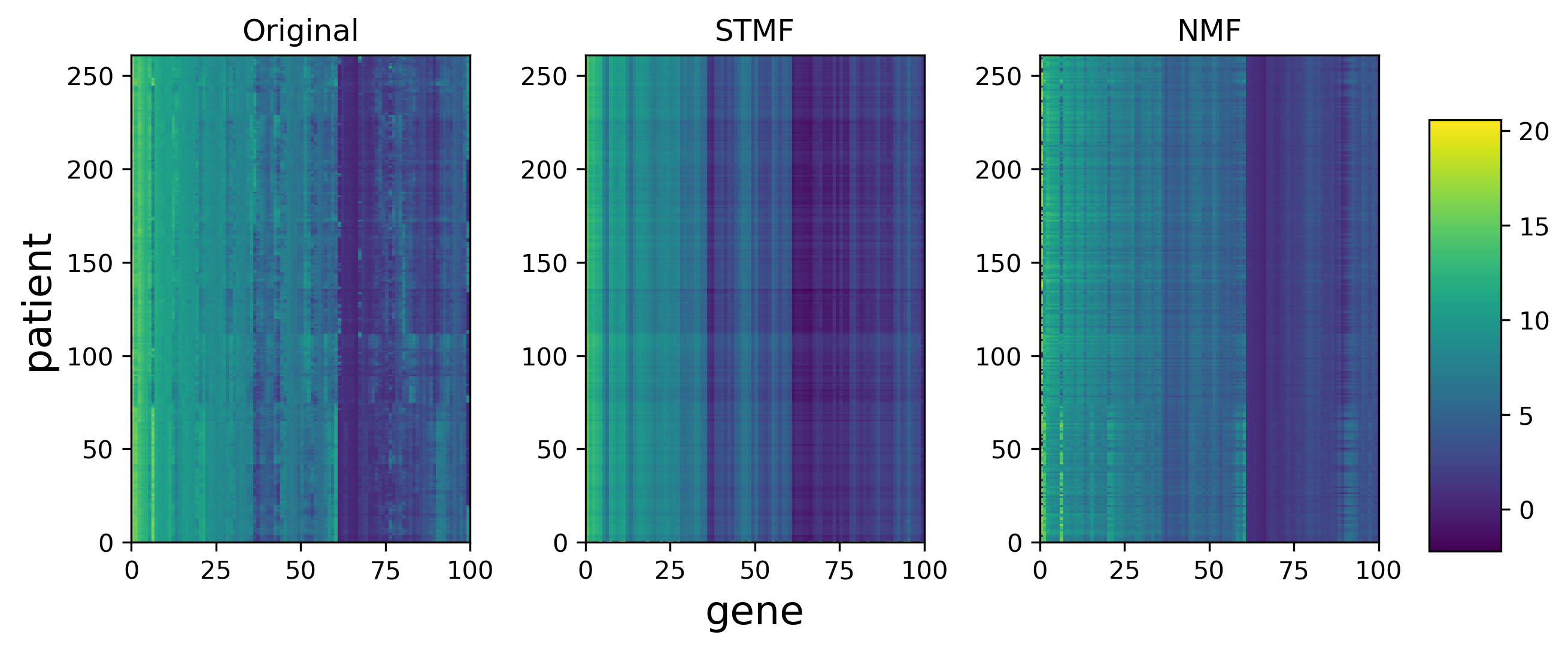}
    \caption{A comparison between \texttt{STMF}'s and \texttt{NMF}'s \\ predictions of rank 3 approximations on  \texttt{SARC} data with $20\%$ missing values.}
    \label{sarc_orig_approx}
\end{figure}

\begin{figure}[!htb] 
    \centering
    \captionsetup{justification=centering}
    \begin{subfigure}{.5\textwidth}
    \includegraphics[height=0.42\textwidth, width=\textwidth]{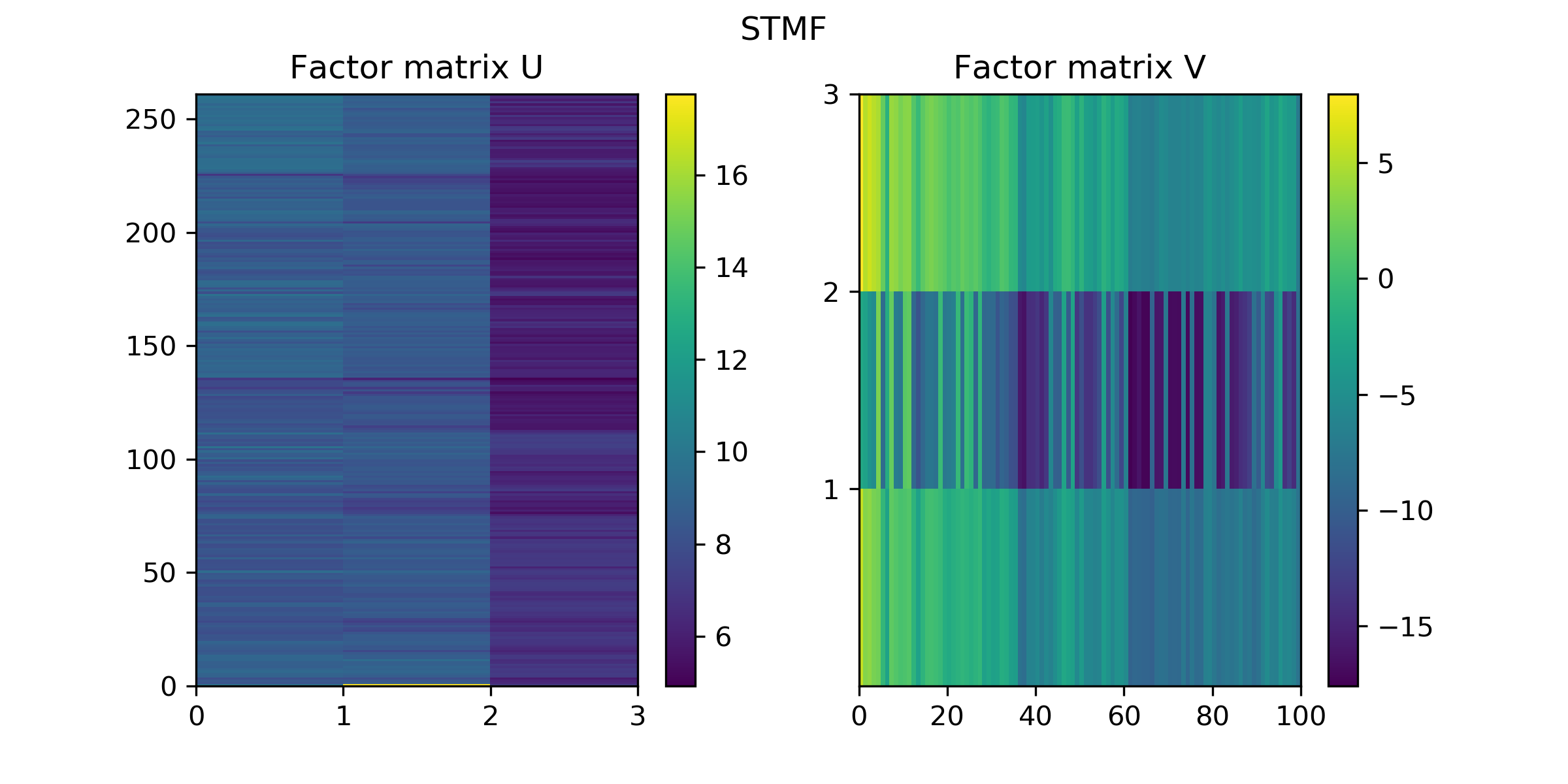}
    \caption{Factor matrices $U_{\texttt{STMF}}, V_{\texttt{STMF}}$ from \texttt{STMF}.}
\end{subfigure}\hfill%
\begin{subfigure}{.5\textwidth}
    \includegraphics[height=0.42\textwidth, width=\textwidth]{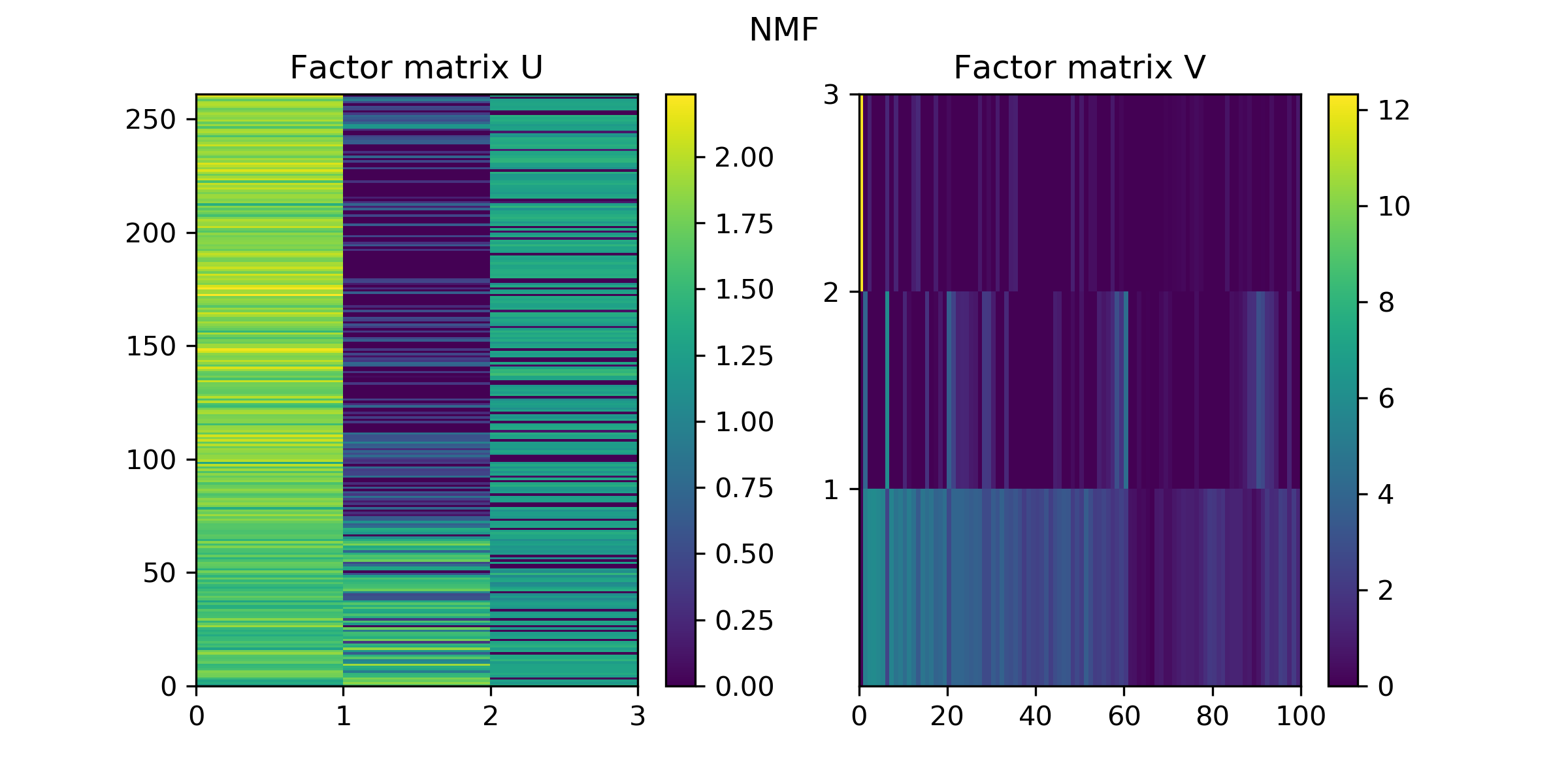}
    \caption{Factor matrices $U_{\texttt{NMF}}, V_{\texttt{NMF}}$ from \texttt{NMF}.}
\end{subfigure}
    \caption{Factor matrices $U_{\texttt{STMF}}, V_{\texttt{STMF}}$ and  $U_{\texttt{NMF}}, V_{\texttt{NMF}}$ from \texttt{STMF} and \texttt{NMF} on \texttt{SARC} data,  respectively.}
    \label{sarc_factor_matrices}
\end{figure}

\begin{figure}[!htb] 
    \centering
    \begin{subfigure}[t]{.5\textwidth}
        \includegraphics[height=0.42\textwidth, width=\textwidth]{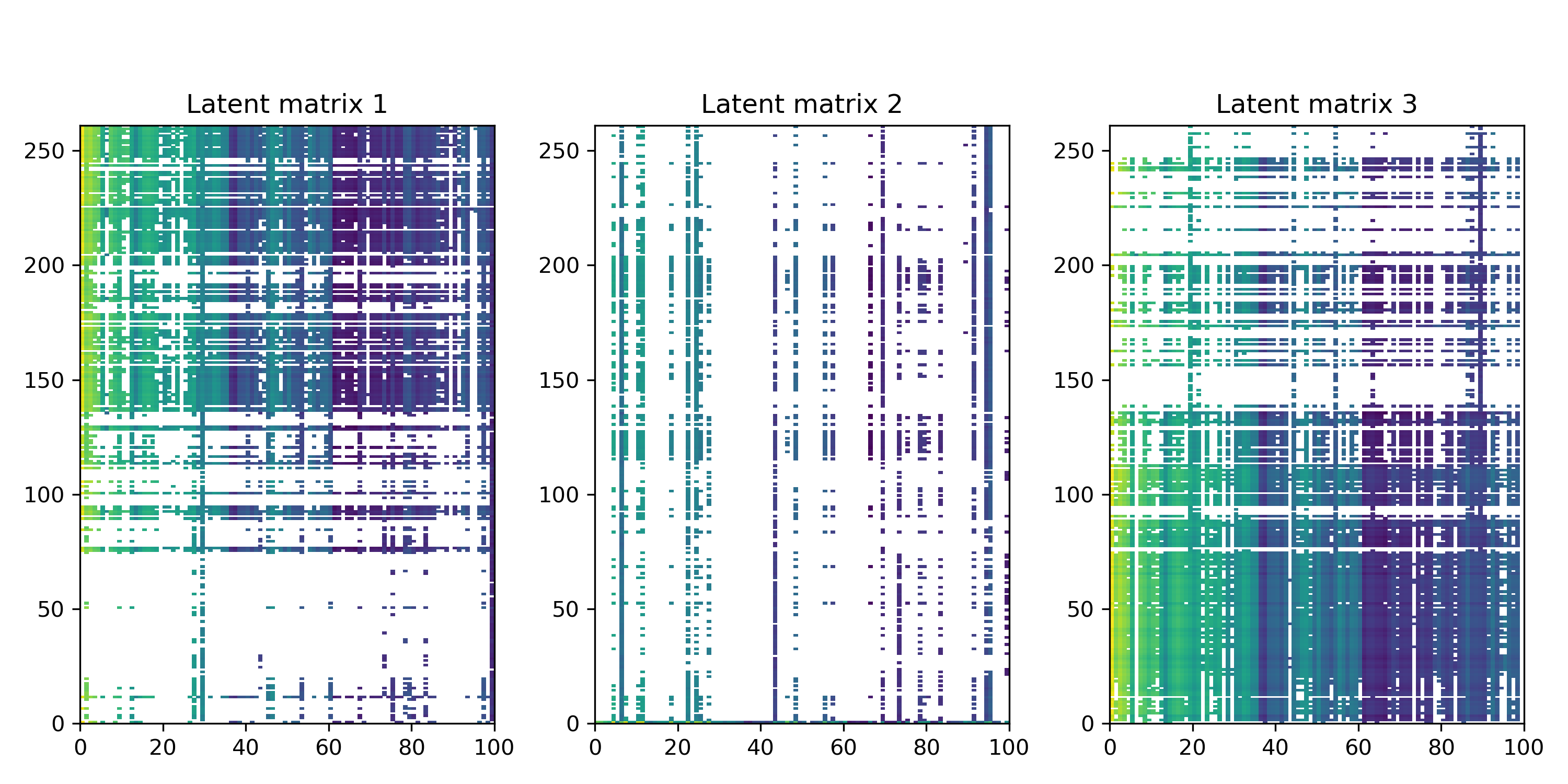}
        \caption{Latent matrices $R_{\texttt{STMF}}^{(i)}$, $i \in \{1, 3\}$, where \\ white represents the element which does not \\ contribute to the approximation $R_{\texttt{STMF}}$.}
    \end{subfigure}\hfill%
    \begin{subfigure}[t]{.5\textwidth}
        \includegraphics[height=0.42\textwidth, width=\textwidth]{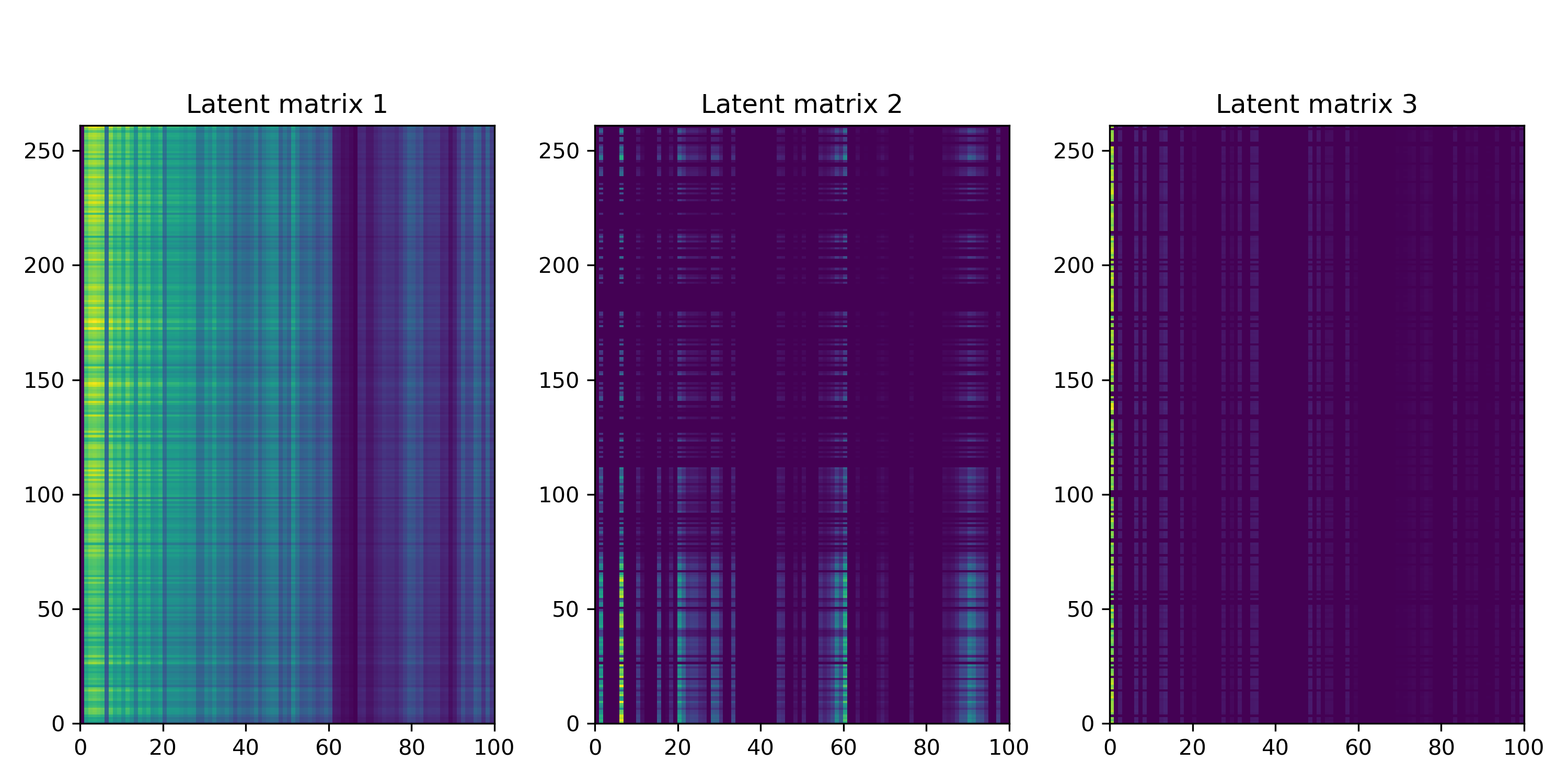}
        \caption{Latent matrices $R_{\texttt{NMF}}^{(i)}$, $i \in \{1, 3\}$.}
    \end{subfigure}
    \caption{\texttt{STMF}'s and \texttt{NMF}'s latent matrices on \texttt{SARC} data.}
    \label{sarc_latent_matrices}
\end{figure}

\begin{figure*}[!htb]
        \centering
        \begin{subfigure}[t]{0.32\textwidth}
            \centering
            \includegraphics[width=\textwidth]{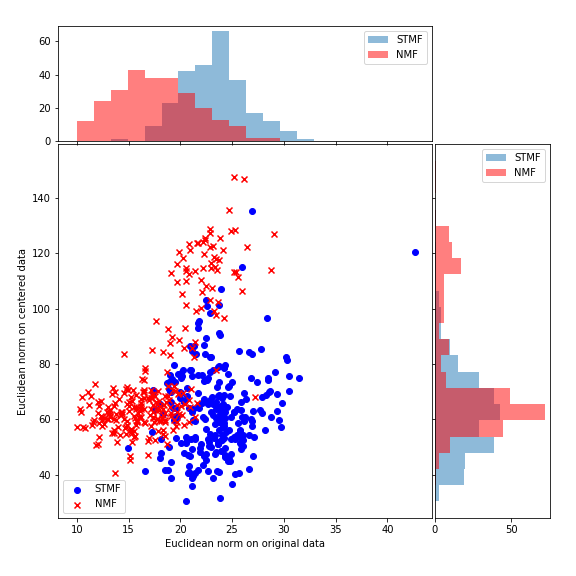}
            \caption{Euclidean norm on centered data}
        \end{subfigure}
        \hfill
        \begin{subfigure}[t]{0.32\textwidth}
            \centering 
            \includegraphics[width=\textwidth]{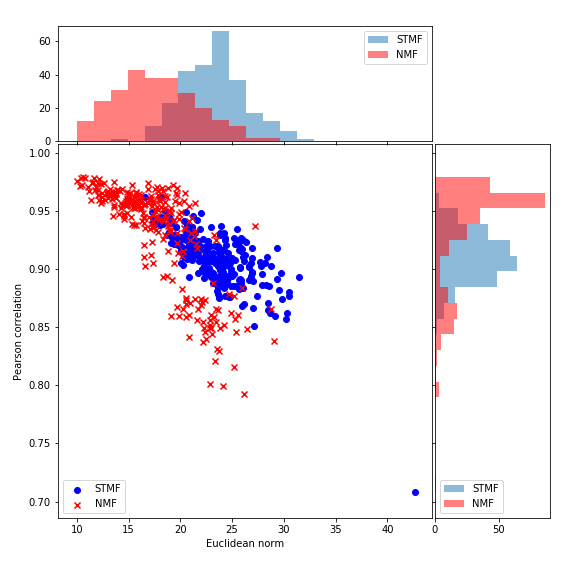}
            \caption{Pearson correlation}
        \end{subfigure}
        \hfill
        \begin{subfigure}[t]{0.32\textwidth}
            \centering 
            \includegraphics[width=\textwidth]{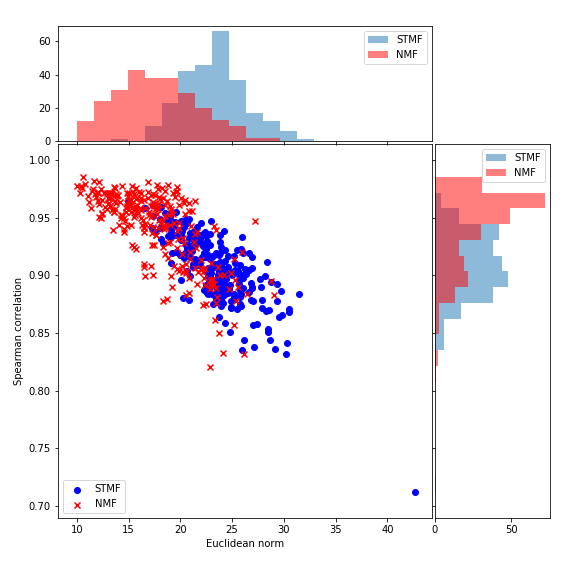}
            \caption{Spearman correlation}
        \end{subfigure}
        \caption{Euclidean norm on centered data, Pearson and Spearman correlation on \texttt{SARC} data.}
        \label{sarc_corrs}
\end{figure*}

\end{document}